%% file: arxiv.tex
\documentclass[10pt,twocolumn,letterpaper]{article}

\usepackage{style/paper}
\usepackage{times}
\usepackage{epsfig}
\usepackage{graphics}
\usepackage{graphicx}
\usepackage{amsmath}
\usepackage{amssymb}
\usepackage{bm}
\usepackage{mathtools}
\usepackage{enumitem}
\usepackage{subcaption}
\usepackage{multirow}
\usepackage[numbers,sort]{natbib}
\usepackage{tabularx}
\usepackage{tablefootnote}

\setlist{leftmargin=0.4cm}
\makeatletter
\newcommand\blfootnote[1]{%
  \begingroup
  \renewcommand\thefootnote{}\footnote{#1}%
  \addtocounter{footnote}{-1}%
  \endgroup
}
\makeatother

\usepackage[pagebackref=true,breaklinks=true,letterpaper=true,colorlinks,bookmarks=false]{hyperref}

\paperfinalcopy 

\begin{document}

\title{Neural Haircut: Prior-Guided Strand-Based Hair Reconstruction}

\author{
Vanessa Sklyarova$^{1}$ \quad 
Jenya Chelishev$^{2,*}$ \quad 
Andreea Dogaru$^{3,*}$ \quad 
Igor Medvedev$^{1}$ \\ 
Victor Lempitsky$^{4}$ \quad 
Egor Zakharov$^{1}$
\vspace{0.3cm}\\
$^1$Samsung AI Center \ \ $^2$Rockstar Games \ \ $^3$FAU Erlangen-Nürnberg \ \ $^4$Cinemersive Labs
}

\ifpaperfinal\thispagestyle{empty}\fi

\input{figures/teaser/teaser}

\input{parts/abstract.tex}
\blfootnote{$^*$ Work done at Samsung AI Center}
\blfootnote{$^\dagger$ https://samsunglabs.github.io/NeuralHaircut/}

\input{notation}

\input{parts/intro.tex}
\input{parts/related.tex}
\input{figures/optim/optim}
\input{parts/method.tex}
\input{parts/experiments.tex}
\input{parts/conclusion.tex}

\input{parts/acknowledgements}

{\small
\bibliographystyle{style/ieee_fullname}
\bibliography{refs}
}

\clearpage
\appendix

\input{parts_suppmat/method}

\input{parts_suppmat/experiments}

\end{document}

%% file: figures/teaser/teaser.tex
\twocolumn[{%
\renewcommand\twocolumn[1][]{#1}%
    \maketitle
    \begin{center}
        \vspace{-0.9cm}
        \centering
        \includegraphics[width=\textwidth]{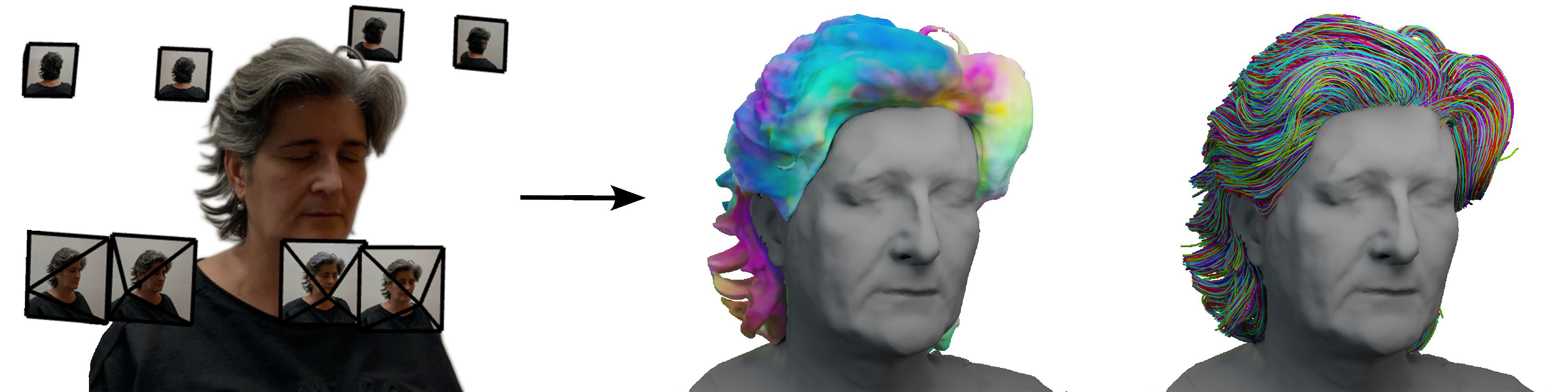}\vspace{0.05cm}
        \begin{tabular}{ccc}
            \hspace{-0.02\textwidth}\multirow{2}{*}{\textbf{Multi-view images}}\hspace{0.16\textwidth} & \textbf{Stage I: Volumetric hair}\hspace{0.08\textwidth} & \textbf{Stage II: Strand-based}\hspace{-0.08\textwidth} \\
            & \textbf{reconstruction}\hspace{0.08\textwidth} & \textbf{reconstruction}\hspace{-0.08\textwidth}
        \end{tabular}
        \vspace{-0.3cm}
        \captionof{figure}{We propose a two-stage pipeline for image-based hair reconstruction. Our first stage reconstructs coarse hair, head, and shoulder geometry using volumetric representations. The second stage fits hair strands to the coarse reconstruction via a joint optimization process that incorporates rendering-based losses and priors learned on the synthetic data.}
        \label{fig:teaser}
    \end{center}
}]

%% file: parts/abstract.tex
\begin{abstract}
   Generating realistic human 3D reconstructions using image or video data is essential for various communication and entertainment applications. While existing methods achieved impressive results for body and facial regions, realistic hair modeling still remains challenging due to its high mechanical complexity. This work proposes an approach capable of accurate hair geometry reconstruction at a strand level from a monocular video or multi-view images captured in uncontrolled lighting conditions. Our method has two stages, with the first stage performing joint reconstruction of coarse hair and bust shapes and hair orientation using implicit volumetric representations. The second stage then estimates a strand-level hair reconstruction by reconciling in a single optimization process the coarse volumetric constraints with hair strand and hairstyle priors learned from the synthetic data. To further increase the reconstruction fidelity, we incorporate image-based losses into the fitting process using a new differentiable renderer. The combined system, named Neural Haircut, achieves high realism and personalization of the reconstructed hairstyles. For video results, please refer to our project page$^\dagger$.
\end{abstract}

%% file: notation.tex
\def\L{\mathcal{L}} 
\def\S{\mathbf{S}} 
\def\p{\mathbf{p}} 
\def\b{\mathbf{b}} 
\def\l{\mathbf{L}} 
\def\d{\mathbf{d}} 
\def\g{\mathbf{g}} 
\def\Zgeom{\mathbf{T}} 
\def\zgeom{\mathbf{z}} 
\def\z{\mathbf{z}} 
\def\y{\mathbf{y}} 
\def\x{\mathbf{x}} 
\def\Zapp{\mathbf{A}} 
\def\zapp{\mathbf{a}} 
\def\E{\mathcal{E}} 
\def\G{\mathcal{G}} 
\def\fhair{f_\text{hair}} 
\def\fbust{f_\text{bust}} 
\def\ahair{\alpha^\text{hair}} 
\def\abust{\alpha^\text{bust}} 
\def\mhair{\mathbf{m}_\text{hair}} 
\def\mbust{\mathbf{m}_\text{bust}} 
\def\m{\mathbf{m}} 
\def\Softrast{\mathcal{R}} 
\def\Pcam{\mathcal{P}} 
\def\D{\mathcal{D}} 
\def\F{\mathcal{F}} 
\def\v{\mathbf{v}} 
\def\ohair{\hat{\mathbf{o}}_\text{hair}} 
\def\obust{\hat{\mathbf{o}}_\text{bust}} 
\def\I{\mathbf{I}} 
\def\n{\mathbf{n}} 
\def\lhair{l_\text{hair}} 
\def\lbust{l_\text{bust}} 

%% file: parts/intro.tex
\section{Introduction}

We propose a new image-based modeling method that recovers human hair from multi-view photographs or video frames. Hair reconstruction remains one of the most challenging problems in human 3D modeling because of its highly complex geometry, physics, and reflectance. Nevertheless, it is critical for many applications, such as special effects, telepresence, and gaming.

In computer graphics, the dominant representation for hair is 3D polylines, or \emph{strands}, which can facilitate both realistic rendering and physics simulation~\cite{Blender}. At the same time, modern image- and video-based human reconstruction systems often model hairstyles using data structures that have fewer degrees of freedom and are easier to estimate, such as meshes with fixed topology~\cite{Grassal2021NeuralHA, Khakhulin2022RealisticOM} or volumetric representations~\cite{Lombardi2019NeuralV, Lombardi2021MixtureOV, Wang2021PriorGuidedM3, Ramon2021H3DNetFH, neus, Fu2022GeoNeus, Yariv2021VolumeRO, Darmon2021ImprovingNI, Zheng2021IMA, Park2020NerfiesDN, Park2021HyperNeRFAH, Gafni2020DynamicNR, Cao2022AuthenticVA, Hong2021HeadNeRFAR, Mihajlovi2022KeypointNeRFGI, Zheng2022PointAvatarDP, Giebenhain2022LearningNP}. As a result, these methods often obtain over-smoothed hair geometries and can only model the ``outer shell'' of the hairstyle without its inner structure.

Accurate strand-based hair reconstruction can be accomplished via controlled lighting equipment and dense capture setup with synchronized cameras, i.e.\ using \emph{light stages}~\cite{Debevec2000AcquiringTR}. Recently, impressive results were achieved~\cite{Nam2019StrandAccurateMH, McGuire2021HumanHI, neuralstrands, Wang2022NeuWigsAN, Wang2021HVHLA} by relying on uniform or structured lighting and camera calibration to facilitate the reconstruction process. The latest work~\cite{neuralstrands} further utilized manual frame-wise annotation of the hair growth directions to achieve physically plausible reconstructions. However, despite the impressive quality of the results, the sophisticated capture setup and the manual pre-processing requirements make such methods unsuitable for many practical applications. Some learning-based methods for hairstyle modeling~\cite{Zhang2019HairGANR3, Zhou2018HairNetSH, Wu2022NeuralHDHairAH, Yang2019DynamicHM, Saito20183DHS, Hu2015SingleviewHM, Chai2016AutoHairFA, Kuang2022DeepMVSHairDH} incorporate hair priors learned from the strand-based synthetic data to ease the acquisition process. However, the accuracy of these methods naturally depends on the size of the training dataset. Existing datasets~\cite{Hu2015SingleviewHM, Wu2022NeuralHDHairAH} typically consist of only a few hundred samples and are inadequately small for handling the diversity of human hairstyles, leading to the low fidelity of the reconstructions.

In this work, we propose a method for hair modeling that uses only image- or video-based data without any additional manual annotations and works in uncontrolled lighting conditions. To achieve that, we have designed a two-stage reconstruction pipeline. The first stage, \emph{coarse} volumetric hair reconstruction, employs implicit volumetric representations and is purely data-driven. The second stage, \emph{fine} strand-based reconstruction, operates at the level of hair strands and relies heavily on priors learned from a small-scale synthetic dataset.

During the first stage, we reconstruct implicit surface representations~\cite{Park2019DeepSDFLC} for hair and bust (head and shoulders) regions. Additionally, we learn a field of hair growth directions, which we call 3D orientations, by matching them through a differentiable projection with hair directions observed in the training images or 2D orientation maps. While this field can facilitate a more accurate hair shape fitting, its primary use case is to constrain the optimization of hair strands during the second stage. To calculate the hair orientation maps from the input frames, we use a classic approach based on image gradients~\cite{Paris2004CaptureOH}.

The second stage relies on pre-trained priors to obtain strand-based reconstructions. We employ an improved parametric model learned from the synthetic data using an auto-encoder~\cite{neuralstrands} to represent individual strands and combine it with a new diffusion-based prior~\cite{ddpm, Karras2022ElucidatingTD} to model their joint distribution, i.e.\ a complete hairstyle. This stage thus reconciles the coarse hair reconstruction obtained in the first stage with the learning-based priors through an optimization process. Lastly, we improve the fidelity of reconstructed hairstyles via differentiable rendering using a new hair renderer based on soft rasterization~\cite{Liu2019SoftRA}.

To summarize, our contributions are:
\vspace{-0.25cm}
\begin{itemize}
    \itemsep-0.1cm
    \item Human head 3D reconstruction method for bust and hair regions, which includes hair orientations;
    \item Improved training procedure for the strand prior;
    \item Latent diffusion-based prior for global hairstyle modeling, which ``interfaces'' with the parametric strand prior;
    \item Differentiable soft hair rasterization technique that leads to more accurate reconstructions than the previous rendering methods;
    \item Strand-fitting process that incorporates all the components discussed above to produce high-quality reconstructions of human hair at the level of strands.
\end{itemize}
\vspace{-0.15cm}

We validate the efficacy of our method on synthetic~\cite{Yuksel2009HairM} and real-world data, for which we use multi-view images from a 3D scanner operating in unconstrained lighting conditions~\cite{Ramon2021H3DNetFH} and monocular videos from a smartphone.

%% file: parts/related.tex
\section{Related work}

\textbf{Human head reconstruction.} Modern approaches have achieved impressive results in modeling static and dynamic human subjects using image and video data. These methods primarily rely on shape priors trained using synthetic datasets or 3D scans. Among the most widespread priors are parametric models~\cite{FLAME:SiggraphAsia2017, 
Pavlakos2019ExpressiveBC, Loper2015SMPLAS, Osman2020STARST, Osman2022SUPRAS} that represent the head as a rigged mesh with a fixed topology. However, these models do not include hair, as it is known to be notoriously difficult to scan. This sparked the development of methods that extend parametric models to include hair using volumetric representations and image or video-based finetuning~\cite{Grassal2021NeuralHA, Khakhulin2022RealisticOM, Gafni2020DynamicNR, Zheng2021IMA, Wang2021PriorGuidedM3, Zheng2022PointAvatarDP} or, more recently, using higher quality 3D scanners~\cite{Ramon2021H3DNetFH, Giebenhain2022LearningNP}. While successfully reconstructing the facial geometry, these methods still achieve low fidelity of hair. They also model only the visible \emph{outer} hair surface and do not reconstruct the inner geometry, limiting the downstream applications.

Hair reconstruction using volumetric representations still has important advantages. Modern volumetric reconstruction methods can handle challenging lighting conditions due to view-dependent modeling of radiance~\cite{Mildenhall2020NeRFRS}. They can also hallucinate the geometry of the outer surface regions unseen in the training samples~\cite{neus, Yariv2021VolumeRO, Gropp2020ImplicitGR}, which is useful for reconstruction using video data. We found volumetric reconstruction methods ideal for use in the first stage, during coarse hair modeling. We further extend them to represent hair and bust geometries separately, which is not done in the previous works. That allows us to incorporate additional supervision for the geometry and more effectively constrain hair strand optimization during the second stage. In addition, the separate bust geometry is used in an auxiliary way to introduce proper occlusion handling into the volumetric and strand-based hair rendering.

\textbf{Strand-based hair reconstruction.} Starting with the seminal work~\cite{Paris2004CaptureOH} on hair reconstruction, image-based methods~\cite{Zhou2018HairNetSH, Wu2022NeuralHDHairAH, Yang2019DynamicHM, Saito20183DHS, Hu2015SingleviewHM, Nam2019StrandAccurateMH, neuralstrands, Kuang2022DeepMVSHairDH} have relied on hair orientation maps~\cite{Paris2004CaptureOH}, or gradients in the image space, to estimate 3D hair strands. These orientation maps can quite effectively bridge the sim-to-real gap and allow some of these methods~\cite{Saito20183DHS, Zhou2018HairNetSH, Kuang2022DeepMVSHairDH} to train using \emph{only} synthetic data, while still generalizing to the real-world images. However, they have multiple practical issues. In order to obtain high-quality orientation maps, hair must be uniformly lit and have no specular highlights. This assumption is quite strong and allows impressive results~\cite{Nam2019StrandAccurateMH, neuralstrands} to be achieved using light stages for data capture. However, it limits the practicality of such methods for real-world reconstruction cases where lighting conditions are hard to control. Non-uniform lighting and low effective resolution of the real-world data lead to orientation maps having excessive noise levels or lacking details. Furthermore, orientation maps encode ``sign-free'' local orientations and do not capture the hair growth direction, which adds extra ambiguity to the reconstruction process.

Some methods~\cite{neuralstrands, Hu2015SingleviewHM} address this using manual annotation of the exact growth direction, which makes the reconstruction process labor-intensive. Others~\cite{Chai2016AutoHairFA, Kuang2022DeepMVSHairDH} employed regressors trained via manual annotation to predict the hair growth directions. However, the solution and the dataset presented in these works are closed-source and remain challenging and costly to reproduce. Lastly, the majority of the strand-based reconstruction method~\cite{Zhang2019HairGANR3, Zhou2018HairNetSH, Wu2022NeuralHDHairAH, Yang2019DynamicHM, Saito20183DHS, Nam2019StrandAccurateMH, Chai2016AutoHairFA, Kuang2022DeepMVSHairDH} model hair strands without explicitly attaching them to the head scalp, which limits the realism of the resulting reconstructions. Our method addresses all these issues by introducing new hairstyle priors that ensure the physical realism of the reconstructed strands and a new coarse-to-fine optimization pipeline that uses prior-guided optimization and differentiable rendering to obtain personalized reconstructions even in non-uniform lighting conditions.




%% file: figures/optim/optim.tex
\begin{figure*}
    \centering
    \includegraphics[width=0.95\textwidth]{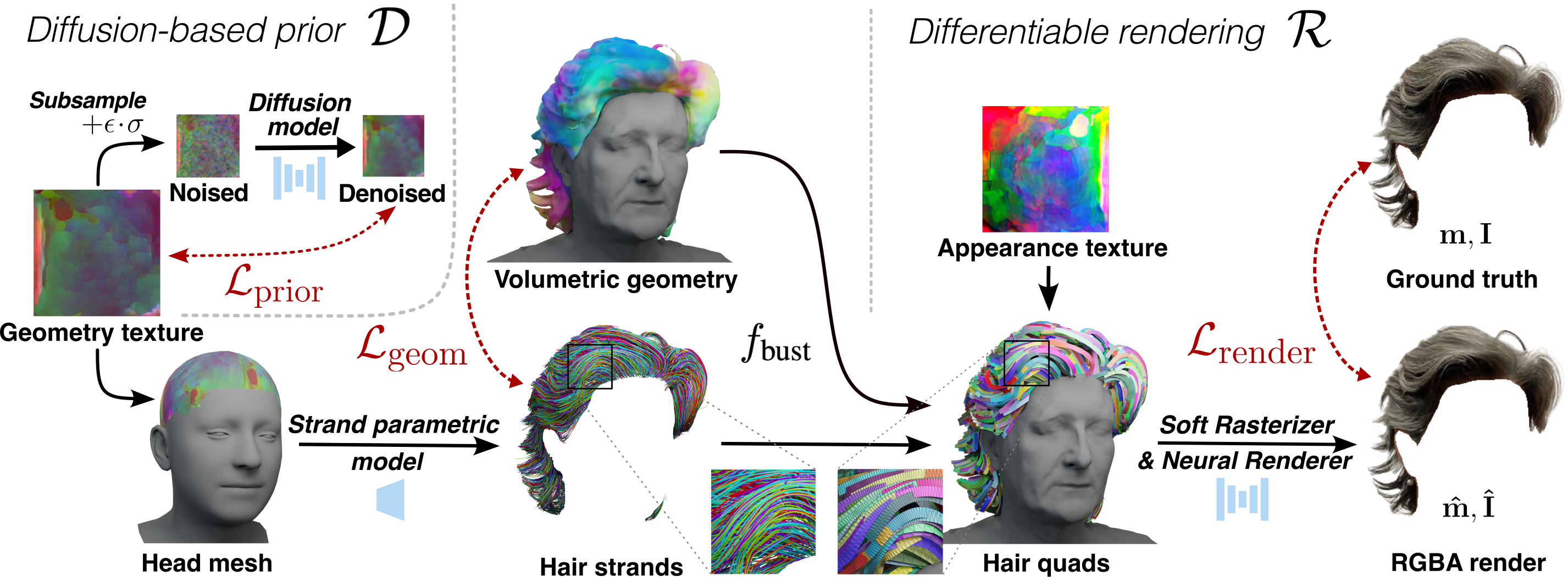}
    \caption{The overview of the second stage of our approach (fine strand-based reconstruction). We use shape texture to represent hair strands and utilize multiple objectives to optimize it. We apply $\mathcal{L}_\text{prior}$ as a regularization penalty using a diffusion network pre-trained on synthetic hairstyles. Then, we use $\mathcal{L}_\text{geom}$ to match the reconstructed strands to geometry and orientation fields parameterized by the implicit function. Lastly, $\mathcal{L}_\text{render}$ is used to match the rendered hair to the ground truth image.}
    \label{fig:optim}
    \vspace{-0.3cm}
\end{figure*}

%% file: parts/method.tex
\section{Method}


\subsection{Overview}

We reconstruct the strand-based hair geometry given a single monocular video or multi-view images in the form of polylines in 3D: $\S = \{ \p^l \}_{l=1}^L$. Our hair reconstruction pipeline consists of two stages. 
First, we obtain a coarse volumetric hair reconstruction in the form of implicit fields. We then reconstruct fine hair strands using optimization of coarse geometry-based, rendering-based, and prior-based terms. The hairstyle prior is obtained separately during pre-training on a synthetic dataset.

\textbf{Hair prior training.} Following~\cite{neuralstrands}, we parameterize the hairstyle using a \emph{latent geometry texture} defined on the head scalp and denoted as $\Zgeom$. The mapping between hair strands and their latent embeddings is provided by the hair parametric model. It has the same architecture and training procedure as the original approach~\cite{neuralstrands}, besides a modified data term that improves the fidelity of curly hair reconstructions. We denote the decoder that produces strands given their latent embeddings as $\G$ and an encoder as $\E$.

We then train a latent diffusion-based prior $\D$, defined on the geometry texture maps $\Zgeom$. We use EDM~\cite{Karras2022ElucidatingTD} formulation that outperforms previous approaches such as DDPM~\cite{ddpm}. We introduce multiple data augmentations that preserve the realism of the hairstyle while training on a small dataset of hairstyles~\cite{Hu2015SingleviewHM} consisting only of a few hundred samples.

\textbf{Stage I: coarse volumetric reconstruction.} We approach coarse reconstruction by estimating hair and bust geometry as signed distance functions (SDFs) $\fhair, \fbust: \mathbb{R}^3 \rightarrow \mathbb{R}$. We train them via volumetric ray marching~\cite{neus} using a shared view-dependent color field $c: \mathbb{R}^3 \times \mathbb{S}^2 \rightarrow \mathbb{R}$. We employ supervision via semantic segmentation masks to ensure that hair and bust regions are non-overlapping. Also, to correctly reconstruct the head scalp, which is typically not visible on training samples, we fit a FLAME~\cite{FLAME:SiggraphAsia2017} head mesh to the scene and use it as a prior for the bust SDF. Lastly, to facilitate strand-based reconstruction, we train an additional field of 3D hair orientations $\beta: \mathbb{R}^3 \rightarrow \mathbb{S}^2$ using the hair signed distance function and match its projections with observed hair strand orientations.

\textbf{Stage II: fine strand-based reconstruction.} We reconstruct hair strands as a geometry texture $\Zgeom$, i.e.\ a dense two-dimensional map of latent hair vectors, where the position on the map corresponds to the position of the hair root on the scalp. At each iteration, we sample $N$ random embeddings $\{ \zgeom_i \}_{i=1}^N$ from the texture $\Zgeom$ and obtain corresponding strands $\{ \S_i \}_{i=1}^N$ using a pre-trained decoder $\G$. These strands are then used to evaluate geometric and rendering-based constraints. In the geometric loss, we penalize strands outside the hair volume and ensure that the visible part of the surface defined by $\fhair$ is uniformly covered. We also match the orientations $\b^l_i$ of the predicted strands, defined as the normalized difference between two consecutive points, to the orientation field $\beta$. 
Here, $\b^l_i = \d^l_i \big/ \|\d^l_i\|_2$ and $\d^l_i =  \p^{l+1}_i-\p^l_i$.

Besides the geometric constraints, we also employ silhouette-based and neural rendering losses. The rendered hair silhouette $\hat\m$ and RGB image $\hat\I$ are then obtained using neural soft hair rasterization denoted as $\Softrast$. The renderer employs a bust surface estimated from $\fbust$ to handle occlusions. The silhouette $\hat\m$ is predicted directly from the sampled strands, while the image render is obtained via a neural hair rendering pipeline inspired by~\cite{neuralstrands}.

Lastly, prior-based regularization is applied directly to the geometry texture $\Zgeom$ using a pre-trained diffusion model. Specifically, we apply random noise to the geometry map and denoise it using a diffusion model $\D$. We then evaluate the reconstruction error of the input map $\Zgeom$ and back-propagate the gradient of this loss back into the texture. This pipeline is inspired by the DreamFusion method~\cite{poole2022dreamfusion}, albeit with some modifications which facilitate training from a small dataset of hairstyles.

The scheme of the fine reconstruction stage is shown in Figure~\ref{fig:optim}. Below, we describe the parts of our approach in more detail.

\subsection{Hair prior training}

Our global hairstyle prior is trained using a strand parametric model and a latent diffusion network, which interface with each other via the geometry texture $\Zgeom$.

\textbf{Hair strand parametric model}. To obtain the latent representations for the hair strands, we follow~\cite{neuralstrands} and train a variational autoencoder, which maps a strand $\S = \{\p^l\}_{l=1}^L$ into a latent vector $\zgeom$ via encoder $\E$, and back via decoder $\G$. During training, we employ the reparameterization trick: $\zgeom = \z_\mu + \epsilon \cdot \z_\sigma$, where $\z_\mu$ and $\z_\sigma$ are parameters of the Gaussian distribution, and $\epsilon \sim \mathcal{N}(\mathbf{0}, \mathbf{I})$. The model is trained using the data term $\L_\text{data}$ and KL divergence~\cite{Kingma2013AutoEncodingVB} term $\L_\text{KL}$ on a synthetic dataset of hair strands. The data term consists of an L2 error between predicted $\hat{\p}^l$ and ground-truth points $\p^l$, as well as the cosine distance for the orientations $\hat{\b}^l$ and $\b^l$, similarly to~\cite{neuralstrands}. Inspired by~\cite{Zhou2018HairNetSH}, we additionally match curvatures $\hat\g^l$ and $\g^l$ of predicted and ground-truth strands to better model curly hair, where
\begin{equation}
    \g^l = \big\| \b^l \times \b^{l+1} \big\|_2.
\end{equation}
Our data term is therefore defined as follows:
\begin{equation}
\begin{split}
   \L_\text{data} = \sum_{l=1}^L \big\| \hat{\p}^l - \p^l \big\|_2^2 & + \lambda_d \big( 1 - \hat\b^l \b^l \big) \\ & + \lambda_c \big\| \hat\g^l - \g^l \big\|_2^2,
\end{split}
\end{equation}
and the final loss is:
\begin{equation}
   \L_\text{VAE} = \L_\text{data} + \lambda_\text{KL} \L_\text{KL} \big( \mathcal{N}(\z_\mu, \z_\sigma) \,\big\|\, \mathcal{N}(\mathbf{0}, \mathbf{I}) \big).
   \label{eq:strand_prior}
\end{equation}

\textbf{Hairstyle diffusion model.} We obtain the latent representation of the synthetic hairstyle consisting of $N$ strands $\{\S_i\}_{i=1}^N$ by first estimating their latent descriptors $\{ \zgeom_i \}_{i=1}^N$ using the pre-trained strand encoder $\E$. We then convert them into a dense texture $\Zgeom$ using nearest neighbor interpolation. To increase the diversity of training samples, we employ augmentations that preserve the realism of the hairstyle before encoding. Also, to speed up the training and further diversify the inputs of the diffusion model, we subsample the full texture into a low-resolution map $\Zgeom_\text{LR}$.

We use the Elucidating Diffusion Moldel (EDM)~\cite{Karras2022ElucidatingTD} to train the denoiser $\D$. Below, we denote a training sample $\Zgeom_\text{LR}$ as $\y$ to be consistent with~\cite{Karras2022ElucidatingTD}, and obtain a noised input: $\x = \y + \epsilon \cdot \sigma$, where $\epsilon \sim \mathcal{N}(\mathbf{0}, \mathbf{I})$, and $\sigma$ is a noise strength. We then predict a denoised input:
\begin{equation}
    \D(\x, \sigma) = c_\text{skip}(\sigma) \cdot \x + c_\text{out}(\sigma) \cdot \F \big( c_\text{in}(\sigma) \cdot \x, c_\text{noise}(\sigma) \big),
\end{equation}
where the $c_\text{skip}$, $c_\text{out}$, $c_\text{in}$ and $c_\text{noise}$ are part of pre-conditioning approach proposed in~\cite{Karras2022ElucidatingTD}, which improves the robustness of $\D$ to the low noise strength $\sigma$, and $\F$ is a neural network. Our training objective also follows~\cite{Karras2022ElucidatingTD}:
\begin{equation}\label{eq:l_diff}
    \mathcal{L}_\text{diff} = \mathbb{E}_{\,\y,\sigma,\epsilon} \Big[\, \lambda_\text{diff}(\sigma) \cdot \big\| \D(\x, \sigma) - \y \big\|_2^2 \,\Big],
\end{equation}
where $\lambda_\text{diff}(\sigma)$ is a weighting function, and the expectation is approximated via sampling.

\subsection{Coarse volumetric reconstruction}

We represent the subject's coarse head geometry in a segmented form using the hair and the bust neural signed distance functions~\cite{Park2019DeepSDFLC} (SDFs). We use the volumetric ray marching approach for neural implicit surfaces, NeuS~\cite{neus} to fit them. We modify NeuS to accommodate multi-label reconstruction as we reconstruct the hair geometry and the bust (head and shoulders) geometry as separate shapes. The training proceeds by approximating a pixel's color $\mathbf{c}$ using the radiance at $N$ points $\x_i$ sampled along the corresponding ray $\v$. The color is predicted as follows:
\begin{equation}
    \hat{\mathbf{c}} = \sum_{i=1}^N T_i \cdot \alpha_i \cdot c\,(\x_i, \v, l, \n),\quad T_i = \prod_{j=1}^{i-1} (1 - \alpha_i),
    \label{eq:vol_rendering}
\end{equation}
where $T_i$ is the accumulated transmittance, $\alpha_i$ is the opacity, l and $\n$ - the blended hair with bust features and normals correspondingly, and $c$ is the view-dependent radiance field. We calculate the opacity $\alpha_i$ of each point along the ray by blending the individual opacities of hair and bust:
\begin{equation}
    \alpha_i = \min \big( \ahair_i + \abust_i,\, 1 \big).
\end{equation}
Besides the color, we also render the bust and the hair masks:
\begin{equation}
    \ohair = \sum_{i=1}^N T_i \cdot \ahair_i,\quad  \obust = \sum_{i=1}^N T_i \cdot \abust_i.
\end{equation}

Our training losses include a photometric L1 loss $\L_\text{color}$, which matches $\hat{\mathbf{c}}$ and $\mathbf{c}$, a mask-based loss $\L_\text{mask}$ that applies binary cross-entropy between the predicted masks and the ground-truth $\mhair$ and $\mbust$, and the regularizing Eikonal term~\cite{Gropp2020ImplicitGR} $\L_\text{reg}$, which is applied for both $\fhair$ and $\fbust$.

Our additional losses include a regularization for the bust shape. Before proceeding with the coarse reconstruction of the subject, we fit a FLAME~\cite{FLAME:SiggraphAsia2017} head mesh into the scene using optimization based on 2D facial landmarks~\cite{bulat2017far}. Using this mesh, we ensure that $\fbust$ includes the head scalp surface region by applying the regularizing constraints denoted as $\L_\text{head}$ that match the SDF to the mesh. To implement this loss, we follow the previous works~\cite{Atzmon2019SALSA, Gropp2020ImplicitGR, Sitzmann2020ImplicitNR} on fitting neural SDFs using mesh-based data and provide its full description in the supplementary materials.

Lastly, we incorporate an additional field of hair growth directions, $\beta$, into the coarse reconstruction. We train it via a differentiable surface rendering~\cite{Fu2022GeoNeus} of $\fhair$. Following~\cite{Fu2022GeoNeus}, we obtain the intersection point $\x_\text{s}$ of the ray $\v$ with the hair surface. We then project the 3D orientation field $\beta(\x_\text{s})$ into the camera $\Pcam$ using Plucker line
coordinates~\cite{Wrobel2001MultipleVG}. The projected direction $\l(\x_s, \beta(\x_\text{s}); \Pcam)$ is then matched to the 2D orientation map~\cite{Paris2004CaptureOH}, estimated from the training images using Gabor filters. The matching loss $\L_\text{dir}$ follows previous works~\cite{Paris2004CaptureOH} on strand-based reconstruction and penalizes the minimum angular difference between the projected and ground-truth orientations. Please refer to the supplementary materials for more details.

Overall, the training objective for the coarse reconstruction is as follows:
\begin{equation}
\begin{split}
    \L_\text{coarse} = \L_\text{color} & + \lambda_\text{mask} \L_\text{mask} + \lambda_\text{reg} \L_\text{reg} \\
    & + \lambda_\text{head} \L_\text{head} + \lambda_\text{dir} \L_\text{dir}.
\end{split}
\end{equation}

\subsection{Fine strand-based reconstruction}

To reconstruct the hair strands, we learn a latent hair geometry texture $\Zgeom$~\cite{neuralstrands}, from which a hairstyle can be decoded using a pre-trained network $\G$. However, instead of directly optimizing this map, we parameterize it with a UNet-like neural network using the so-called deep image prior~\cite{deepimageprior}. We found such parameterization to not require additional smoothing~\cite{neuralstrands} of the sparse gradients from the decoded strands. Below, we denote such new parameterization as $\Zgeom_\theta$. 

The training proceeds by sampling $N$ points on the scalp part of the fitted FLAME mesh and decoding them into strands $\{ \S_i \}_{i=1}^N$, each strand consisting of $L$ points: $\S_i = \{ \p^l_i \}_{l=1}^L$. We then evaluate the following objectives: geometry-based losses $\L_\text{geom}$ that match the strands to the coarse geometry, photometric constraints $\L_\text{render}$ calculated via differentiable rendering, and finally, a diffusion-based prior loss $\L_\text{prior}$. Below we describe them in more detail.

\textbf{Geometry-based losses.} To ensure that the optimized strands lie inside the coarse hair volume, we employ a loss $\L_\text{vol}$ that penalizes the points on the strands that stray outside of it:
\begin{equation}
    \L_\text{vol} = \sum_{i=1}^{N} \sum_{l=1}^L \mathbb{I} \big[ \fhair(\p^l_i) > 0 \big] \big( \fhair(\p^l_i) \big)^2,
\end{equation}
where $\mathbb{I}$ denotes the indicator function.

Additionally, to make the learned strands densely cover the visible part of the coarse hair surface, denoted as $\mathcal{S}$, we minimize the error between $K$ random points $\x_k$ sampled on this surface and their nearest points on the strands, denoted as $\p_k$. This loss $\L_\text{chm}$ is exactly equal to the one-way Chamfer distance between the visible part of the coarse hair surface and the learned strands:
\begin{equation}
    \L_\text{chm} = \sum_{k=1}^K \big\| \x_k - \p_k \big\|_2^2,
\end{equation}

Lastly, we calculate the distance between the hair orientations and the implicit field $\beta$ at all points on the strands that are closer to the visible hair surface $\mathcal{S}$ than some small threshold $\tau$. We denote these $M$ points as $\p_m$ and their orientations as $\b_m$. The resulting orientations loss $\L_\text{orient}$ can be written as follows:
\begin{equation}
\L_\text{orient} = \sum_{m=1}^{M} \big(\, 1 - \big| \b_m \cdot \beta(\p_m) \big|\, \big).
\end{equation}
We penalize the orientations of strands near the outer hair surface because the photometric nature of the orientation loss $\L_\text{dir}$ makes the field $\beta$ learn accurate orientations only in this region. We describe the procedure for estimating this surface using $\fhair$ and $\fbust$ in the supplementary materials.

Overall, the total geometry loss is the following:
\begin{equation}
    \L_\text{geom} = \L_\text{vol} + \lambda_\text{chm} \L_\text{chm} + \lambda_\text{orient}\L_\text{orient}.
\end{equation}

\textbf{Rendering-based losses.} We have developed a new approach for the differential rendering of hair strands to improve the visible hair geometry. We note that the previous hair rasterization approaches~\cite{neuralstrands} rely on graphics API~\cite{woo1999opengl} line rasterization algorithms, e.g.\ Bresenham's line algorithm \cite{bresenham}. While being computationally efficient, such methods only provide the gradients w.r.t.\ the first element of the line segments z-buffer, Figure~\ref{fig:rasterization_comparation} (a). At the same time, for the task of mesh inverse rendering, it was shown to be highly beneficial~\cite{Liu2019SoftRA} to propagate the gradient into multiple z-buffer elements. Inspired by the success of this \emph{soft rasterization} method~\cite{Liu2019SoftRA} for meshes, we adapt it for the differentiable rendering of hair strands Figure~\ref{fig:rasterization_comparation} (b).

First, we convert the hair strands into the so-called \emph{hair quads}~\cite{Yuksel2009HairM}. They consist of a stripe-like mesh, which follows the strand trajectory and has normals oriented toward the camera, see close-ups in Figure~\ref{fig:optim}. The vertices of the resulting quad mesh are fully differentiable w.r.t.\ the strands, and we include the quad generation algorithm into the supplementary materials. We then render this mesh using soft rasterization. We include the zero iso-surface of $\fbust$ obtained using Marching Cubes~\cite{Lewiner2003EfficientIO} into the rendering pipeline to handle hair-bust occlusions. Contrary to the previous rasterization methods, in our approach the segmentation mask for the hair is \emph{directly} predicted from the hair geometry using a soft silhouette shader~\cite{10.1145/3415263.3419160}, which allows unconstrained gradient flow into the geometry from the mask-based objectives. To render the color, we follow~\cite{neuralstrands} and use a neural rendering approach that can handle the view-dependent reflectance of the hair. Specifically, we train a neural appearance texture $\Zapp$ similarly to the geometry texture $\Zgeom$ and use it in conjunction with a rendering U-Net to produce the renders, similarly to~\cite{neuralstrands}. 

As a result of the hair rasterization pipeline $\Softrast$ described above, we obtain both the hair silhouette $\hat\m$ and the images $\hat\I$ in a fully differentiable way:
\begin{equation}
    \hat\m, \hat\I = \Softrast_\phi \big( \{ \S_i \}_{i=1}^N, \fbust,  \Pcam \big),
\end{equation}
where $\phi$ denotes the trainable parameters of the appearance texture and a rendering UNet, and $\mathcal{P}$ are the camera parameters. We then apply L1 losses $\L_\text{mask}$ and $\L_\text{rgb}$ to match the predicted silhouette and the color to the ground truth $\m$ and $\I$. The final rendering loss is the weighted sum of these terms:
\begin{equation}
    \L_\text{render} = \L_\text{rgb} + \lambda_\text{mask} \L_\text{mask}.
\end{equation}

\input{figures/rasterization/rasterization_comparation}

\textbf{Diffusion-based prior.} To apply the pre-trained diffusion prior, we use a Score Distillation Sampling (SDS) approach from the DreamFusion work~\cite{poole2022dreamfusion}. In this method, the pre-trained diffusion model is used to guide the optimization of a neural radiance field~\cite{Mildenhall2020NeRFRS} by providing it with the gradients in the image space. These gradients originate from the same loss used to train a diffusion model, in our case, $\L_\text{diff}$, Eq.~\ref{eq:l_diff}. However, instead of back-propagating this loss through the denoising neural network $\mathcal{F}$, the SDS approach assumes the gradients w.r.t. the noised input $\x$ to be identity: $\partial{\mathcal{F}} / \partial{\mathcal{\x}} = \mathcal{I}$. However, we found such a trick to be required only for DDPM~\cite{ddpm} training formulation used in DreamFusion, while for the EDM~\cite{Karras2022ElucidatingTD} that we use, proper back-propagation through the denoising network $\mathcal{F}$ leads to better results. Therefore, in our case, the prior regularization term $\L_\text{prior} \equiv \L_\text{diff}$. 

To calculate this loss, we employ the same procedure as during the training of the diffusion model. We sample random noise $\epsilon$ and the noise level $\sigma$ and apply them to the geometry map. Then, we perform random sub-sampling to decrease the resolution of $\Zgeom_\theta$ before forwarding it through the diffusion model. We back-propagate the loss $\L_\text{prior}$ directly into the parameters $\theta$ of the geometry texture $\Zgeom_\theta$ while keeping the weights of the denoiser frozen.

Overall, the optimization objective for the strand-based reconstruction stage is the following:
\begin{equation}
    \L_\text{fine} =    \L_\text{geom} + \lambda_\text{render} \L_\text{render} + \lambda_\text{prior} \L_\text{prior}.
\end{equation}

%% file: figures/rasterization/rasterization_comparation.tex
\begin{figure}
    \centering
    \includegraphics[width=\linewidth]{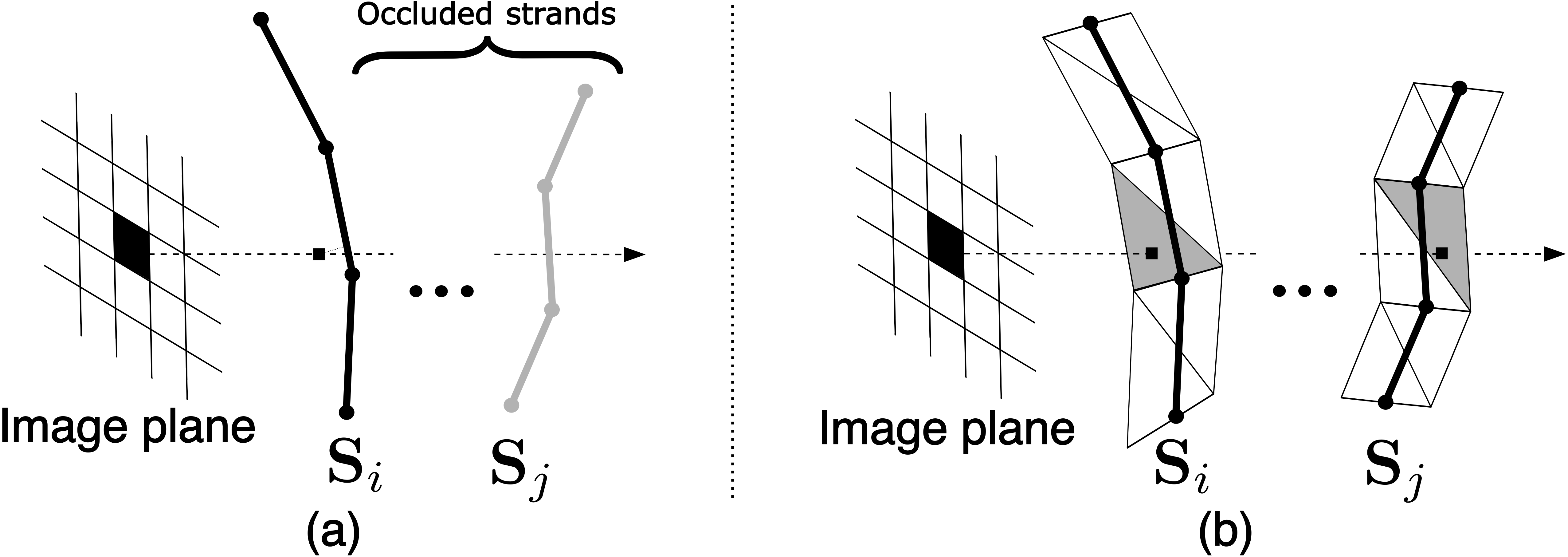}
    \vspace{-0.5cm}
    \caption{(a) Differentiable hair rasterization algorithm of~\cite{neuralstrands} propagates the gradient only into the first element of z-buffer. (b) Our proposed hair rasterization based on quads leverages soft hair rasterization~\cite{Liu2019SoftRA} and passes gradients into multiple elements of the z-buffer to achieve better reconstructions.\vspace{-0.3cm}}
    \label{fig:rasterization_comparation}
\end{figure}

%% file: parts/experiments.tex
\section{Experiments}

\input{figures/qualitative_comp/qualitative_comp}

We use the USC-HairSalon~\cite{Hu2015SingleviewHM} dataset to pre-train the strand parametric model and hairstyle diffusion module. This dataset consists of 343 hairstyles aligned with the template bust mesh. We then evaluate our method using both synthetic and real-world data. We use two synthetic scenes~\cite{Yuksel2009HairM} to conduct a quantitative comparison using the ground-truth strand-based geometry. For the real-world data, we use H3DS Dataset~\cite{Ramon2021H3DNetFH} of multi-view images with non-uniform lighting and monocular video data captured using a smartphone.


\subsection{Implementation details}


To train our method on real-world data we use off-the-shelf methods~\cite{MODNet, cdgnet} to obtain segmentation masks for the hair and bust. To parameterize the geometry texture map, we use a UNet network that predicts it from a constant mesh grid. Similarly to the diffusion model training, we calculate the denoising  error $\L_\text{prior}$ by averaging over the mini-batch of different offsets, noise $\epsilon$, and noise levels $\sigma$. In total, our reconstruction pipeline takes three days per subject on a single NVIDIA RTX 4090: one day for the first, and two days for the second stage. For more training details and hyperparameters, please refer to the supplementary materials.




\subsection{Real-world evaluation}


\textbf{Baselines.} We compare our method against popular 3D reconstruction approaches~\cite{neus, Oechsle2021ICCV}, as well as methods~\cite{Wu2022NeuralHDHairAH, Kuang2022DeepMVSHairDH} designed for strand-based reconstruction, using publicly available scenes from the H3DS~\cite{Ramon2021H3DNetFH} dataset. \textbf{NeuS}~\cite{neus} is a multi-view reconstruction approach that learns the scene geometry as the zero level-set of a signed distance function using volume rendering. This method can reconstruct non-Lambertian surfaces, making it well-suited for hair reconstruction. 
\textbf{UNISURF}~\cite{Oechsle2021ICCV} is another multi-view approach based on occupancy fields learned via volume rendering. This method is specifically tailored to handle semi-transparent objects, such as hair.
\textbf{DeepMVSHair}~\cite{Kuang2022DeepMVSHairDH} is a multi-view image-guided method for realistic strand-based reconstruction that can operate in a sparse-view scenario and under non-uniform lighting conditions. Due to memory constraints, this method is trained to produce reconstructions using twelve views. We also provide comparisons with a single image \textbf{NeuralHDHair}~\cite{Wu2022NeuralHDHairAH} method in the supplementary materials.


The qualitative results are shown in Figure~\ref{fig:geom_compare}. Note that our segmented modeling approach allows us to reconstruct realistic hair alongside the accurate bust geometry. It sets us apart from all the baseline methods, which can only reconstruct coarse hair geometry. Furthermore, our approach is able to reconstruct strands in poorly visible regions.

\input{figures/colmap_polina/polina}

Finally, we evaluate our method in a challenging case of monocular video capture. The results are provided in Figure \ref{fig:in-the-wild}. Our method is fully capable of handling this challenging use case and keeps the high fidelity and realism of the reconstructed hair. We include more examples in the supplementary materials. 

\input{figures/table_quant}
\input{figures/ablation_difu/ablation_difu}

\subsection{Ablation study}\label{sec:ablation}

We conduct an ablation study using both synthetic and real-world datasets. Here, we also include the comparison with \textbf{Neural Strands}~\cite{neuralstrands}, since our approach, in some aspects, builds on top of it. However, we cannot compare our method against it directly as this method is closed-source, requires manual annotation of the hair growth directions, and relies on an MVS-based hair reconstruction method~\cite{Nam2019StrandAccurateMH}, which is sensitive to non-uniform lighting conditions and poorly handles hair specularities.




We employed the same approach for quantitative evaluation as in \cite{neuralstrands}. We first render the ground-truth strands using Blender~\cite{Blender} and reconstruct them. We then follow~\cite{neuralstrands, Nam2019StrandAccurateMH} and measure precision, recall, and F-score between our predicted strands and ground truth using both distance and angular errors as thresholds. The comparison results are shown in Table~\ref{tab:comparison}. First, we see that our rendering loss $\L_\text{rgb}$ improves over the base $\L_\text{render}$ from~\cite{neuralstrands} in terms of precision, while the rendering mask loss $\L_\text{mask}$ achieves better recall and an aggregated F-score. Finally, our complete model $\L_\text{fine}$, which combines together geometric, rendering, and diffusion-based losses, further improves these results, achieving the highest recall and F-score across all experiments.

In Figure~\ref{fig:ablation}, we conduct a qualitative ablation study to evaluate the effect of a diffusion prior $\L_\text{prior}$ and a curvature loss term in the strand parametric model. Notice that the diffusion-based prior achieves substantially higher realism of the internal part of the hairstyle. For the curvature loss, its effect is visible when modeling curly hairstyles. Additional qualitative and quantitative results are provided in the supplementary materials.



%% file: figures/qualitative_comp/qualitative_comp.tex
\begin{figure*}
    \begin{tabular}{ccccc}
        \includegraphics[width=0.185\textwidth]{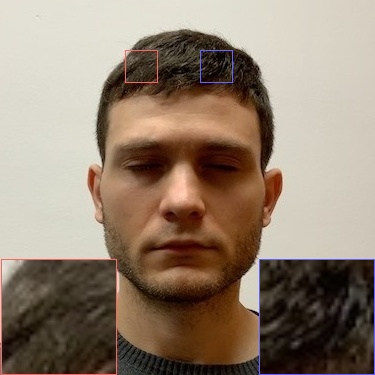} & \hspace{-0.31cm}
        \includegraphics[width=0.185\textwidth]{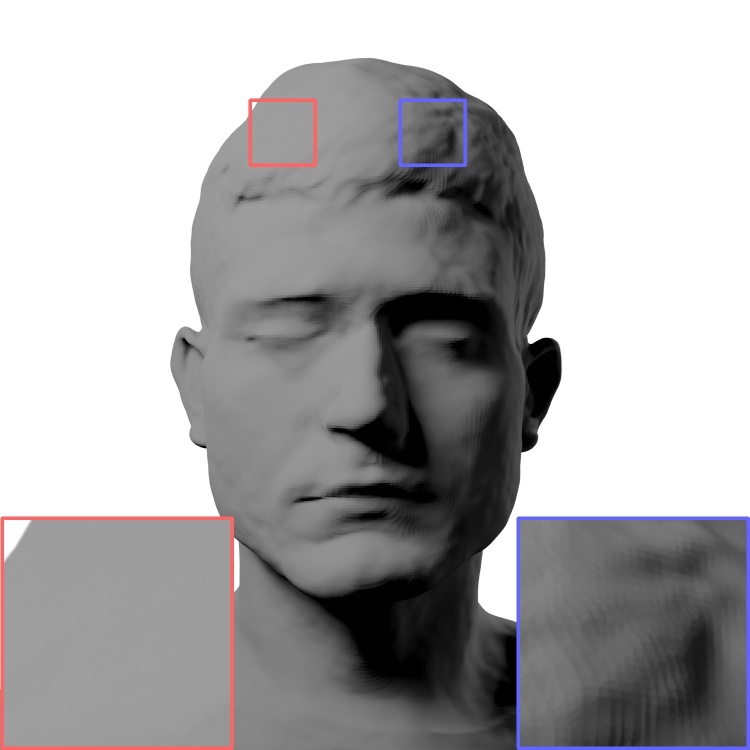} & \hspace{-0.31cm} 
        \includegraphics[width=0.185\textwidth]{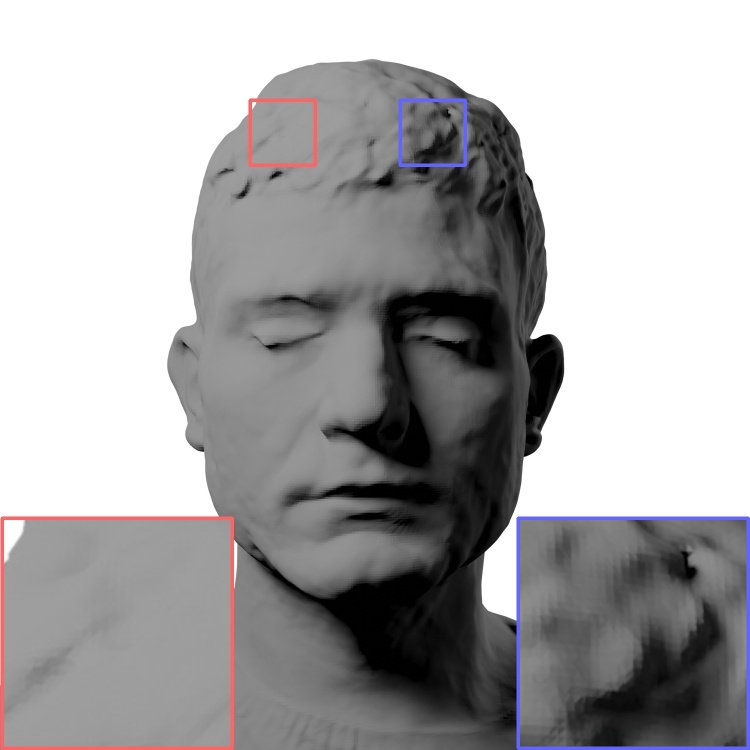} & \hspace{-0.31cm} 
        \includegraphics[width=0.185\textwidth]{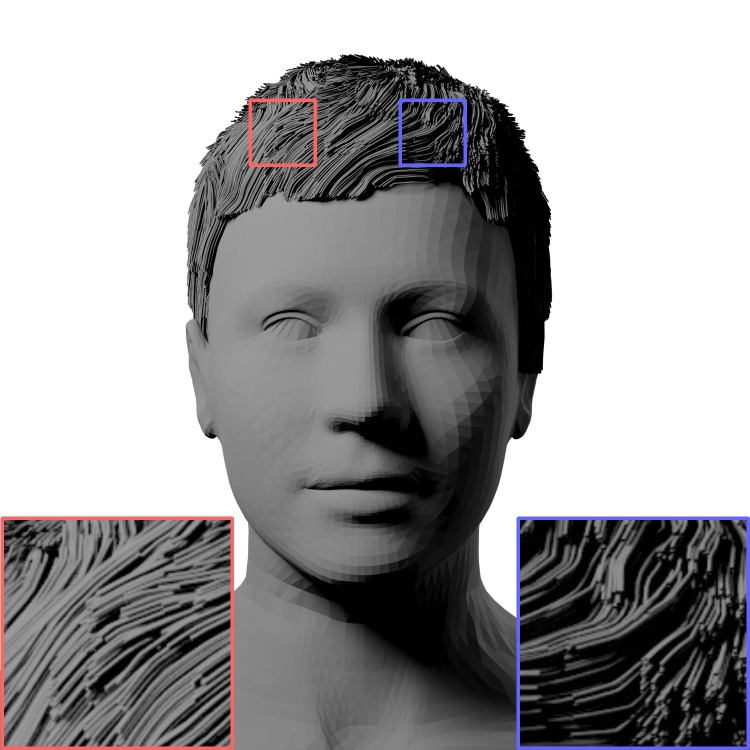} & \hspace{-0.31cm} 
        \includegraphics[width=0.185\textwidth]{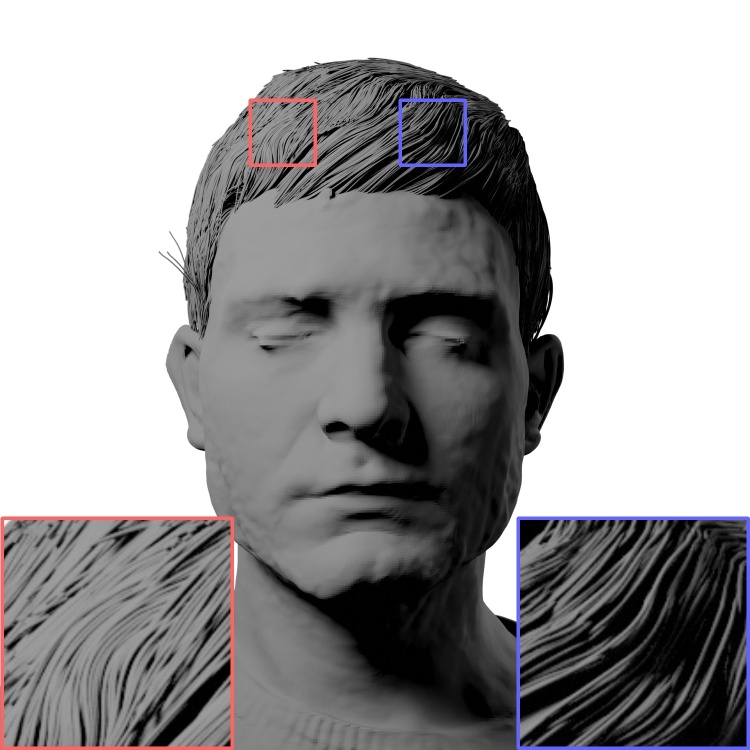} \\ %
        \includegraphics[width=0.185\textwidth]{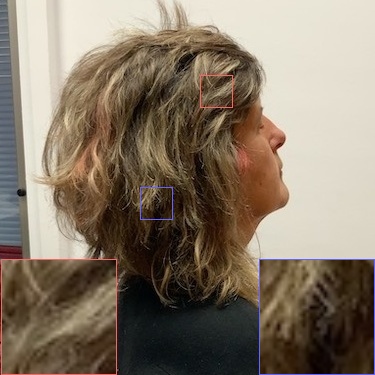} & \hspace{-0.31cm}
       \includegraphics[width=0.185\textwidth]{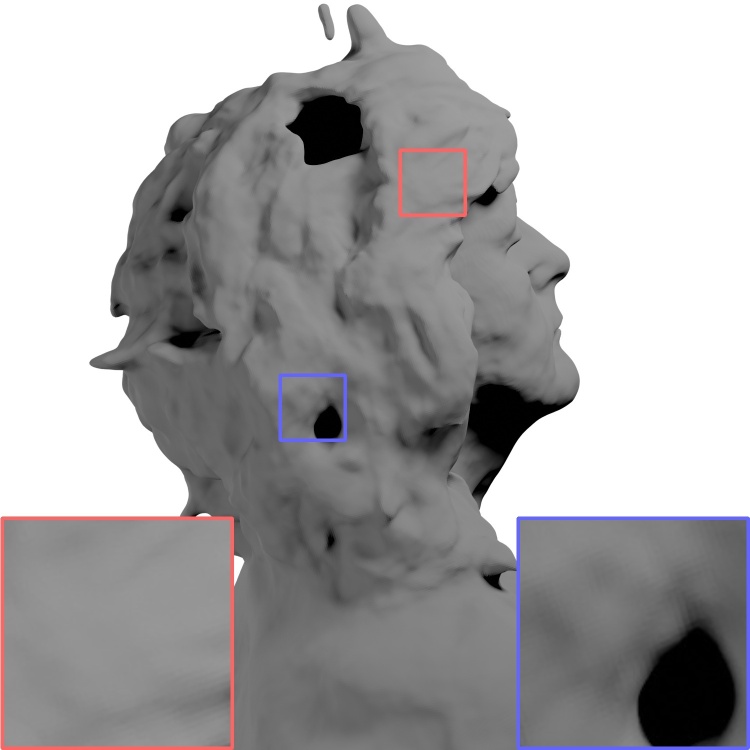} & \hspace{-0.31cm} 
        \includegraphics[width=0.185\textwidth]{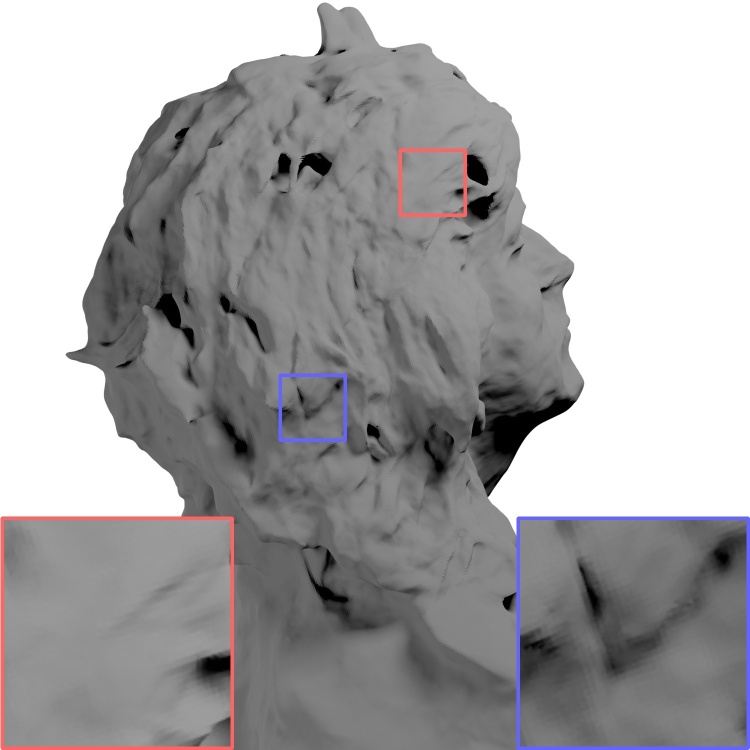} & \hspace{-0.31cm} 
        \includegraphics[width=0.185\textwidth]{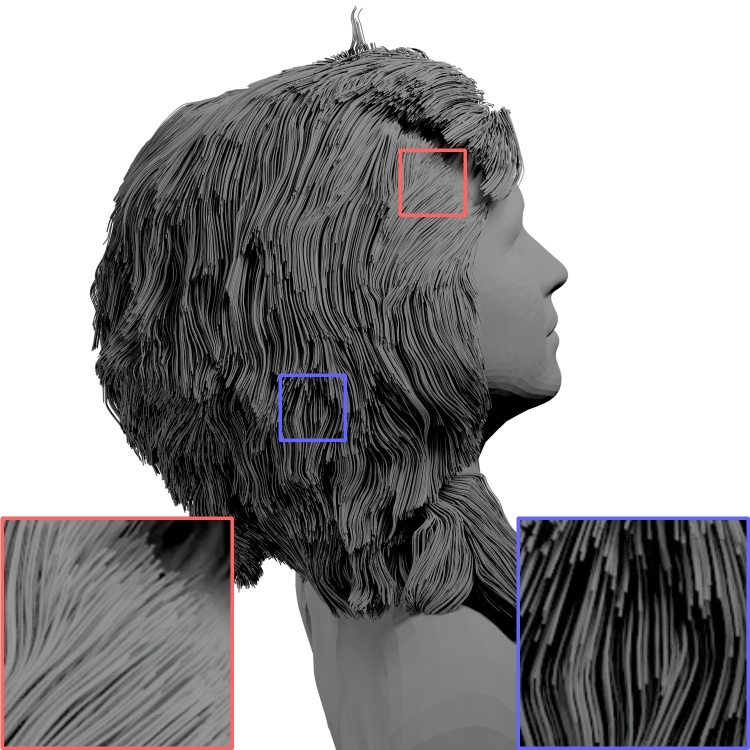} & \hspace{-0.31cm} 
        \includegraphics[width=0.185\textwidth]{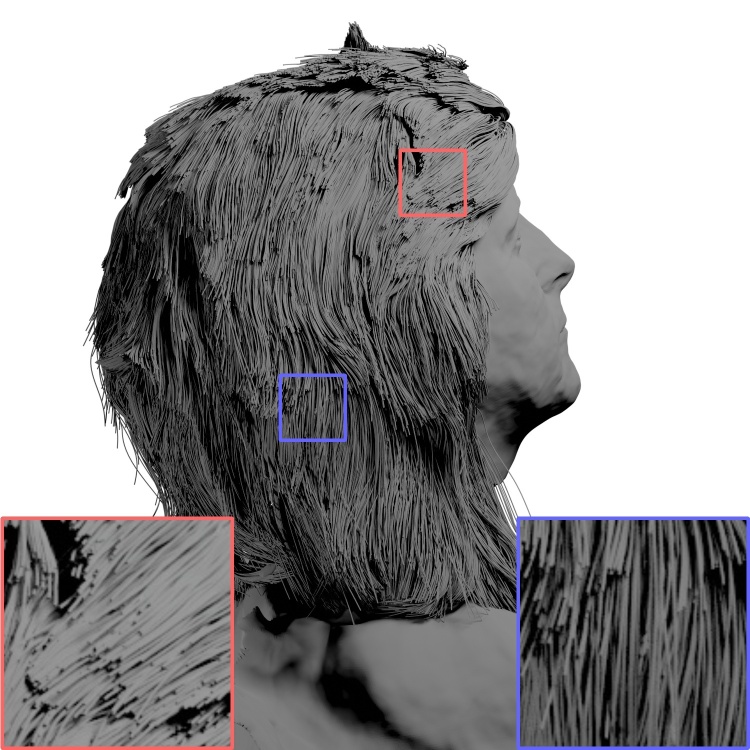} \\ %
        \includegraphics[width=0.185\textwidth]{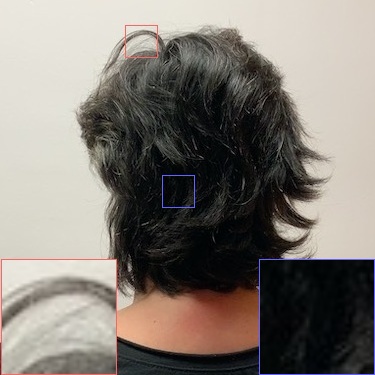} & \hspace{-0.31cm}
        \includegraphics[width=0.185\textwidth]{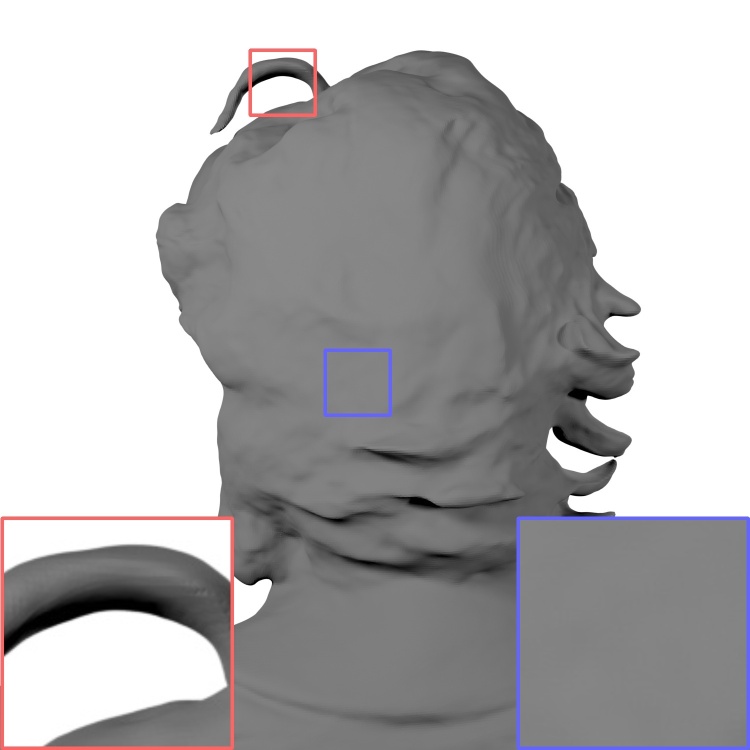} & \hspace{-0.31cm} 
        \includegraphics[width=0.185\textwidth]{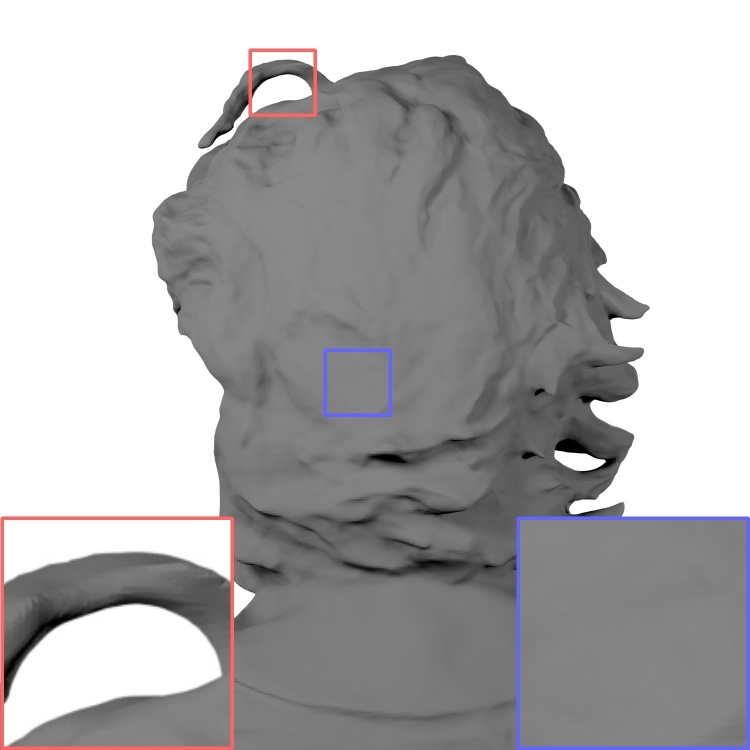} & \hspace{-0.31cm} 
        \includegraphics[width=0.185\textwidth]{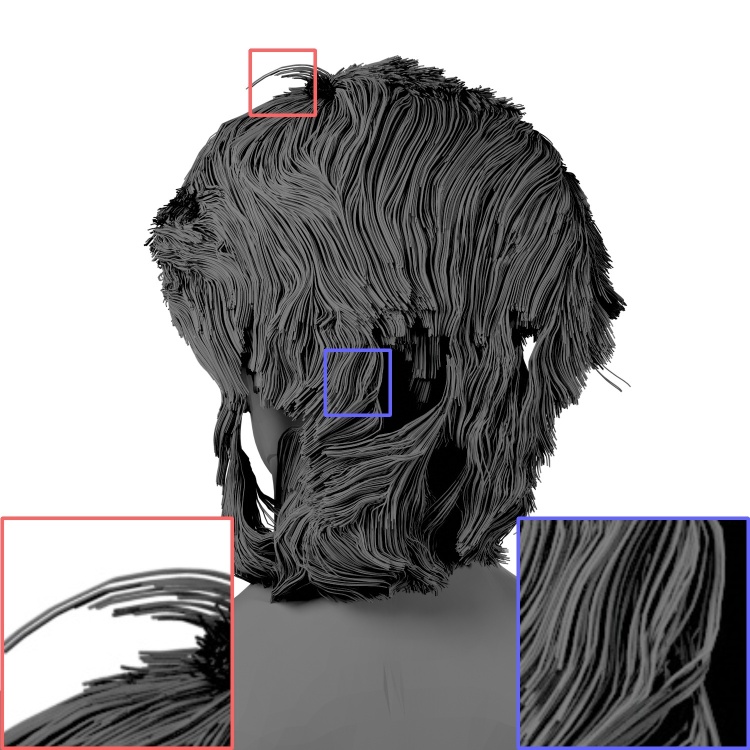} & \hspace{-0.31cm}        
        \includegraphics[width=0.185\textwidth]{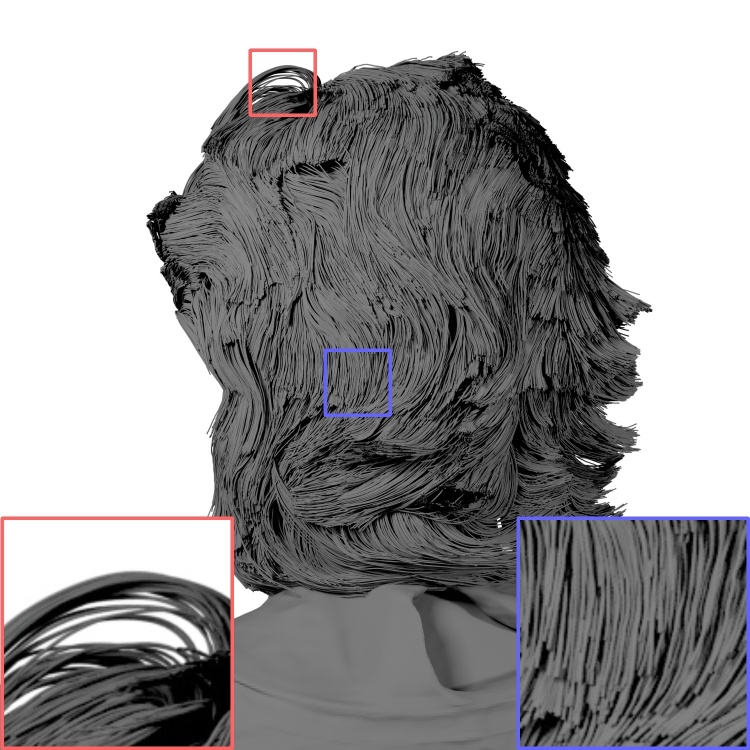} \\ %
        \textbf{Image}  & \hspace{-0.31cm} \textbf{UNISURF}& \hspace{-0.31cm} \textbf{NeuS}  & \hspace{-0.31cm} \textbf{DeepMVSHair} & \hspace{-0.31cm} \textbf{Ours}
    \end{tabular}
    \vspace{-0.35cm}
    \caption{We compare our method with volumetric and strand-based 3D reconstruction systems using a real-world multi-view dataset~\cite{Ramon2021H3DNetFH}. While baseline volumetric approaches~\cite{neus, Oechsle2021ICCV} can only produce coarse hair geometry, our method is able to reconstruct fine details using strands. We also achieve more robust and accurate results than the existing multi-view hair reconstruction methods~\cite{Kuang2022DeepMVSHairDH}. For additional results, please refer to the supplementary materials. Digital zoom-in is recommended.}
    \label{fig:geom_compare}
    \vspace{-0.4cm}
\end{figure*}


%% file: figures/colmap_polina/polina.tex
\begin{figure*}
    \begin{tabular}{cccccc}
        \includegraphics[width=0.1\textwidth]{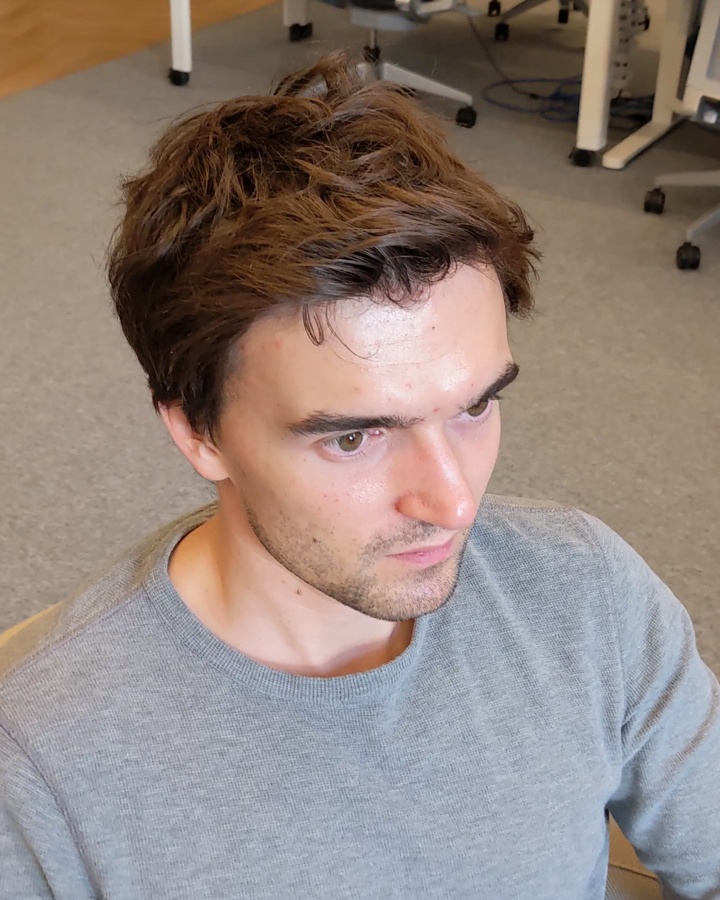} & \hspace{-0.46cm}
        \multirow{2}{*}[0.6in]{\includegraphics[width=0.19\textwidth]{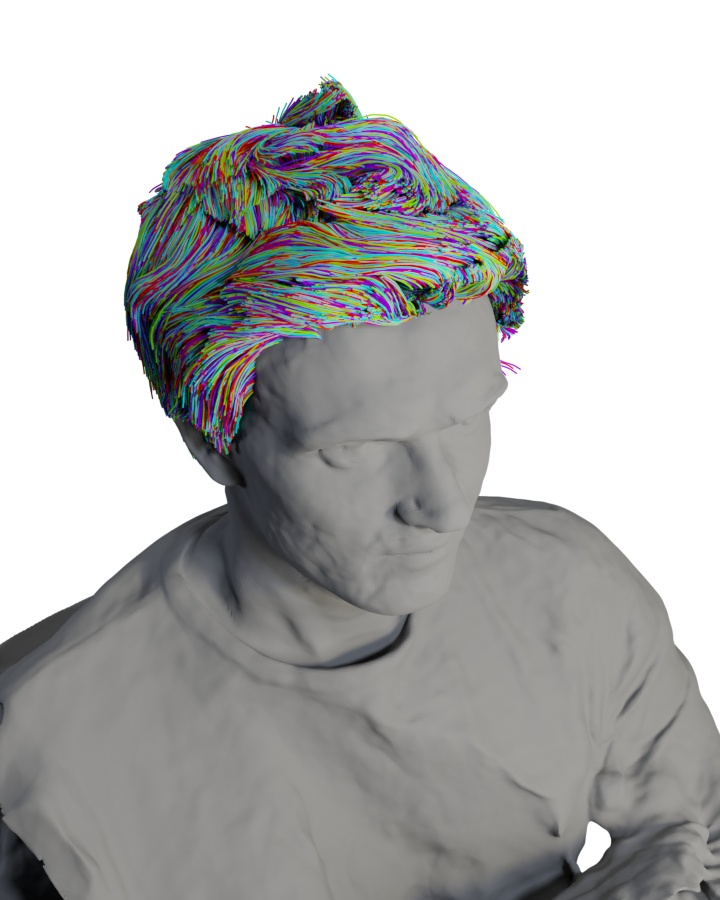}} & \hspace{-0.46cm}
        \multirow{2}{*}[0.6in]{\includegraphics[width=0.19\textwidth]{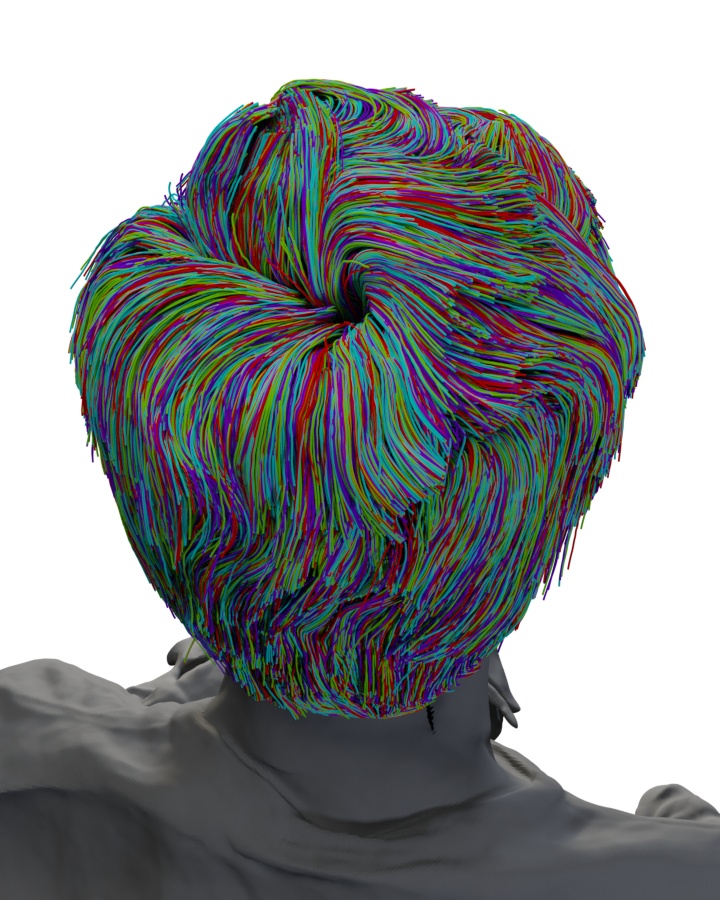}} 
        \hspace{-0.1cm}
        \includegraphics[width=0.1\textwidth]{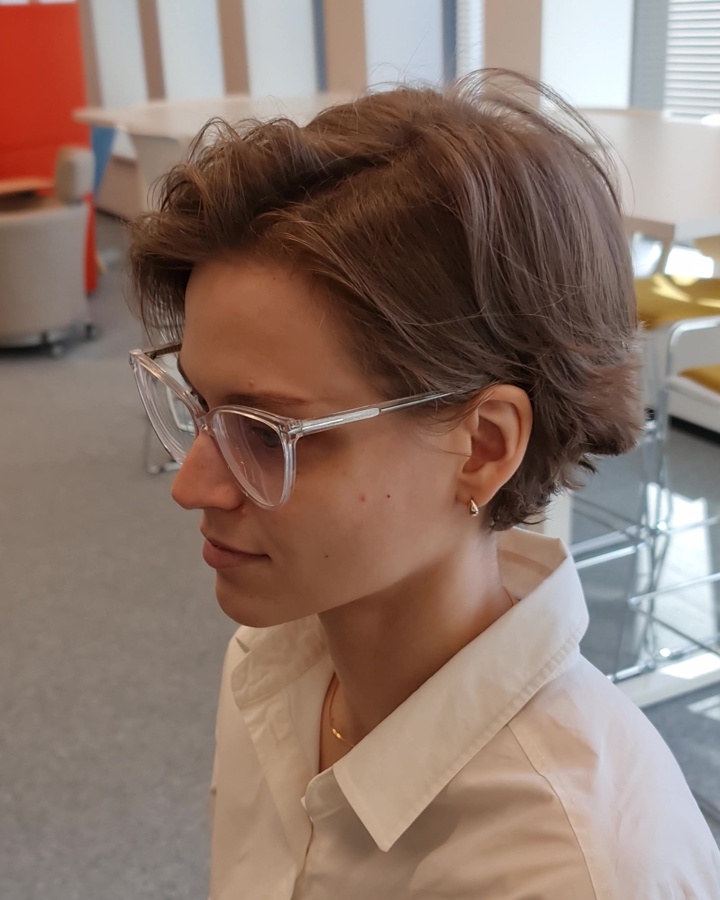} & \hspace{-0.46cm}
        \multirow{2}{*}[0.6in]{\includegraphics[width=0.19\textwidth]{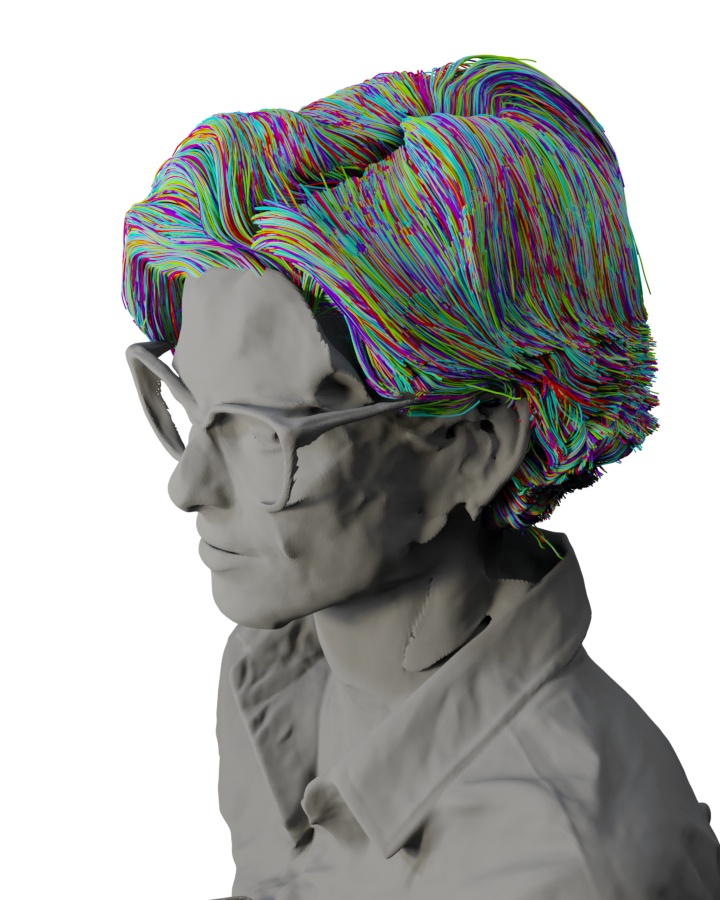}} & \hspace{-0.46cm}
        \multirow{2}{*}[0.6in]{\includegraphics[width=0.19\textwidth]{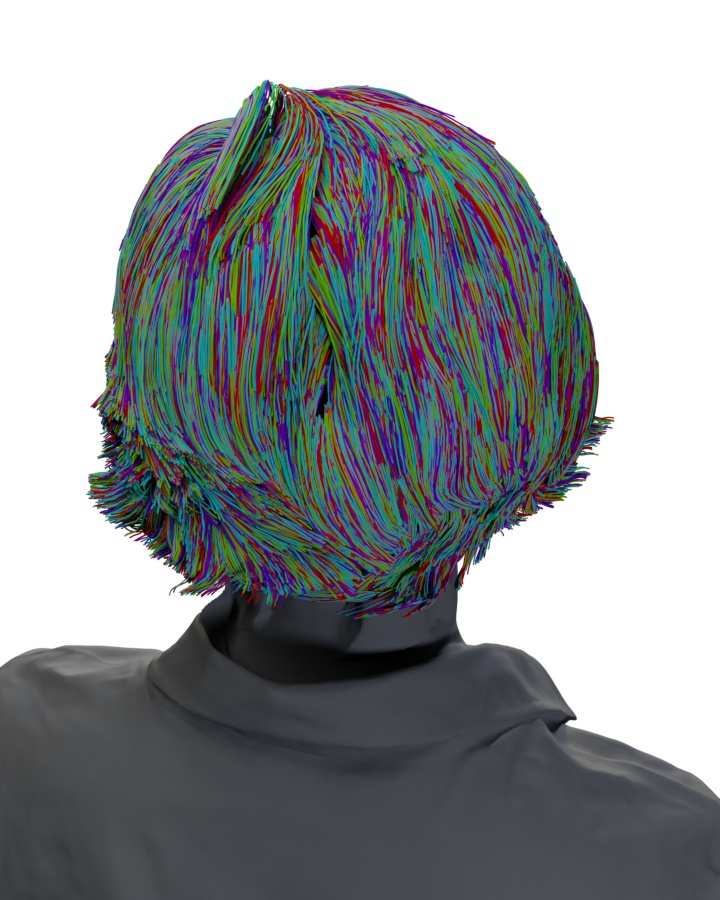}} \\
        \includegraphics[width=0.1\textwidth]{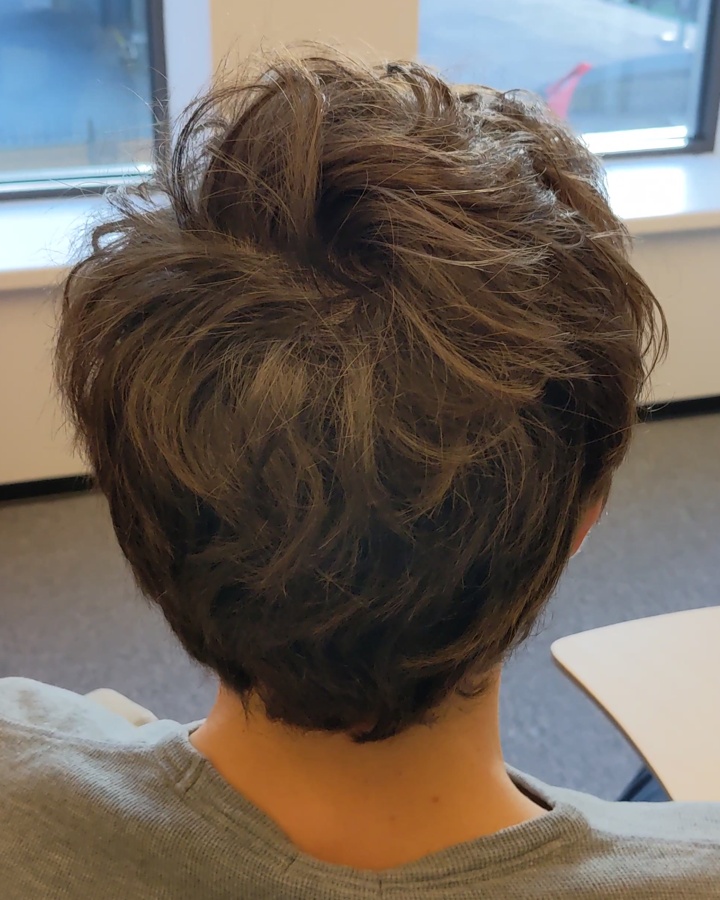} & & &\hspace{-5.5cm}\includegraphics[width=0.1\textwidth]{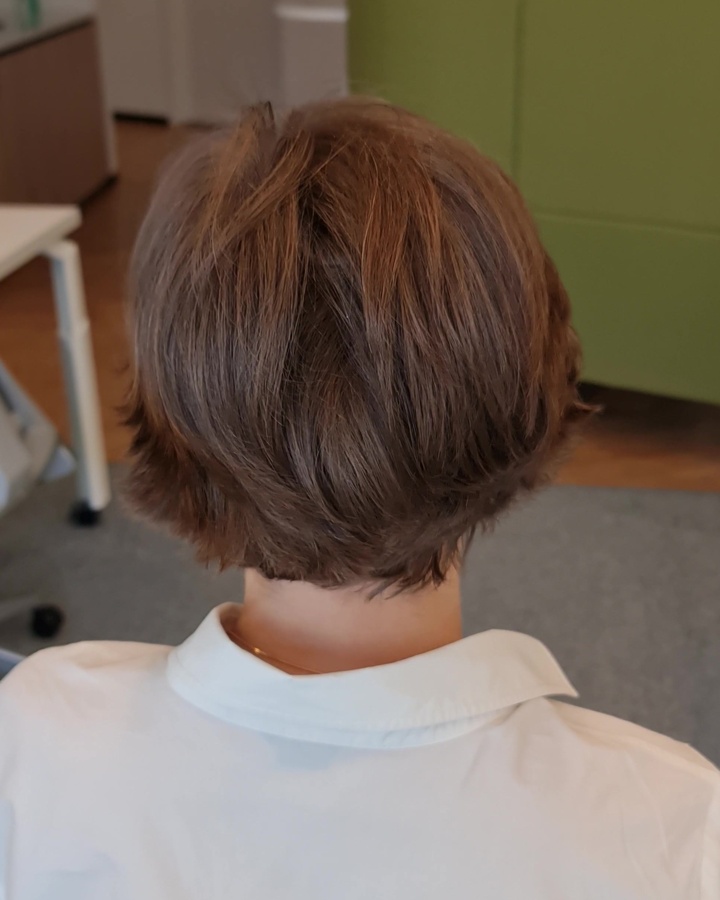}  & &
    \end{tabular}
    \vspace{-0.3cm}
    \caption{Our method can obtain high-fidelity hair reconstructions even from a monocular video. For more results, please refer to the supplementary materials.}
    \vspace{-0.2cm}
    \label{fig:in-the-wild}
\end{figure*}

%% file: figures/table_quant.tex
\begin{figure}
    \centering
    \resizebox{\linewidth}{!}{
    \begin{tabular}{l rrr | rrr | rrr}
        \setlength{\tabcolsep}{0pt}
        & \multicolumn{9}{c}{\textbf{Thresholds: mm} $/$ \textbf{degrees}} \\
        \textbf{Method} & $2 / 20$ & $3 / 30$ & $4 / 40$ & $2 / 20$ & $3 / 30$ & $4 / 40$ & $2 / 20$ & $3 / 30$ & $4 / 40$ \\
        \cline{2-10}
        & \multicolumn{3}{c}{\textbf{Precision}} & \multicolumn{3}{c}{\textbf{Recall}} & \multicolumn{3}{c}{\textbf{F-score}} \\
        \hline
        $\L_\text{geom}$	&	57.3	&	81.9	&	90.4	&	7.8	&	13.8	&	19.8	&	13.7	&	23.5	&	32.5	\\
w/ $\L_\text{render}$~\cite{neuralstrands}	&	58.6	&	82.4	&	91.0	&	8.0	&	13.9	&	21.5	&	14.1	&	23.7	&	34.7	\\
w/ $\L_\text{rgb}$	&	\textbf{60.5}	&	\textbf{83.2}	&	\textbf{91.5}	&	7.6	&	13.8	&	21.0	&	13.5	&	23.7	&	34.1	\\
w/ $\L_\text{mask}$	&	56.5	&	81.5	&	90.4	&	8.7	&	14.7	&	21.0	&	15.0	&	24.9	&	34.1	\\
$\L_\text{fine}$	&	52.9	&	78.1	&	88.4	&	\textbf{9.8}	&	\textbf{17.8}	&	\textbf{26.3}	&	\textbf{16.4}	&	\textbf{28.7}	&	\textbf{40.3}	\\
    \end{tabular}
    }
    \vspace{-0.3cm}
    \captionof{table}{We provide an extensive quantitative evaluation of individual components of our method. Please refer to Section~\ref{sec:ablation} for the discussion.}
    \label{tab:comparison}
\end{figure}

%% file: figures/ablation_difu/ablation_difu.tex
\begin{figure}
    \begin{tabular}{ccc}
        \includegraphics[width=0.31\linewidth]{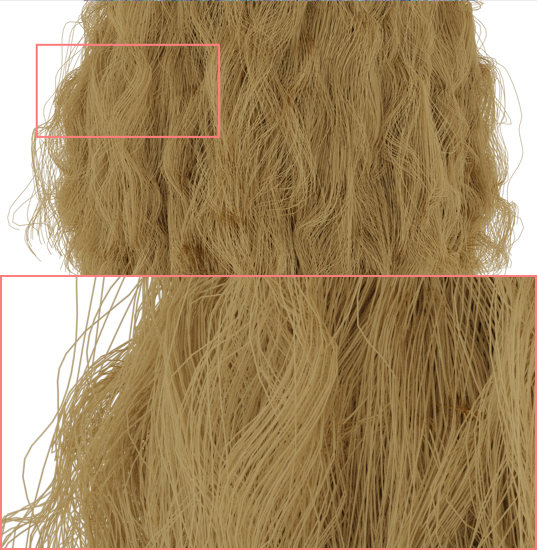}\hspace{-0.37cm}\vspace{-0.13cm} & 
        \includegraphics[width=0.31\linewidth]{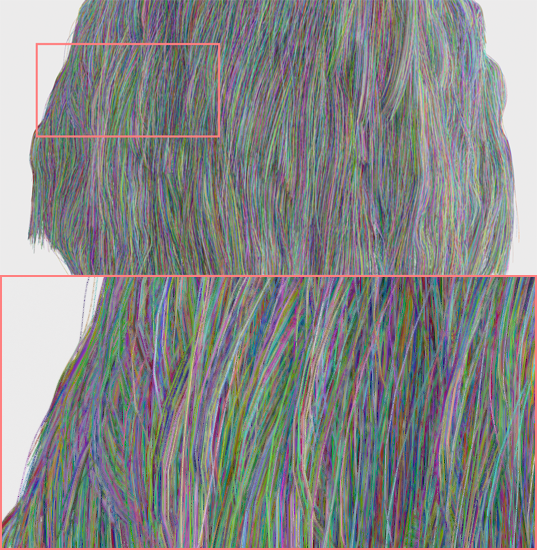}\hspace{-0.37cm} &
        \includegraphics[width=0.31\linewidth]{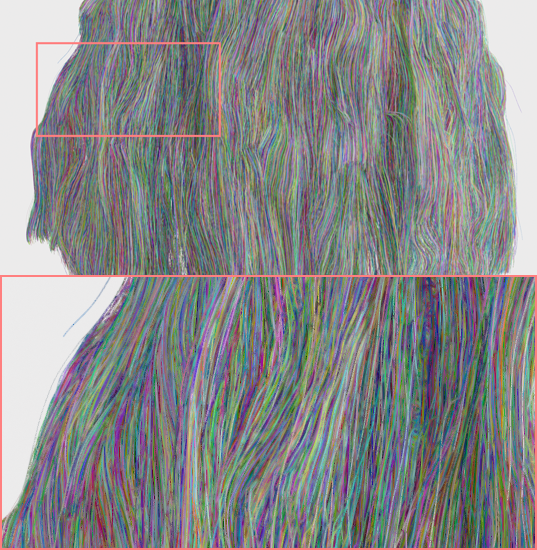}\hspace{-0.37cm} \\
        \textbf{GT}\hspace{-0.37cm}\vspace{-0.11cm} & \textbf{w/o Curv.}~\cite{neuralstrands}\hspace{-0.37cm} & \textbf{Ours}\hspace{-0.37cm}
    \end{tabular}
    \begin{tabular}{cc}
        \includegraphics[width=0.47\linewidth]{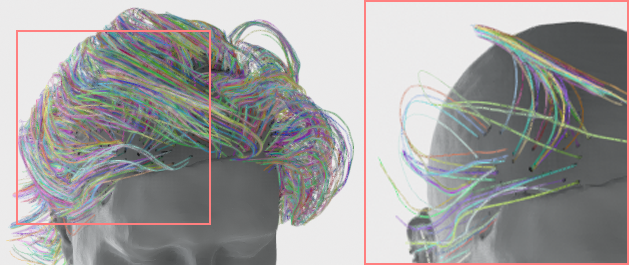}\hspace{-0.37cm}\vspace{-0.1cm} &
        \includegraphics[width=0.47\linewidth]{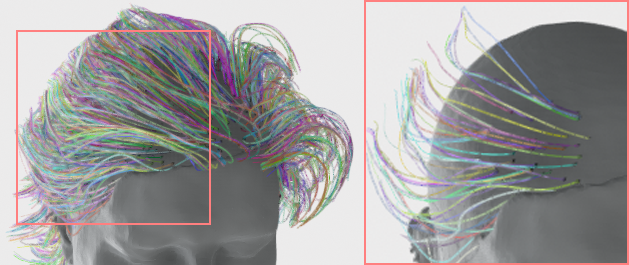}\hspace{-0.37cm} \\
        $\L_\text{geom}$\hspace{-0.37cm}\vspace{-0.3cm} & w/ $\L_\text{prior}$\hspace{-0.37cm}
    \end{tabular}
    \captionof{figure}{Ablation on curvature (top) and diffusion losses (bottom). The incorporation of curvature loss allows us to better model curly strands, while the diffusion tackles the problems with hair growth directions and unrealistic angles (insets show a subset of hairs for clarity).}
    \label{fig:ablation}
    \vspace{-0.3cm}
\end{figure}

%% file: parts/conclusion.tex
\section{Discussion and limitations}

We have presented a method capable of detailed human hair reconstruction from monocular videos with uncontrolled lighting. To achieve that, we employ both volumetric and strand-based hair representations and combine them with differential hair rendering and global hairstyle priors. We demonstrate the efficacy of our approach by conducting extensive qualitative and quantitative evaluations.

The main limitations of our method can be seen in Figures~\ref{fig:in-the-wild}-\ref{fig:ablation}, as well as additional scenes in the supplementary materials. Our system still struggles to represent curly hair and relies on accurate hair and body segmentation masks to produce the reconstructions. In principle, it is possible to address these limitations by extending the dataset for the hairstyle prior training, as well as employing more robust human matting systems.

%% file: parts/acknowledgements.tex
\section*{Acknowledgements}
We thank Samsung ML Platform for providing the computational resources for this work. We also sincerely thank Zhiyi Kuang for aiding us with the DeepMVSHair comparison and Youyi Zheng --- for providing the reconstructions of NeuralHDHair. We also thank David Svitov for his insightful suggestions on diffusion models.

%% file: parts_suppmat/method.tex
\section{Implementation and training details}

\subsection{Datasets preprocessing}
    
We train our hair priors on the publicly available USC-HairSalon~\cite{Hu2015SingleviewHM} synthetic dataset, which consists of 343 hairstyles with up to 10,000 strands aligned with a template bust mesh. Additionally, we match the FLAME head mesh~\cite{FLAME:SiggraphAsia2017} with the template and obtain a UV mapping for the scalp region using Blender\cite{Blender}. We evaluate our method using real-world H3DS~\cite{Ramon2021H3DNetFH} multi-view dataset, monocular videos, and synthetic Cem Yuksel's Hair Models~\cite{Yuksel2009HairM}.

\paragraph{USC-HairSalon.}

To train a parametric prior for individual hair strands, we follow~\cite{neuralstrands} approach to pre-processing. We map each strand into a local tangent-bitangent-normal (TBN) basis using the vertices from the closest face to its root location on the FLAME head mesh. While the normal vector in this basis is calculated using the head mesh and is therefore consistent for nearby strands, to ensure consistency in the other two vectors, we orient the tangent vector in a way that aligns with the $u$ direction of the UV texture coordinates map. The bitangent vector is then defined as a cross-product between the normal and the tangent. The origin of this new coordinate system is the strand's root, so after alignment, each strand originates from $\mathbf{0}$.

We increase the diversity of hair strands following~\cite{neuralstrands} and augment their aligned versions using flipping, stretching and squeezing, and rotations around the normal. On top of that, we also add realistic curliness augmentations and cutting into the mix. We apply the same augmentations besides rotations and flipping to the entire hairstyle for the diffusion-based prior training.

\paragraph{H3DS.} We evaluate our approach using a public subset of a multi-view H3DS dataset~\cite{Ramon2021H3DNetFH}. Each of its scenes has 32 views evenly spaced around the subject. However, since the subject is being moved during the capture because of the non-uniform coverage of the camera setup, their extrinsic parameters are not accurately estimated for some of the scenes. This results in poor performance across all reconstruction methods, and we remove such scenes from the evaluation. We also process these images using human matting~\cite{MODNet} and semantic segmentation~\cite{cdgnet} networks to obtain hair and bust masks. Lastly, we calculate orientation maps using a set of 180 Gabor filters $G_b$ with variances $\sigma_{x}=1.8$ and $\sigma_{y}=2.4$, frequency $\omega = 0.23$, a zero phase offset $\psi$, and a rotation angle $b$, measured in radians. We then obtain an orientation angle $a$ for each pixel: $a = \arg\max_b | G_b |$, and additionally calculate its variance as
\begin{equation}
    \text{Var}[a] =  \sum_{b \in [0, \pi)} \dfrac{\big| G_b \big|}{\sum_o \big| G_o \big|} \cdot \min \{ | a - b |^2,\ | a - b \pm \pi |^2 \} .
\end{equation}

\paragraph{Monocular videos.} We conduct an additional evaluation of our method by training on monocular videos. For this setup, we place subjects in a chair and ask them to remain stationary during the capture session, which lasts around one minute and is produced using a Samsung Note20 Ultra smartphone. Then, we subsample 60 frames from the video, ensuring that they are equally spaced around the subject and have no motion blur. For that, we use image quality assessment networks~\cite{IQA}. We then perform structure-from-motion using COLMAP~\cite{schoenberger2016sfm, schoenberger2016mvs} to obtain initial values for camera intrinsic and extrinsic parameters.   Lastly, we obtain segmentation masks and orientation maps for the training frames following the procedure described previously. We find that additional camera fitting procedure described in~\cite{lin2021barf} launched for first 10,000 iterations during the first stage could improve the quality of reconstructions.

\paragraph{Cem Yuksel's Hair Models.} For the quantitative evaluation, we chose two medium-length hairstyles: curly and straight, from a popular synthetic dataset~\cite{Yuksel2009HairM}. We used a separate dataset from USC-HairSalon for evaluation to avoid bias in the calculated metrics. By using Blender~\cite{Blender}, we generate 70 views with a resolution of $2048 \times 2048$ for each scene for training, which includes both RGB and segmentation masks. We then calculate the orientation maps using the same procedure based on Gabor filters.

\subsection{Hair prior training}

Each hair strand in the synthetic dataset is represented as a set of L points: $\mathbf{S} = \{\mathbf{p}_l \}_{l=1}^L$, while each hairstyle sample consists of $M$ strands: $\{\mathbf{S}_i \}_{i=1}^M$. The number of points per each strand is $L = 100$ across the whole dataset, while the number of strands $M$ varies from sample to sample. 

\paragraph{Hair strand parametric model.}

As stated previously, we map the individual hair strands into a TBN basis and augment them. Then, we encode the aligned 3D points using an encoder $\E$, a one-dimensional ResNet-50~\cite{Resnet}, into the mean $\zgeom_\mu \in \mathbb{R}^{64}$ and sigma $\zgeom_\sigma \in \mathbb{R}^{64}$. We then perform a reparameterization trick $\zgeom = \zgeom_\mu + \zgeom_\sigma \cdot \epsilon$ and decode the resulting latent vector into a strand via a decoder $\G$. Instead of predicting individual points, we follow~\cite{neuralODE} and predict offsets $\d_i^l = \p_i^{l+1} - \p_i^l$. We use a modulated SIREN~\cite{SIREN} network consisting of two MLPs with 8 layers and 256 channels for the decoder architecture following~\cite{neuralstrands}. We use this module to individually decode each offset on the strand given its index $l$ and the latent vector $\zgeom$ as inputs. The index $l$ is normalized and used in the periodic activation functions. The resulting points on the strands are obtained by accumulating the offsets:
\begin{equation}
    \p_i^l = \sum_{j=1}^{l-1} \d_i^j,\quad l = 2 \dots L,
\end{equation}
and $\p_i^1 = \mathbf{0}$ due to alignment. The training of $\E$ and $\D$ proceeds using the training objective described in the main paper. For optimization we use Adam~\cite{Adam} with cosine annealing of learning rate from $10^{-4}$ to $10^{-5}$ and weights  $\lambda_{d}=0.05$, $\lambda_{c}=1$, $\lambda_\text{KL} = 10^{-4}$. After training, the weights of these networks remain frozen, and $\zgeom_\mu$ is used as $\zgeom$.

\paragraph{Hairstyle diffusion model.} We prepare a training sample for the diffusion model by first mapping a hairstyle $\{ \S_i \}_{i=1}^M$ with random strand origins $\p_i^1$ into a hairstyle whose origins span a uniform grid on the FLAME scalp texture map. For that, we use nearest neighbors interpolation. We use the texture with resolution ${256 \times 256}$ for both the hairstyle prior training and fine-tuning. After that, we apply the common augmentations for the entire hairstyle which were described in the previous section in the basis calculated as an average over the basis of its strands components and map them into a latent texture $\Zgeom = \{ \zgeom_{ij} \}_{i,j=1,1}^{256,256}$ using $\E$. Lastly, we subsample this texture into a low-resolution version $\Zgeom_\text{LR} \in \mathbb{R}^{32 \times 32}$ using the random integer offsets $s_i \in [0, 7]$ and $s_j \in [0, 7]$. The low-resolution texture can then be obtained as follows:
\begin{equation}
\begin{aligned}
    \Zgeom_\text{LR} = \{ \z_{ij} \mid i & = s_i + 8q, \\
    j & = s_j + 8r,\ q,r = 0 \dots 31 \}. 
\end{aligned}
\end{equation}
Such subsampling allows us to generate exactly $8^2 = 64$ different training samples per hairstyle, boosting the diversity of the dataset and speeding up the training of the prior.

We then follow EDM~\cite{Karras2022ElucidatingTD} training pipeline and sample $\epsilon$ from a standard normal distribution and a noise level $\sigma$ from a log-normal distribution with the mean $-1.2$ and sigma $1.2$. We obtain a noised texture $\x$ as:
\begin{equation}
    \x = \Zgeom_\text{LR} + \sigma \cdot \epsilon.
\end{equation}
Then, for training we use an equivalent simplified version of $\L_\text{diff}$:
\begin{equation}
\begin{aligned} 
    \mathcal{L}_\text{diff} =
    \mathbb{E}_{\y, \sigma, \epsilon}\, \Big[\ \Big\| & \F \big( c_\text{in}(\sigma) \cdot \x, c_\text{noise}(\sigma) \big) - \\
    & \frac{1}{c_\text{out}(\sigma)} \big( \y - c_\text{skip}(\sigma) \cdot \x \big) \Big\|_{2}^{2}\ \Big].
\end{aligned}
\end{equation}
For derivations, please refer to~\cite{Karras2022ElucidatingTD}. In Figure~\ref{fig:diffusion_generation_suppmat}, we show the samples of a pre-trained diffusion model. These hairstyles look sparse, as they only contain $32^2 = 1024$ strands.

For diffusion model we use UNet architecture from EDM~\cite{Karras2022ElucidatingTD} and optimize it using AdamW~\cite{AdamW} with learning rate $10^{-4}$,  $\beta=[0.95, 0.999]$, $\epsilon = 10^{-6}$, and weight decay $10^{-3}$. For scheduling, we use an inverse decay learning rate schedule with  inverse multiplicative factor = 20000, factor = 1, and warmup = 0.99. All training on synthetic dataset took 2 days on a single NVIDIA RTX 4090.

After training, the diffusion network $\F$ has its weights frozen. 

\input{figures_suppmat/diffusion_generation/diffusion_generation}

\subsection{Coarse volumetric reconstruction}

\paragraph{FLAME fitting.}
For each scene, we fit a FLAME head mesh using keypoint-based objectives. We detect the ground truth keypoints for the face using an OpenPose~\cite{OpenPose1, OpenPose2, OpenPose3, OpenPose4} and Face Alignment~\cite{bulat2017far} detectors and filter out the frames where the face is not visible. 
Then, we optimize w.r.t. the FLAME shape and pose parameters by minimizing the difference between the projected head mesh keypoints and the detected ones. First, we optimize global rotation, translation, and scale parameters and then additionally fit shape starting from PIXIE~\cite{PIXIE:3DV:2021} initialization and turning on the shape regularization.  
We use a pipeline similar to DECA ~\cite{DECA} for visible keypoints projection and L-BFGS ~\cite{LBFGS} optimizer with a learning rate set to $0.5$.

\paragraph{Volumetric reconstruction.}
To calculate $\ahair_i$ and $\abust_i$, we follow the approach described in NeuS~\cite{neus} and convert the SDF values $\fhair(\x_i)$ and $\fbust(\x_i)$ into opacities for a given ray $\v_i$. To obtain a set of points $\{\x_i\}_{i=1}^N$ used in ray marching, we apply the iterative importance sampling algorithm from~\cite{Mildenhall2020NeRFRS} using blended opacities $\alpha$.

We then use the FLAME mesh to provide additional training signals for the occluded bust regions, which cannot be correctly reconstructed by simply minimizing the difference between the rendered and ground truth colors and silhouettes. Following prior works for fitting SDFs to mesh-based geometry~\cite{Atzmon2019SALSA, Gropp2020ImplicitGR, Sitzmann2020ImplicitNR}, we regularize the implicit SDF to vanish near the surface of the mesh and match its gradients $\nabla_\x \fbust(\x)$ to the surface normals $\n (\x)$ of the closest point on the FLAME mesh. Additionally, we penalize the non-zero hair occupancies $\ahair_i$ inside the bust mesh. 

We calculate this loss by reusing the points $\x_i$ sampled during ray marching to make the training process more efficient. We split these points into two groups: those who lie inside the volume $\Omega_\text{head}$ bounded by the mesh, and the ones that lie outside: $\Omega_\text{out} = \Omega \setminus \Omega_\text{head}$. We additionally sample a set of points $\x^\text{head}_i$ on the surface of the head mesh, denoted as $\Omega_0$, to evaluate surface-based constraints. The final loss is denoted as $\L_\text{head}$:
\begin{equation}
\begin{aligned}
    \L_\text{head} = \sum_{\x^\text{head}_i \in \, \Omega_0} & \big| \fbust(\x^\text{head}_i) \big| \, + \\ & 0.1\cdot \big( 1 - \nabla_{\x^\text{head}_i} \fbust(\x^\text{head}_i) \cdot \n (\x^\text{head}_i) \big) \\ +
   \sum_{\x_i \in \Omega_\text{out}} \ & 0.1 \cdot  \exp \big( -\gamma \cdot \big| f_\text{bust}(\x_i) \big| \big) \\ +
    \sum_{\x_i \in \Omega_\text{head}} & \big| \ahair_i \big| \, ,
\end{aligned}\label{eq:sdf_head}
\end{equation}
where $\cdot$ denotes a dot product, and $\gamma \gg 1$ is a constant.

To calculate an orientation loss, we follow~\cite{Nam2019StrandAccurateMH} and use Plucker line coordinates~\cite{Wrobel2001MultipleVG} to project the orientation field $\beta$ at point $\x_s$ along the ray $\v$ into the camera $\mathcal{P}$, the projected 2D direction in the camera coordinates is denoted as $\mathbf{L} ( \x_s, \beta(\x_s), \mathcal{P})$. Then, we measure the angle  $\hat{a}_{\v}$ between the predicted direction and the camera $y$-axis, which is module $\pi$, i.e.\ in the range $[0, \pi)$. The direction loss between this predicted angle and the ground-truth orientation $a_\v$ with its variance $\text{Var}[a_\v]$ in the hair region are measured as follows:
\begin{equation}
    \L_\text{dir} = \sum_{\v} \frac{\mhair(\v)}{ \text{Var}^2 [a_\v] } \min \big\{ | a_\v - \hat{a}_\v |,\ | a_\v - \hat{a}_\v \pm \pi | \big\},
\end{equation}
where $\mhair(\v)$ denotes a hair mask value at the rendered pixel, corresponding to the ray $\v$, and the sum is across all rays in the batch.

\paragraph{Network architecture.}
We use a similar network architecture as NeuS~\cite{neus}, which consists of three MLPs to encode SDF $\fhair$ for hair geometry, SDF for head $\fbust$, and scene color, respectively. 
The geometry networks have 8 hidden layers with a hidden size of 256, Softplus with $\beta=100$ as the activation function, and a skip connection from the input to the fourth hidden layer.

The hair geometry network first transforms the input points via positional encoding with 8 harmonics and then passes them through the MLP to predict their SDF, $\fhair\in\mathbb{R}$, features $\lhair\in \mathbb{R}^{256}$, and orientations $\beta\in\mathbb{R}^{3}$. The activation of the orientation head is the Tanh function. 
As the rest of the bust has lower frequency details, the input points of the head geometry network are positionally encoded with 6 harmonics. Similarly, the network predicts the bust SDF $\fbust\in\mathbb{R}$ and feature vectors $\lbust\in \mathbb{R}^{256}$.

We have a joint color network, which is modeled by an MLP with 4 linear layers with a hidden size of 256. As input it takes the spatial location $x_{i}$, the view direction $\v$, the normal vector of SDF, $\n = \nabla{f(x_{i})}$, and a 256-dimensional feature vector, $l$. To combine the feature vectors and normals of the hair, $\lhair, \nabla{\fhair(x_{i})}$, with the bust, $\lbust, \nabla{\fbust(x_{i})}$, we first calculate the blending weight $w_{i}$ for each point as follows:

\begin{equation}
    w_i = \frac{\abust_i}{\abust_i +\ahair_i+\varepsilon},
\end{equation}
where $\abust_i,\ahair_i$ - are individual opacities of bust and hair correspondingly and $\varepsilon=10^{-5}$ is used for numerical stability.
Then we blend the features and the normals accordingly:

\begin{equation}
    l = w_{i}\cdot \lbust +(1-w_{i})\cdot \lhair
\end{equation}

\begin{equation}
    \n = w_{i}\cdot \nabla{\fbust(x_{i})}+(1-w_{i})\cdot \nabla{\fhair(x_{i})}
\end{equation}

We train volumetric reconstruction using Adam~\cite{Adam} optimizer with learning rate equal to $5\cdot 10^{-4}$ and weights: $\lambda_\text{color}=1$, $\lambda_\text{mask} = 0.1$, $\lambda_\text{reg}= 0.1$, $\lambda_\text{head} = 0.1$, $\lambda_\text{dir}= 0.1$ for 300,000 iterations.

\subsection{Fine strand-based reconstruction}

For fine strand-based optimization we use Adam~\cite{Adam} with learning rate set to $10^{-3}$  and MultiStep annealing with $\gamma=0.5$ and milestones = $[4\cdot10^{4}, 6\cdot10^{4}, 8\cdot10^{4}]$. Also, we use the following weights of losses: $\lambda_\text{chm.}=1.$, $\lambda_\text{orient}=0.01$, $\lambda_\text{prior}=10^{-3}$, $\lambda_\text{render}=10^{-3}$, $\lambda_\text{mask}=0.01$.

\paragraph{Texture parametrization.} Our geometry $\Zgeom$ and appearance texture have resolution $256\times 256$ with number of channels 64 and 16 correspondingly. They are both parameterized using a UNet, similar to deep image prior~\cite{deepimageprior}. We share network parameters between these textures and predict them from a constant grid of UV coordinates. We pre-process them using a positional encoding~\cite{Mildenhall2020NeRFRS} before feeding into the network, which consists of 6 sine and cosine functions.

\paragraph{Visible surface extraction.} We obtain the visible hair surface $\mathcal{S}$ from a hair SDF $\fhair$ and a bust SDF $\fbust$ to use it in orientation $\L_\text{orient}$ and chamfer $\L_\text{chm.}$ losses. To obtain it, we first extract zero-level iso-surfaces from both implicit functions using Marching Cubes~\cite{Lewiner2003EfficientIO}. Then, we render both hair and bust meshes using a set of cameras from the chosen dataset. Due to limited top views in H3DS~\cite{Ramon2021H3DNetFH} we additionally consider top cameras to prevent the appearance of big holes in hair SDF geometry. Finally, we select all the hair faces that are visible from at least one view and use the resulting mesh as $\mathcal{S}$ in all losses.


    

\paragraph{Soft rendering.}

\begin{figure}
    \includegraphics[scale=0.3]{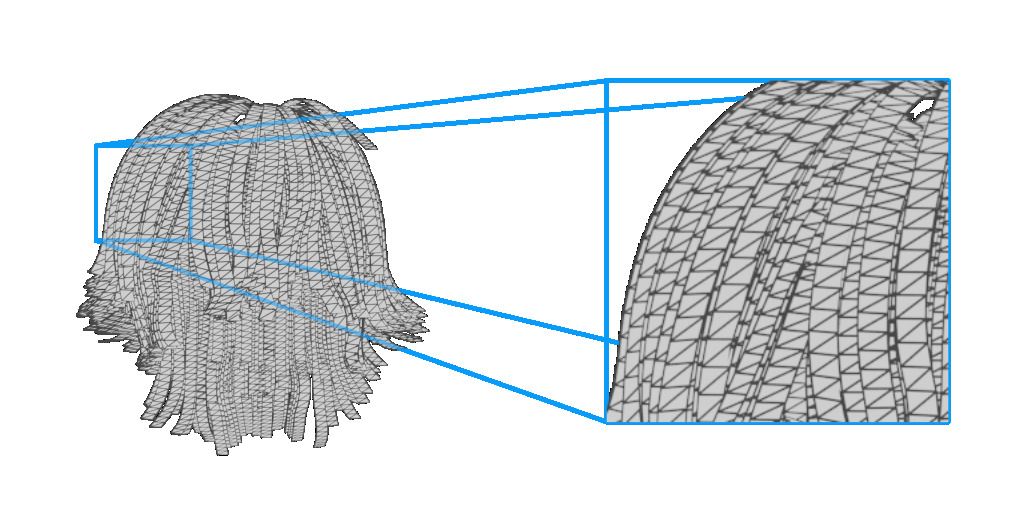}
    \caption{Hair quads produced by the view-aware generation: geometry is built in a way that most of the quads are facing the camera plane.}
    \label{fig:hair_quads}
\end{figure}

For differentiable soft rasterization, we use Pytorch3D~\cite{Pytorch3d} framework.  Our full soft rasterization pipeline consists of three steps. First, we generate quad geometry for each strand in a hairstyle and orient these polygonal quads so that most of the produced faces are aligned with the camera plane, see Fig. \ref{fig:hair_quads}. It requires calculating Frenet–Serret frame \cite{Crane:2013:DGP} for each point of a strand and then generating additional vertices while considering only $XY$ coordinates in camera space:
\begin{equation}
    \mathbf{S}_\text{gen.} = \mathbf{S} \pm [\mathbf{N}_{XY}, 0],
\end{equation}
where $\mathbf{S}$ -- hair strand represented by 3D vertices, $\mathbf{N}_{XY}$ -- normal vector to a strand projection onto a camera plane, by $[\cdot, \cdot ]$ we denote a simple concatenation. Such view-aware generation prevents quads from being oriented orthogonal to the view plane, which effectively increases the number of samples as if they were oriented randomly.

Secondly, we rasterize obtained quads for hair within head to obtain z-buffer with the nearest faces to each pixel. Finally, we blend it using sigmoid probability map. For rasterization we use blur radius = $10^{-4}$, faces per pixel = 16 and image size = 512 due to memory restrictions and for blending: $\sigma=10^{-5}$ and $\gamma=10^{-5}$.  


The input of our rendering UNet network consists of soft rasterized appearance descriptors $\in \mathbb{R}^{16}$ concatenated with hard rasterized orientations $\hat{a}$ transformed to $\mathbb{R}^3$ using sine and cosine functions. For the cosine, we additionally split it into two channels, corresponding to the positive and negative components, and take their absolute value. This way, we ensure that all channels are normalized in $[0, 1]$. While the rasterized appearance features are the same for the whole strand, orientations are different for each point. They contribute to the ability of neural rendering to model the view-dependent changes in hair color since the projected orientations contain information about both hair strand local growth direction and camera view direction. For rendering UNet, we use architecture similar to~\cite{Rakhimov_2022_CVPR}. 

%% file: figures_suppmat/diffusion_generation/diffusion_generation.tex
\begin{figure*}
    \begin{tabular}{ccccc}
        \includegraphics[width=0.185\textwidth]{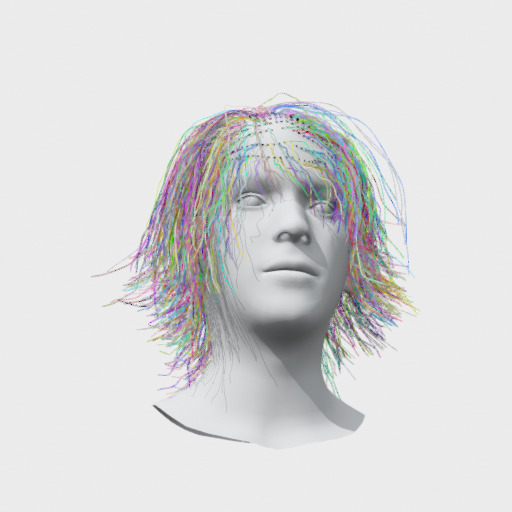} & \hspace{-0.31cm}
        \includegraphics[width=0.185\textwidth]{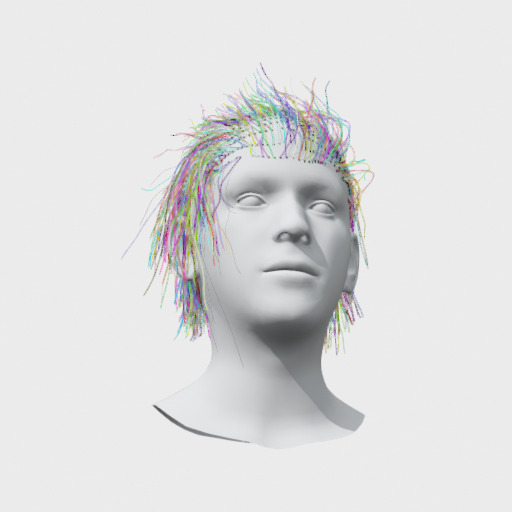} & \hspace{-0.31cm} 
        \includegraphics[width=0.185\textwidth]{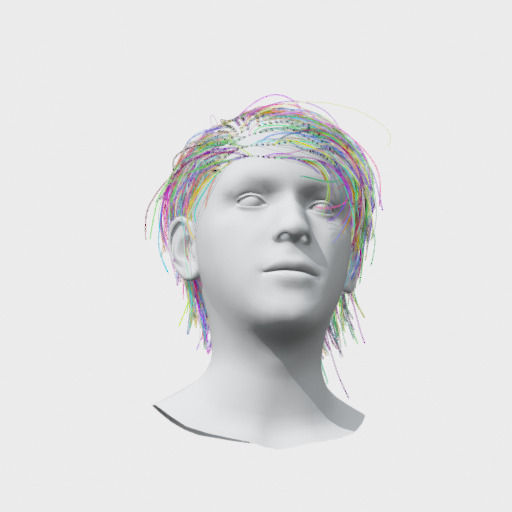} & \hspace{-0.31cm} 
        \includegraphics[width=0.185\textwidth]{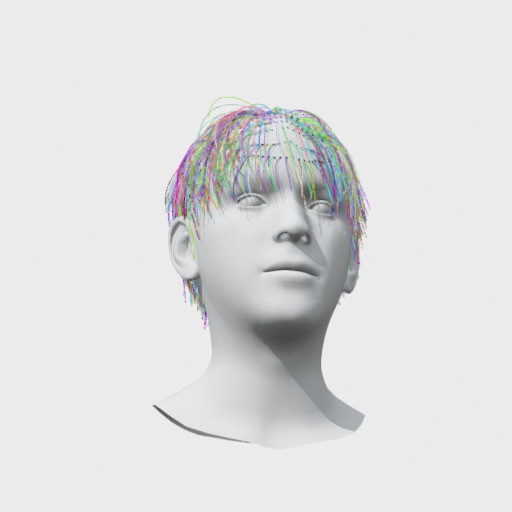} & \hspace{-0.31cm} 
        \includegraphics[width=0.185\textwidth]{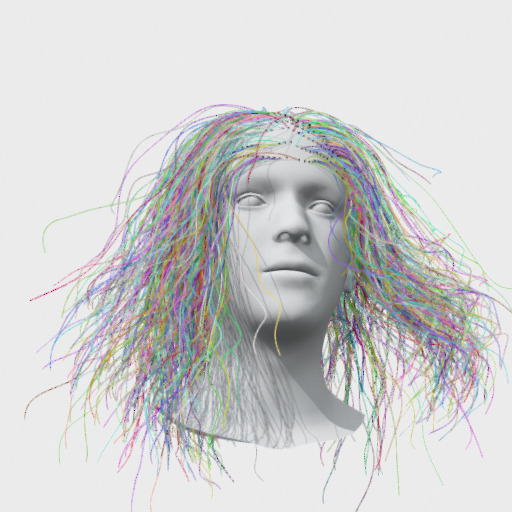} 
    \end{tabular}
    \caption{Random hairstyles produced by a pre-trained diffusion model. Each sample consists of $1024$ individual strands}
    \label{fig:diffusion_generation_suppmat}
\end{figure*}

%% file: parts_suppmat/experiments.tex
\section{Additional Ablations and Results}

\input{figures_suppmat/table_ablation}
\input{figures_suppmat/table_hyperparameters}

\input{figures_suppmat/comp_NeuralHDAvatar/comp_neuralhdavatar}

\paragraph{Evaluation protocol.}

For metrics evaluation we sample 50,000 strands, the same number as in ground-truth hairstyle. Furthermore, we linearly interpolate the number of points one each strand in ground-truth hairstyle to 100 to prevent biased metrics.

\paragraph{Real-world evaluation.}

We show an extended comparison of our approach with reconstruction methods for different viewing angles (see Figure~\ref{fig:geom_compare_suppmat}) on H3DS~\cite{Ramon2021H3DNetFH}. Furthermore, we visualize the reconstructions for additional scenes in Figure~\ref{fig:geom_compare_suppmat_additional}. 
Our approach can handle both long and short hairstyles, proving its versatility in achieving realistic reconstructions in different scenarios. Finally, we provide results of our system obtained on the same twelve views as used in DeepMVSHair~\cite{Kuang2022DeepMVSHairDH} (see Figure~\ref{fig:geom_compare_suppmat_12_views}).

We also provide an extended comparison with the one-shot reconstruction method NeuralHDHair\cite{Wu2022NeuralHDHairAH}, see Figure~\ref{fig:neural_hd_hair_suppmat}. The main advantage of our method is the higher fidelity of hair reconstructions, which are obtained jointly with the personalized bust models. 

Lastly, we include additional monocular video reconstruction examples. In total, we present the results for five scenes with various hairstyles: long, short, and curly, see Figure~ \ref{fig:colmap_scene_jenya_suppmat}, \ref{fig:colmap_scene_ksyusha_suppmat} and \ref{fig:colmap_scene_nastya_suppmat}.
\paragraph{Ablation on losses.}\label{sec:ablation_suppmat}

We provide an extended ablation study in Table~\ref{tab:comparison_suppmat} and Figure~\ref{fig:ablation_geom_supplem}. It contains separate results for curly and straight synthetic hair, which extend quantitative metrics presented in the main paper. We additionally include an ablation study for the individual components of the geometry loss $\L_\text{geom}$ in the upper section and importance of $\L_\text{rgb}$ in the bottom.
Our full method, $\L_\text{fine}$, achieves the best performance in terms of both Recall and aggregated F-score for both scenes. 
All terms in $\L_\text{geom}$ clearly contribute to the overall performance.
Without $\L_\text{chm}$ the generated hairstyle does not cover the whole hair volume. Since we use only a one-way chamfer, not two-way, to attract strands to the outer surface, discarding $\L_\text{vol}$ leads to unrealistic strands outside the hair region, see Figure~\ref{fig:ablation_geom_supplem}. Without $\L_\text{orient}$ we obtain the same hairstyle coverage, but strands have random orientations on the surface which significantly decreases the realism. Furthermore,  from Table~\ref{tab:comparison_suppmat}  you could see the decrease in performance for both scenes without using $\L_\text{rgb}$. 

We also provide an extended comparison with Neural Strands~\cite{neuralstrands}. We have re-implemented a strand generator network, a differentiable rasterizer, neural hair rendering, texture parametrization, and a training procedure.  Due to the unavailability of the ground-truth 3D strand segments and hair growth directions, we replace their geometry loss with our $\mathcal{L}_\text{geom}$. We refer to the resulting method as Neural Strands$^*$
and present the results in Tab.~\ref{tab:comparison_suppmat} (last row) and Fig.~\ref{fig:neural_strands_comp}. Our method outperforms Neural Strands~\cite{neuralstrands} both quantitatively (across most metrics) and qualitatively. 

\input{figures_suppmat/metrics_visual}

\paragraph{Hyperparameters study.}\label{sec:ablation_suppmat}

We conduct an ablation study on important hyperparameters used in the soft rendering part in order to make sure that the chosen ones are optimal. Results on curly synthetic hair scene are provided in Table~\ref{tab:comparison_hyperparameters}. We varied different number of faces per pixel and images used at each iteration in soft rasterization. In our final model by default, we use faces per pixel = 16 and batch size = 1. From the table, we could see that neither increase nor decrease of these hyperparameters helps to improve results compared to our final model. The quality of renders is also the same. 

\input{figures_suppmat/ablation_geom/ablation_geom}

\input{figures_suppmat/qualitative/qualitative}
\input{figures_suppmat/qualitative/additional_qualitative}
\input{figures_suppmat/colmap_jenya_last/colmap_jenya}
\input{figures_suppmat/colmap_ksyusha_last/colmap_ksyusha_additional}
\input{figures_suppmat/colmap_nastya/colmap_nastya_additional}
\input{figures_suppmat/colmap_person/colmap_person}
\input{figures_suppmat/qualitative/comparison_12_views}

%% file: figures_suppmat/table_ablation.tex
\begin{figure*}
    \centering
    \resizebox{0.95\linewidth}{!}{%
    \begin{tabular}{l rrr | rrr | rrr || rrr | rrr | rrr}
        \setlength{\tabcolsep}{0pt}
        & \multicolumn{9}{c}{\textbf{Straight hair}} & \multicolumn{9}{c}{\textbf{Curly hair}} \\
        \textbf{Method} & $2 / 20$ & $3 / 30$ & $4 / 40$ & $2 / 20$ & $3 / 30$ & $4 / 40$ & $2 / 20$ & $3 / 30$ & $4 / 40$ & $2 / 20$ & $3 / 30$ & $4 / 40$ & $2 / 20$ & $3 / 30$ & $4 / 40$ & $2 / 20$ & $3 / 30$ & $4 / 40$ \\
        \cline{2-19}
        & \multicolumn{3}{c}{\textbf{Precision}} & \multicolumn{3}{c}{\textbf{Recall}} & \multicolumn{3}{c}{\textbf{F-score}}& \multicolumn{3}{c}{\textbf{Precision}} & \multicolumn{3}{c}{\textbf{Recall}} & \multicolumn{3}{c}{\textbf{F-score}} \\
        \hline

        $\L_\text{geom}$	&	63.8	&	88.6	&	94.9	&	9.9	&	16.2	&	21.2	&	17.1	&	27.4	&	34.7&	50.8	&	75.1	&	85.9	&	5.7	&	11.3	&	18.4	&	10.2	&	19.6	&	30.3\\
        
 w/o $\L_\text{chm}$	&	82.9 &	95.0&	97.1&	4.5&	8.9	&	14.2&	8.5&	16.3&24.8	&	51.0&	73.8&	84.6&		3.9&	8.4&	14.3	&7.2 &15.1 &24.5\\
w/o $\L_\text{vol}$	&	48.3&	71.8	&79.4	&10.1	&	21.7&	32.2&	16.7&	33.3 &45.8	&20.1	&35.3	&45.5	&5.7	&12.4		&21.2	&	8.9	&18.4&28.9\\
w/o $\L_\text{orient}$ &	31.7&	56.2&69.0	&	6.0	&12.1	&17.8	&	10.1	&19.9	&	28.3&21.5	&43.7	&59.8	&4.7	&10.3		&17.7	&	7.7	&16.7& 27.3\\
\hline 
w/ $\L_\text{render}$~\cite{neuralstrands}	&	68.4	&	89.4	&	95	&	9.8	&	15.7	&	23.6	&	17.1	&	26.7	&	37.8 &	48.7	&	75.3	&	87.0	&	6.2	&	12.0	&	19.3	&	11.0	&	20.7	&	31.6\\
w/ $\L_\text{rgb}$	&	71.6	&	90.4	&	95.2	&	9.1	&	15.6	&	22.5	&	16.1	&	26.6&	36.4	&	49.3	&	76.0	&	87.7	&	6.1	&	12.0	&	19.4	&	10.9	&	20.7&	31.8\\
w/ $\L_\text{mask}$	&	63.5	&	88.2	&	94.6	&	11.1	&	17.3	&	22.5	&	18.9&	28.9	&	36.4	&	49.4	&	74.7	&	86.1	&	6.3	&	12.1	&	19.5	&	11.2	&	12.1	&	31.8\\
$\L_\text{fine} $ w/o $\L_\text{rgb}$ 	&	59.8	&	84.1	&	92.2	&	12.9	&	22.8	&	31.3	&	21.2	&	35.9	&	46.7	&	45.1	&	71.1	&	83.6	&	6.3	&	12.4	&	20.3	&	11.1	&	21.1	&	32.7 \\
$\L_\text{fine}$	&	59.9	&	84.1	&	92.1	&	13.1	&	22.7	&	31.5	&	21.5	&	35.8&	46.9&	45.8	&	72.1	&	84.6	&	6.4	&	12.8	&	21.0	&	11.2	&	21.7	&	33.6 \\

Neural Strands$^*$~\cite{neuralstrands}&		74.0	&	81.8	&	85.3	&	12.8	&	20.5	&	28.8 &	21.8&	32.8&	43.1&	38.4&	59.8&	72.4&	7.9	&	15.1&	23.8	&	13.1	&	24.1	&	35.8 \\
    \end{tabular}
    }
    \captionof{table}{We provide an extended quantitative evaluation of individual components of our method with per-scene metrics. Our full method with $\L_\text{fine}$ outperforms others in terms of Recall and F-score for both scenes. For a detailed discussion, please refer to Section~\ref{sec:ablation_suppmat}.}
    \label{tab:comparison_suppmat}
\end{figure*}

%% file: figures_suppmat/table_hyperparameters.tex
\begin{figure}
    \centering
    \resizebox{\linewidth}{!}{
    \begin{tabular}{l rrr | rrr | rrr }
        \textbf{Method} & $2 / 20$ & $3 / 30$ & $4 / 40$ & $2 / 20$ & $3 / 30$ & $4 / 40$ & $2 / 20$ & $3 / 30$ & $4 / 40$ \\
        \cline{2-10}
        & \multicolumn{3}{c}{\textbf{Precision}} & \multicolumn{3}{c}{\textbf{Recall}} & \multicolumn{3}{c}{\textbf{F-score}} \\
        \hline

         final\ model	&	45.8	&	72.1	&	84.6	&	6.4	&	12.8	&	21.0	&	11.2	&	21.7	&	33.6\\
        
        32 faces per pixel	& 42.8 & 67.8 & 81.4 & 5.8 &	11.7 & 19.5 & 10.2 & 20.0 & 31.5 \\
        1 face per pixel & 42.4 & 69.5 & 83.7 & 5.9 & 12.2 & 20.7 & 10.4 & 20.8 & 33.2 \\
        
        batch size 4 & 38.8 & 66.6 & 81.9 & 6.2 & 12.6 & 20.8 & 10.7 & 21.2 & 33.2 \\
        batch size 8 & 43.7 & 67.9 & 81.4 & 6.3 & 12.8 & 21.1 & 11.0 & 21.5 & 33.5 \\
    \end{tabular}
}   
    \captionof{table}{We provide an extended quantitative evaluation of hyperparameters of our method. Our final model outperforms others, showing that its set of hyperparameters is optimal.}
    \label{tab:comparison_hyperparameters}
\end{figure}

%% file: figures_suppmat/comp_NeuralHDAvatar/comp_neuralhdavatar.tex
\begin{figure}
    \begin{center}
    \includegraphics[width=0.15\textwidth]{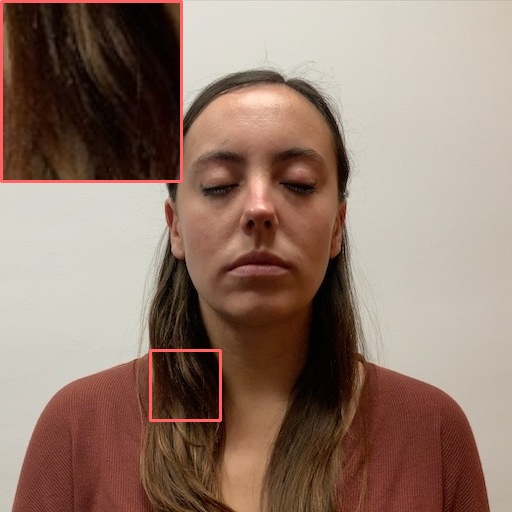}  
    \includegraphics[width=0.15\textwidth]{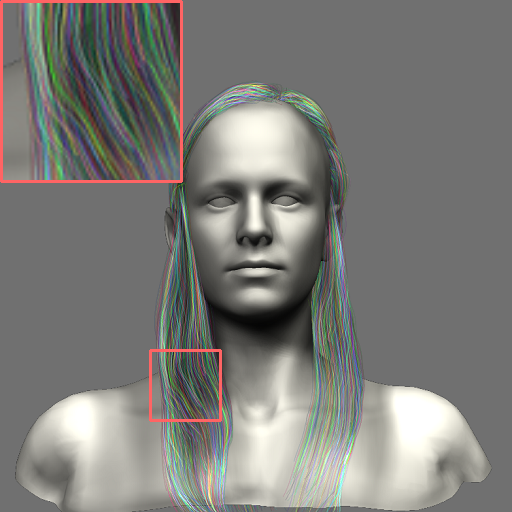}  
    \includegraphics[width=0.15\textwidth]{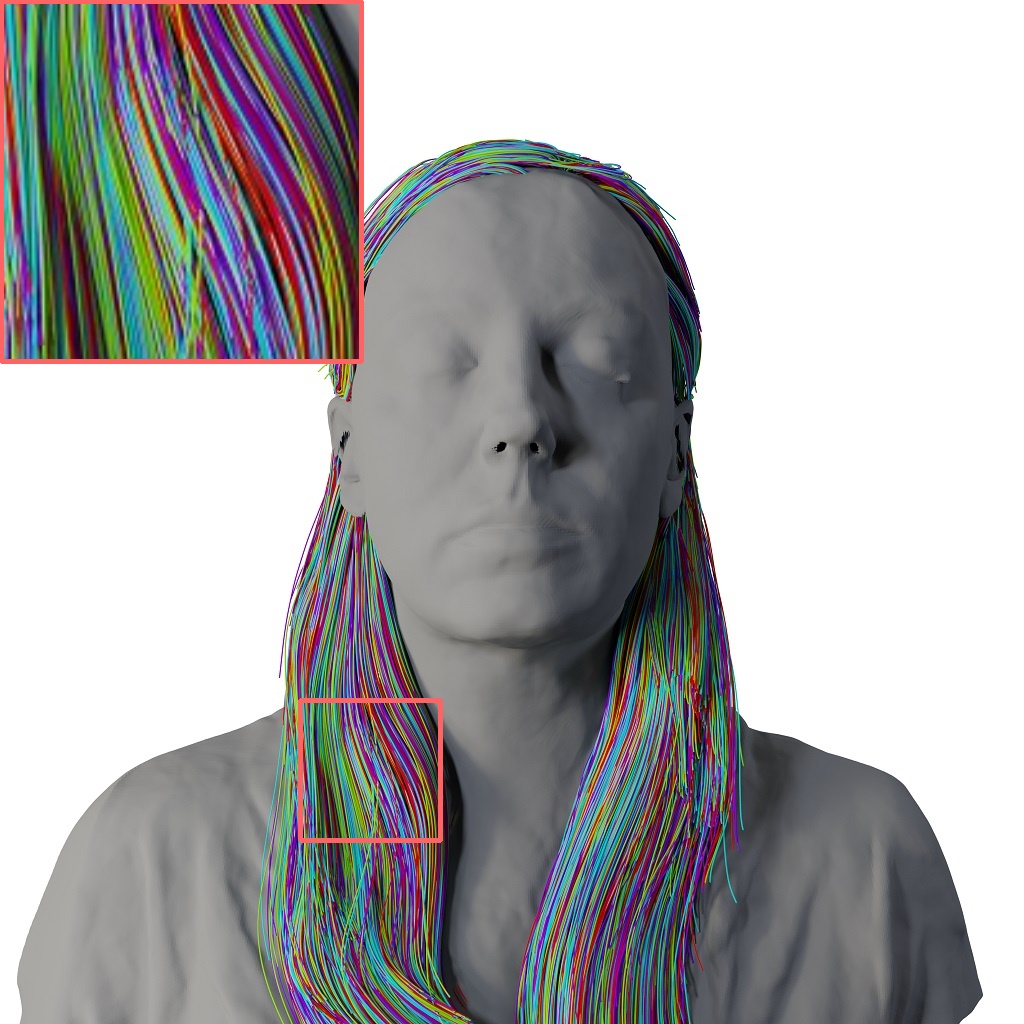}  \\
    
    \includegraphics[width=0.15\textwidth]{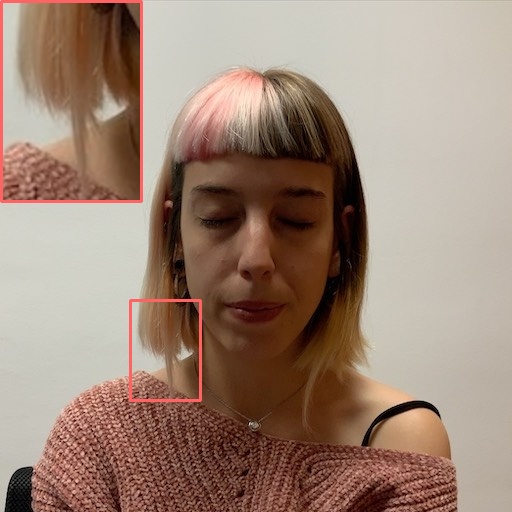}  
    \includegraphics[width=0.15\textwidth]{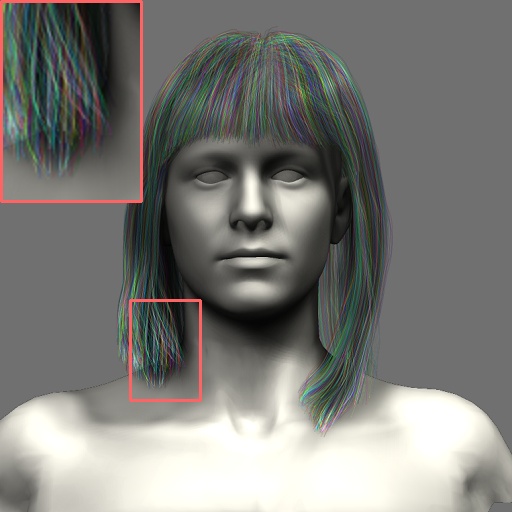}
    \includegraphics[width=0.15\textwidth]{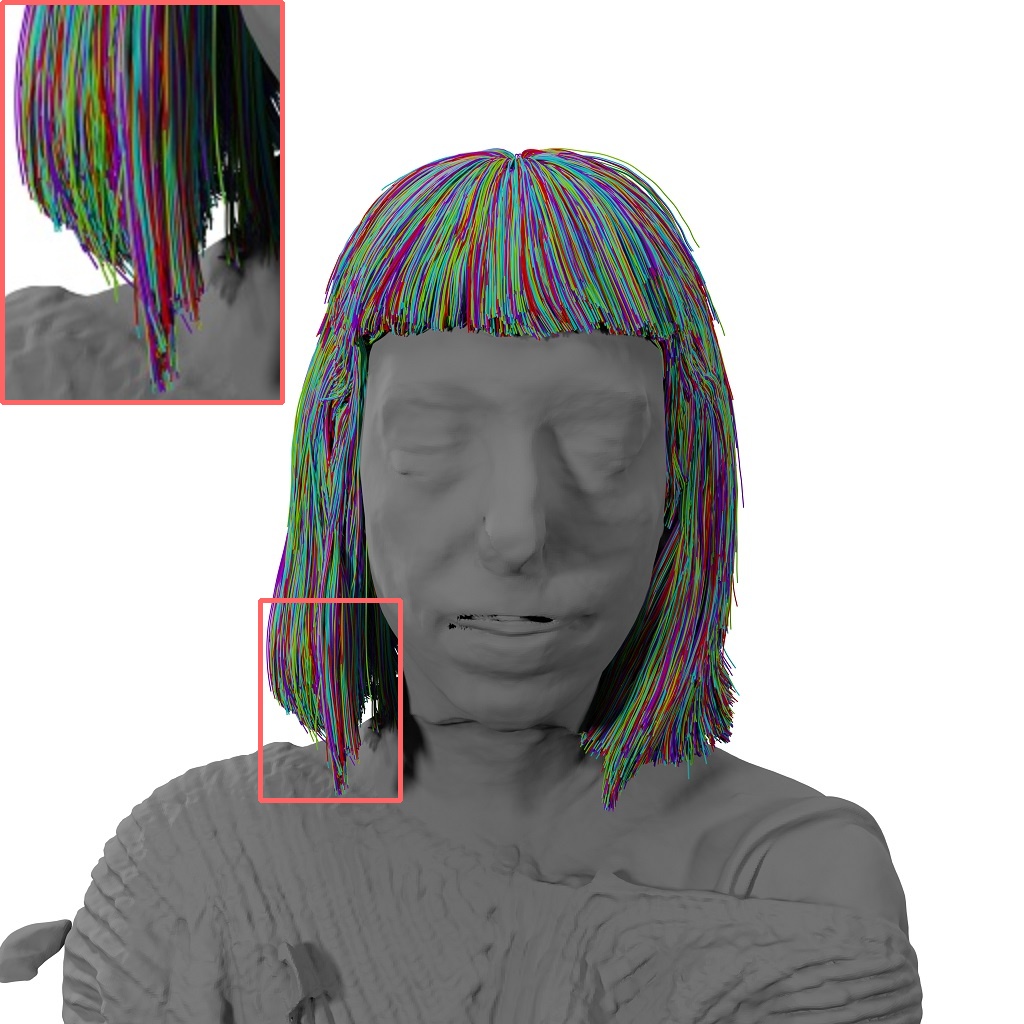}  \\
    
    \includegraphics[width=0.15\textwidth]{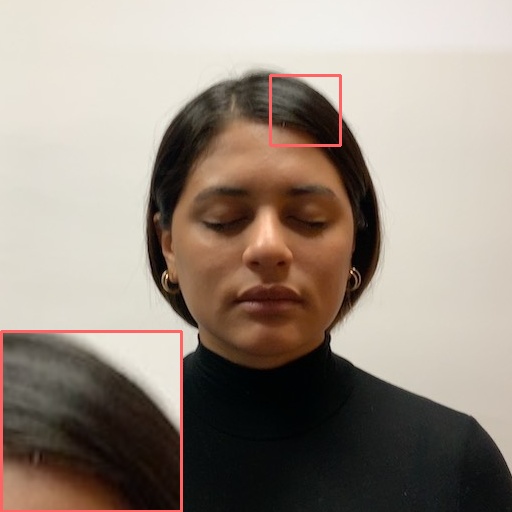}
    \includegraphics[width=0.15\textwidth]{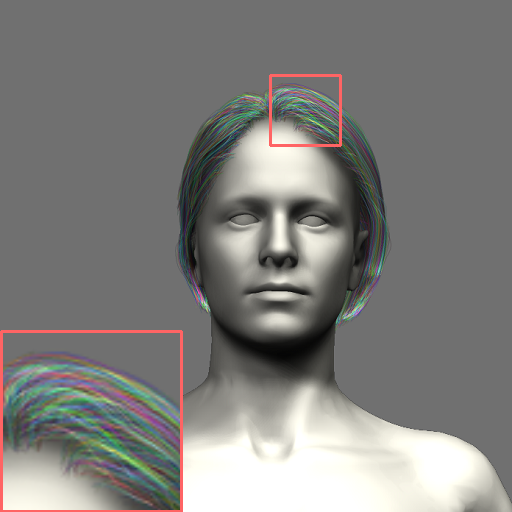}  
    \includegraphics[width=0.15\textwidth]{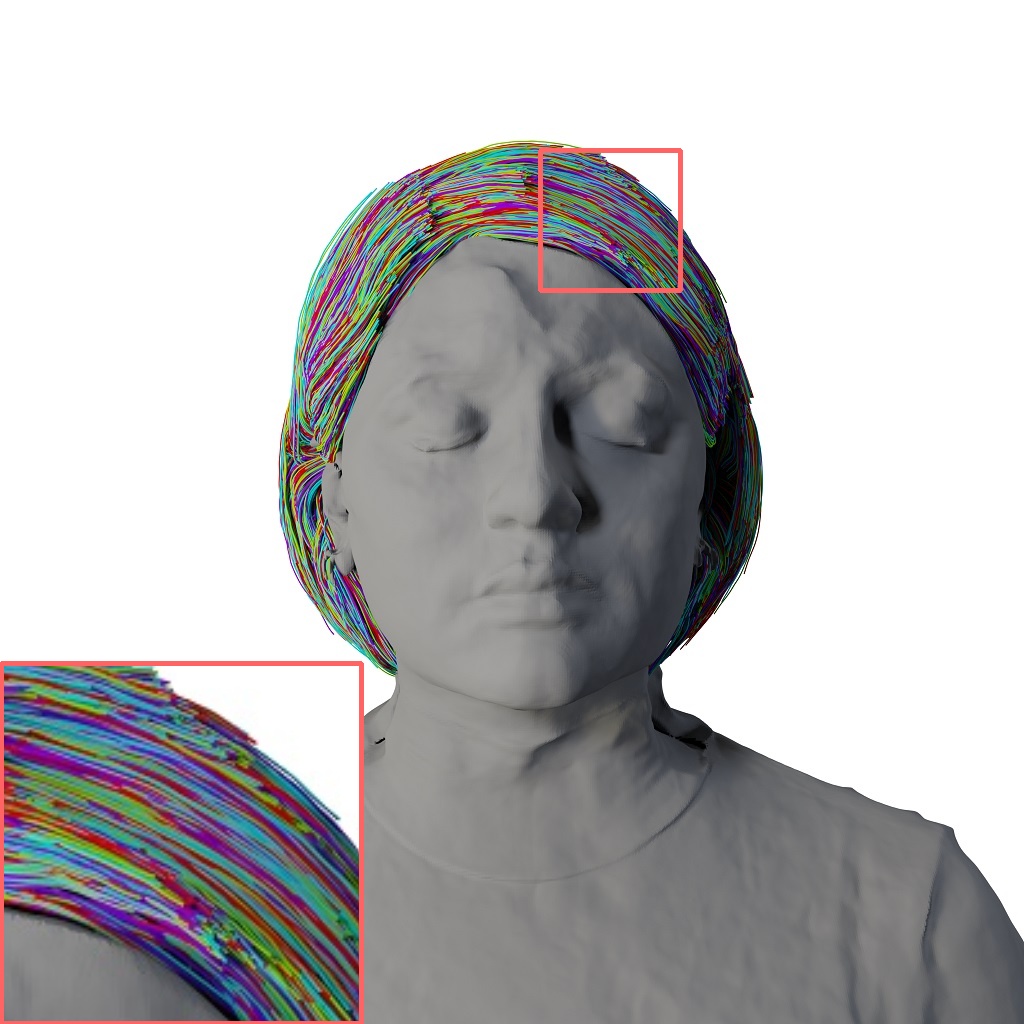} \\

    \includegraphics[width=0.15\textwidth]{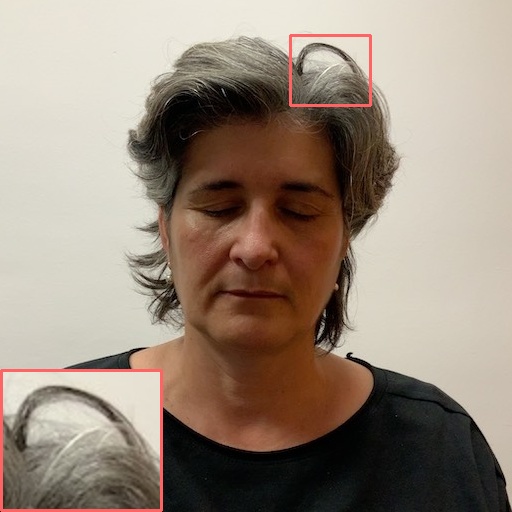}  
    \includegraphics[width=0.15\textwidth]{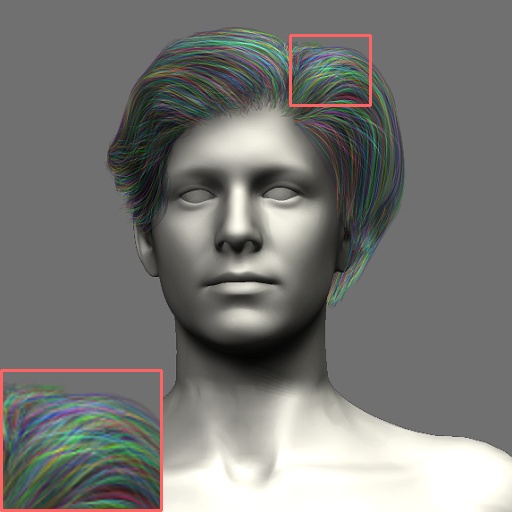}  
    \includegraphics[width=0.15\textwidth]{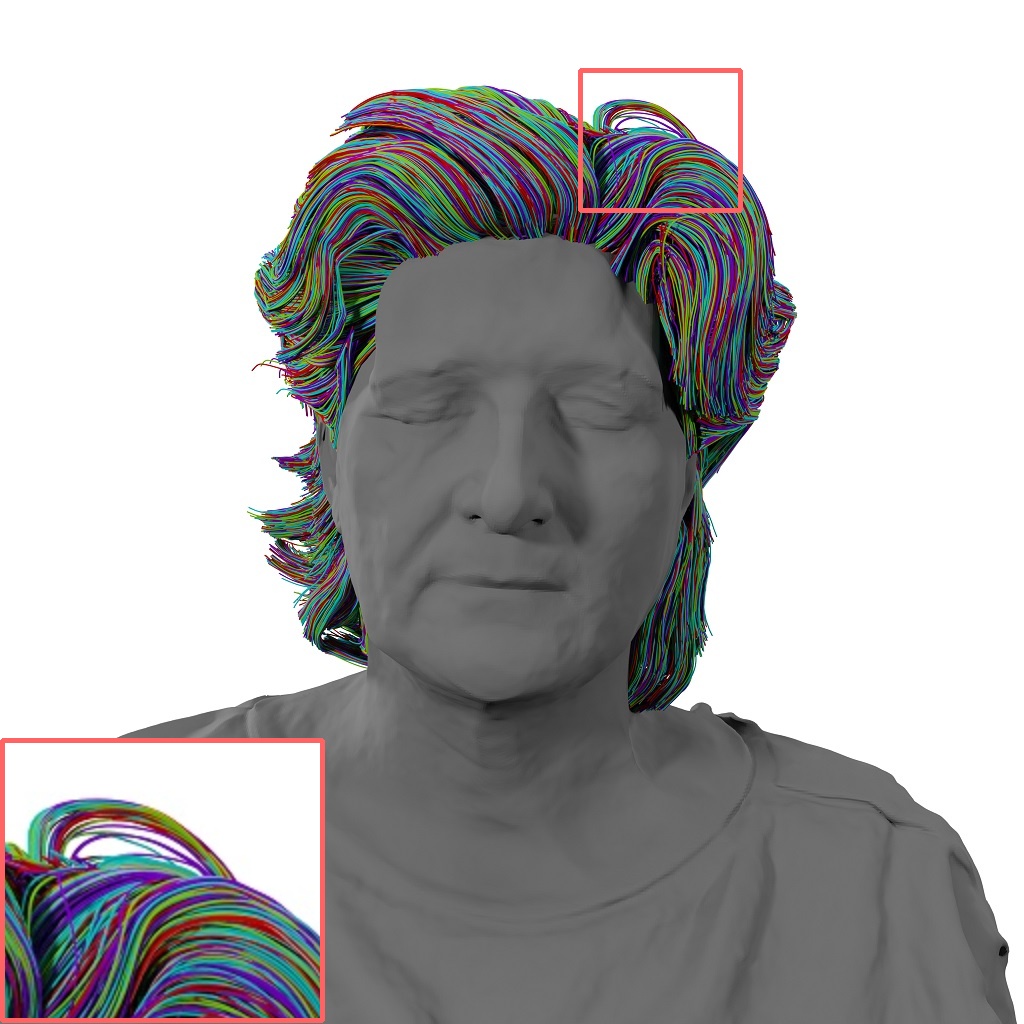}    
    \end{center}
    \vspace{-0.4cm}
    \caption{Comparison of our multi-view method (right) with a single-shot NeuralHDHair~\cite{Wu2022NeuralHDHairAH} system (middle). Digital zoom-in is recommended.}\label{fig:neural_hd_hair_suppmat}
\end{figure}

%% file: figures_suppmat/metrics_visual.tex
\begin{figure}
    \begin{tabular}{ccc}
    \includegraphics[width=0.15\textwidth]{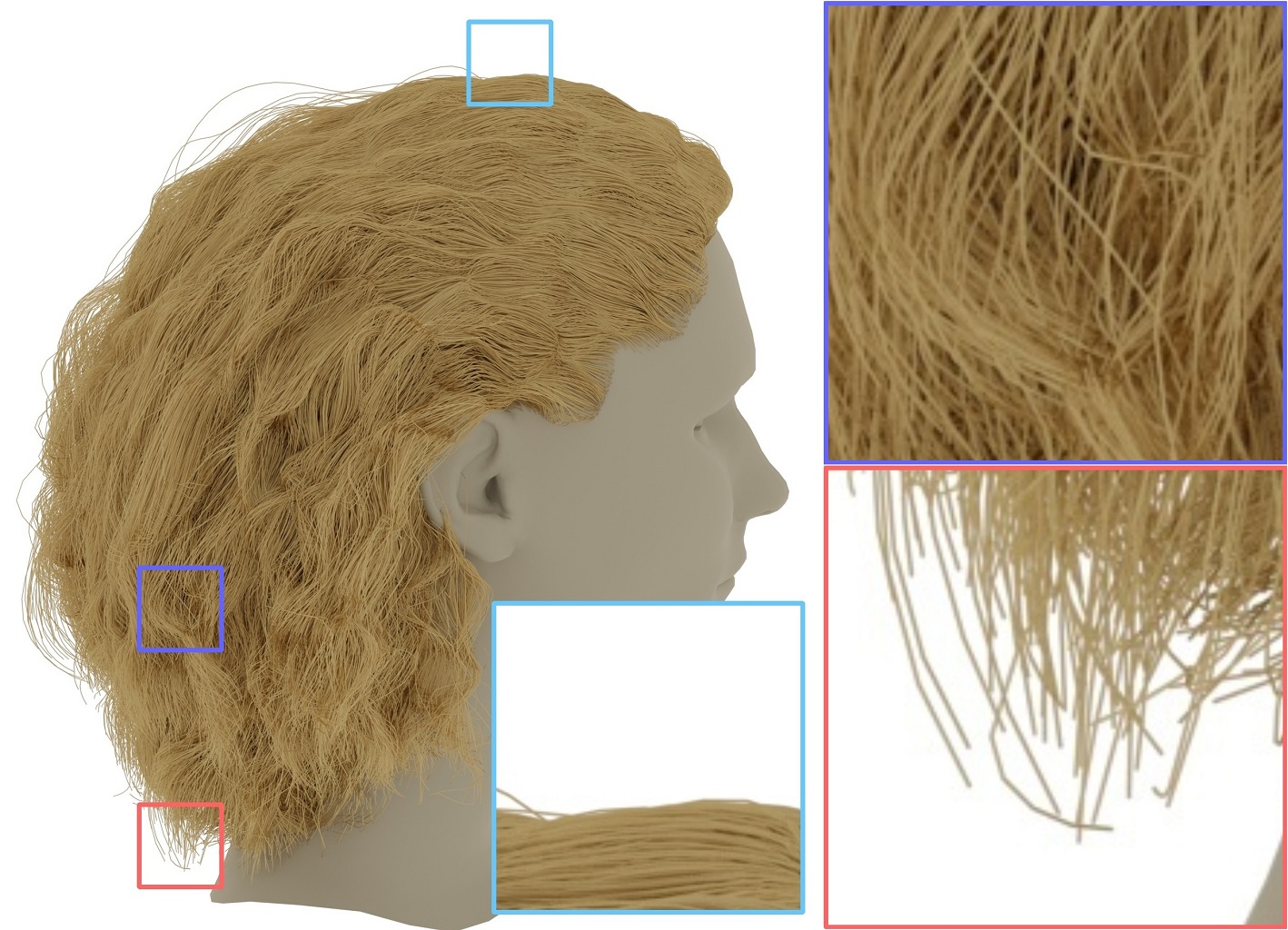} & \hspace{-.4cm}
    \includegraphics[width=0.15\textwidth]{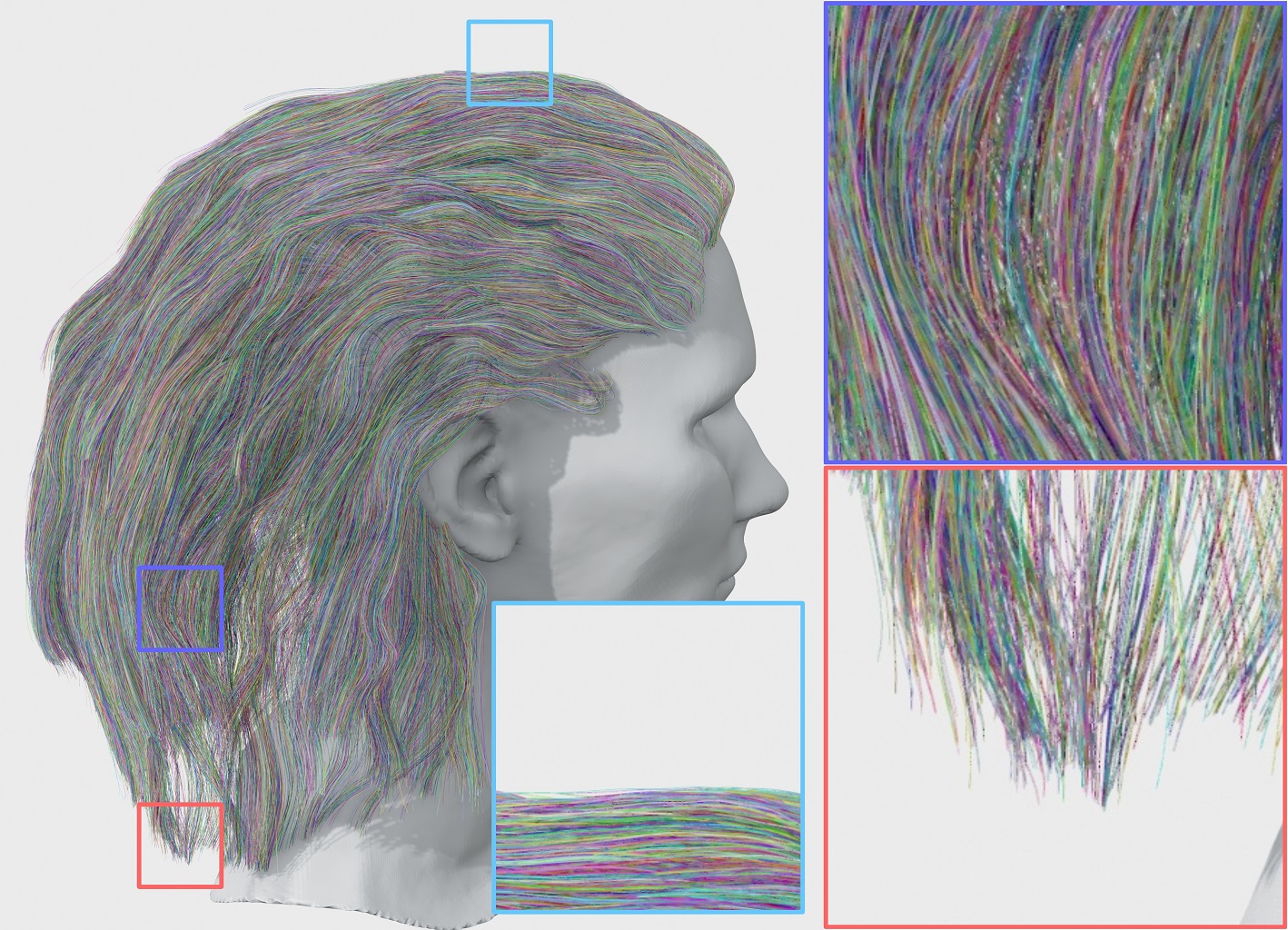}  & \hspace{-.4cm}
    \includegraphics[width=0.15\textwidth]{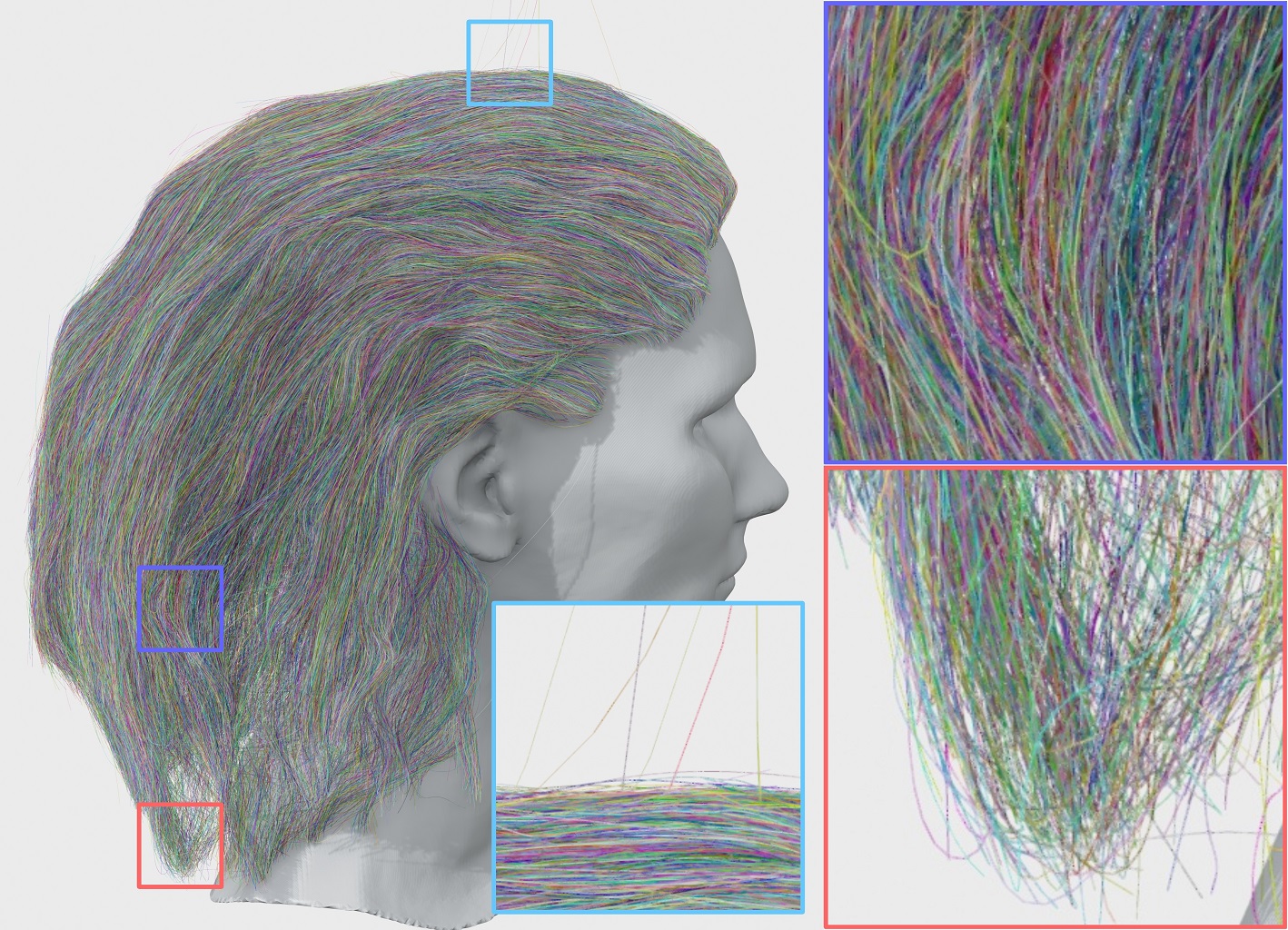}  \\
    \textbf{Image} & \hspace{-.4cm} \textbf{Ours} & \hspace{-.4cm} \textbf{Neural Strands$^*$} \\
    \end{tabular}
    \vspace{-0.3cm}
    \caption{A comparison with Neural Strands~\cite{neuralstrands} Our method obtains higher-quality hairstyles that lack the unrealistic curls visible in the top and bottom parts of the \cite{neuralstrands} reconstruction.}
    \label{fig:neural_strands_comp}
    \vspace{-0.4cm}
\end{figure}

%% file: figures_suppmat/ablation_geom/ablation_geom.tex
\begin{figure}
    \begin{center}
        \includegraphics[width=0.22\textwidth]{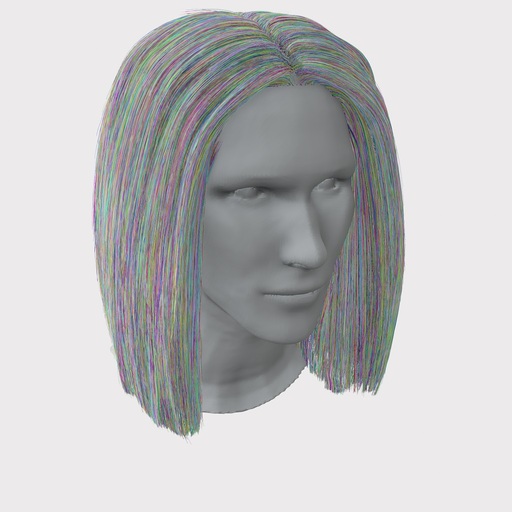}
        \includegraphics[width=0.22\textwidth]{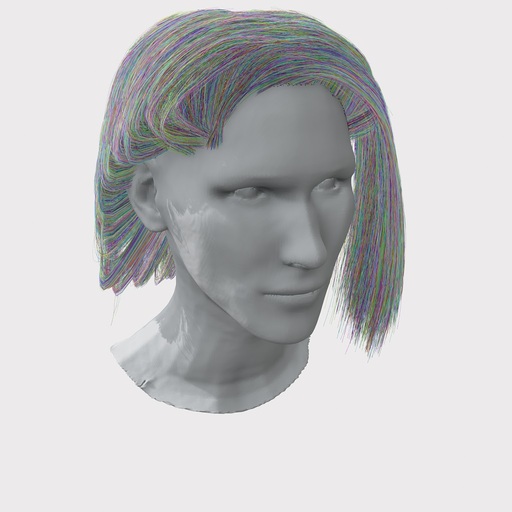}  \\ %
        \textbf{$\L_\text{geom}$}  \hspace{2.4cm} \textbf{w/o $\L_\text{chm}$}  \\ 
        \includegraphics[width=0.22\textwidth]{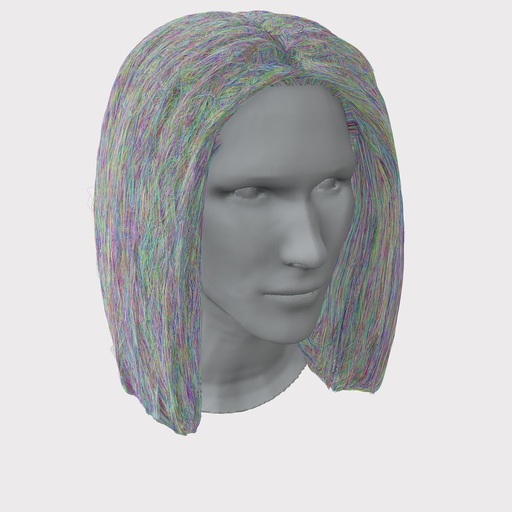}
       \includegraphics[width=0.22\textwidth]{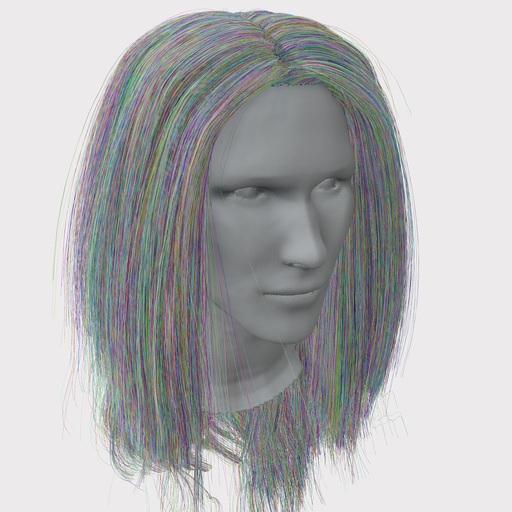}\\
       \textbf{w/o $\L_\text{orient}$}   \hspace{1.8cm} \textbf{w/o $\L_\text{vol}$}  \\ 
    \vspace{0.1cm}
    \caption{Ablation on individual components of geometry loss $\L_\text{geom}$. Without chamfer loss $\L_\text{chm}$ strands doesn't cover the whole hair silhouette. Removing orientation loss $\L_\text{orient}$ leads to random directions while removing the volume loss $\L_\text{vol}$ results in uncontrolled strands growing outside the hair region.}
    \label{fig:ablation_geom_supplem}
\end{center}
\end{figure}

%% file: figures_suppmat/qualitative/qualitative.tex
\begin{figure*}
    \begin{tabular}{ccccc}
        \includegraphics[width=0.185\textwidth]{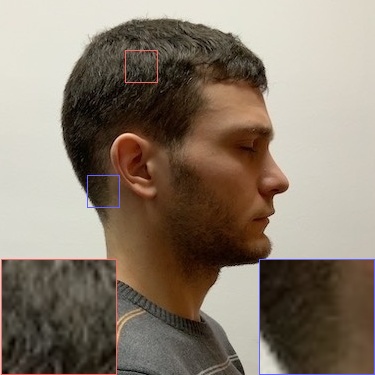} & \hspace{-0.31cm}
        \includegraphics[width=0.185\textwidth]{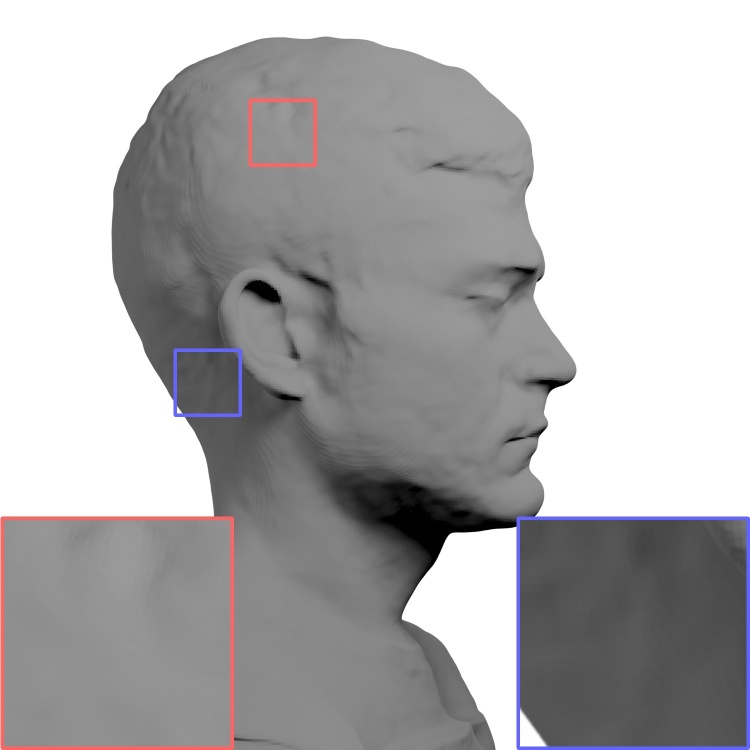} & \hspace{-0.31cm} 
        \includegraphics[width=0.185\textwidth]{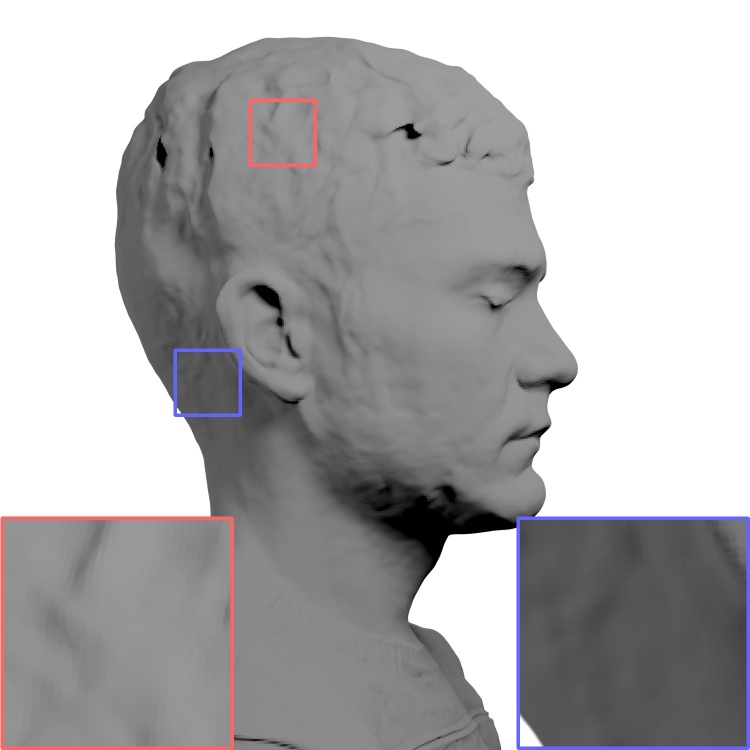} & \hspace{-0.31cm} 
        \includegraphics[width=0.185\textwidth]{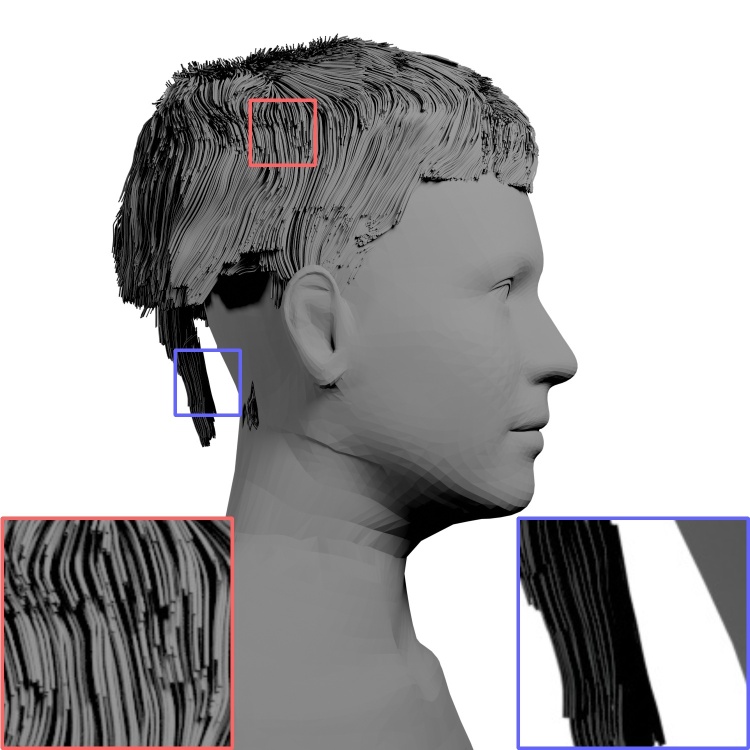} & \hspace{-0.31cm} 
        \includegraphics[width=0.185\textwidth]{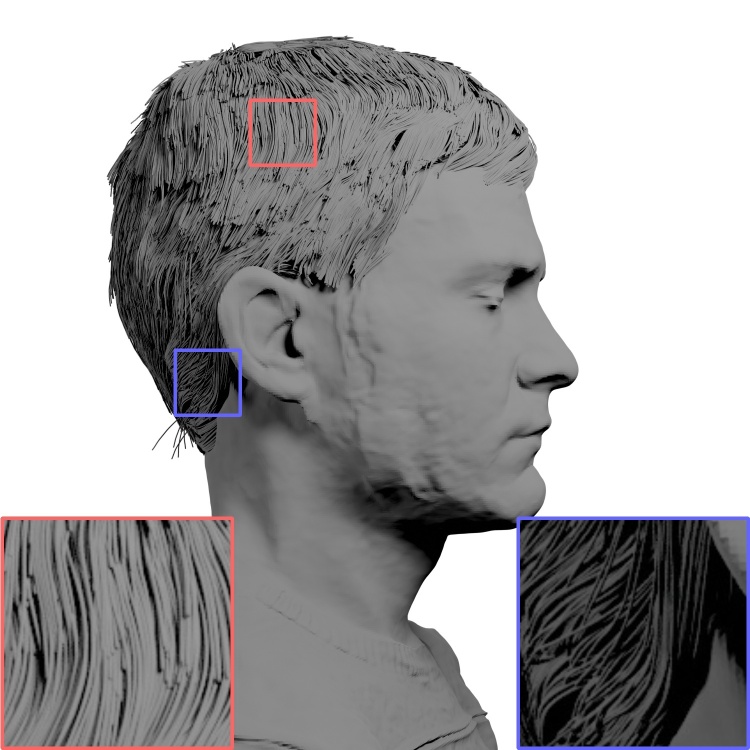} \\ %
        \includegraphics[width=0.185\textwidth]{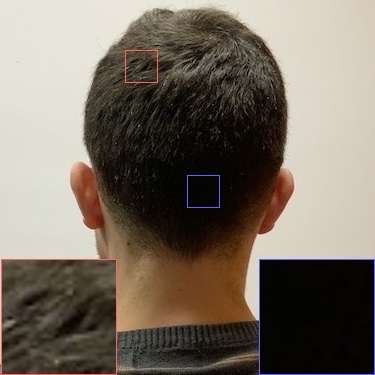} & \hspace{-0.31cm}
        \includegraphics[width=0.185\textwidth]{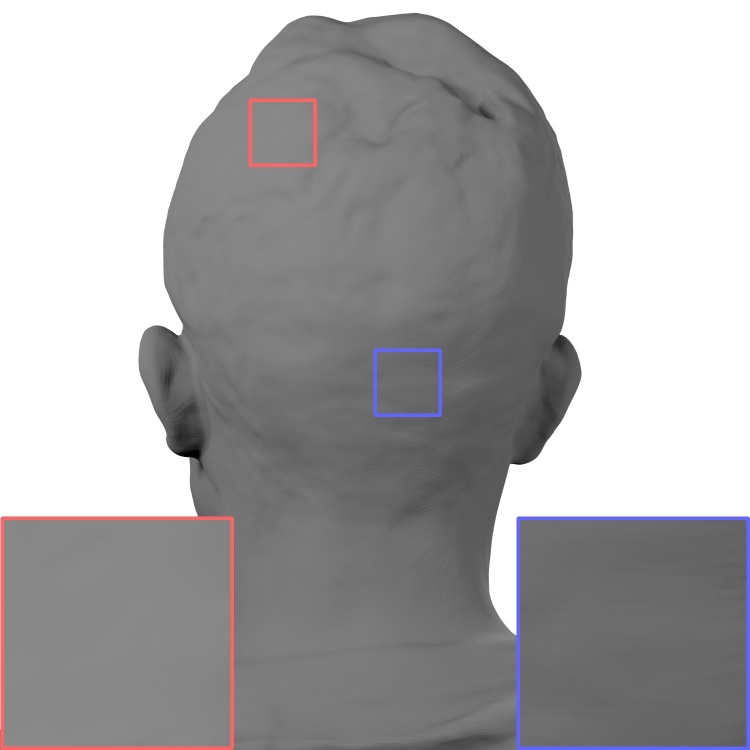} & \hspace{-0.31cm} 
        \includegraphics[width=0.185\textwidth]{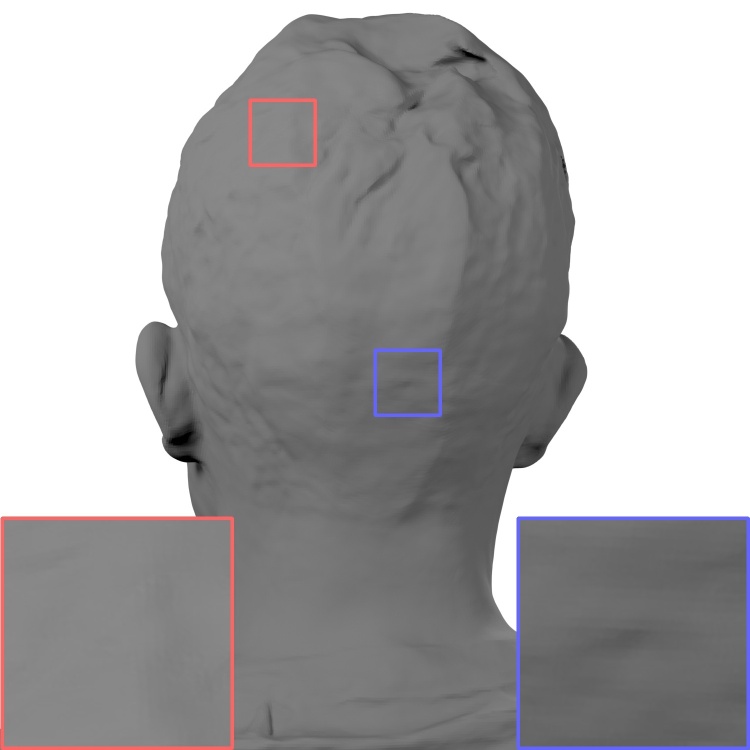} & \hspace{-0.31cm} 
        \includegraphics[width=0.185\textwidth]{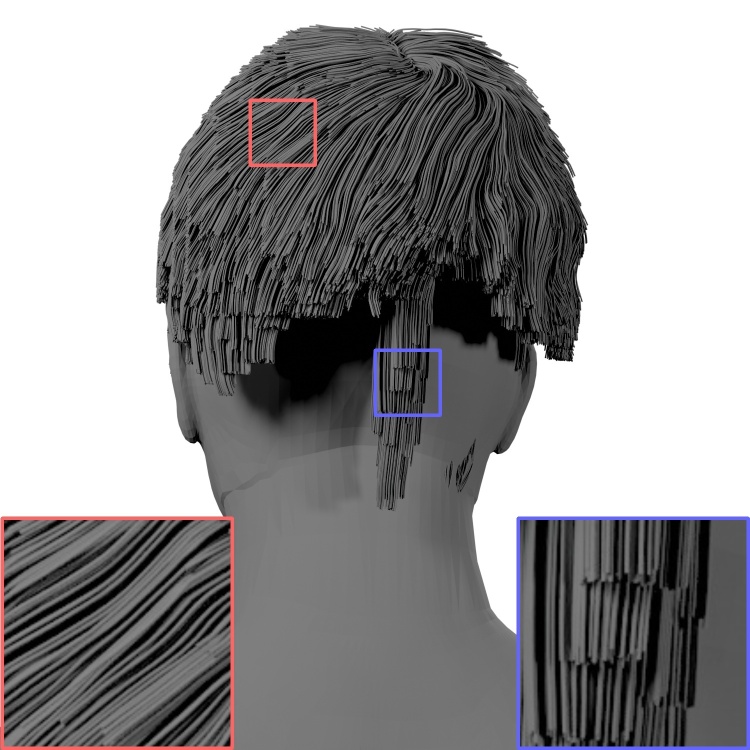} & \hspace{-0.31cm} 
        \includegraphics[width=0.185\textwidth]{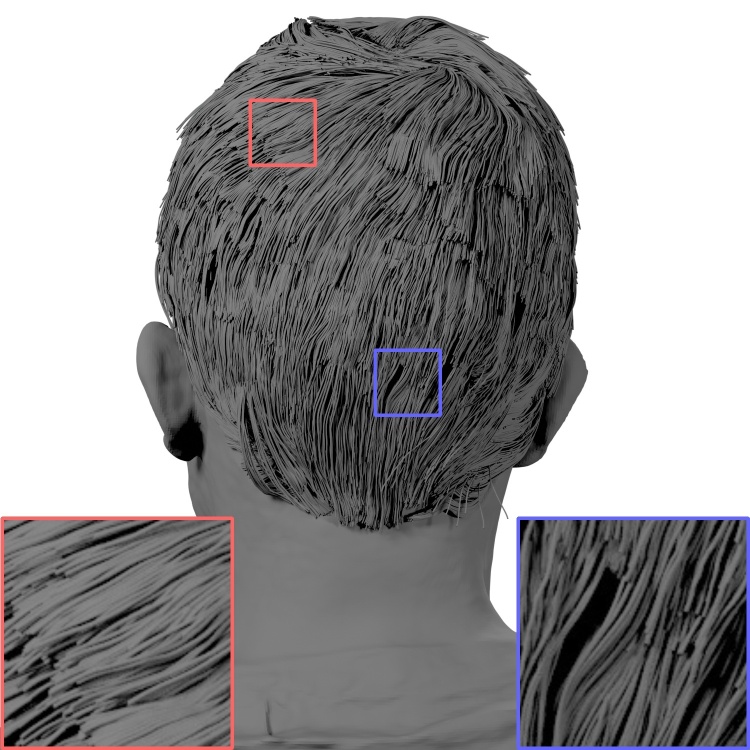} \\ %
        \includegraphics[width=0.185\textwidth]{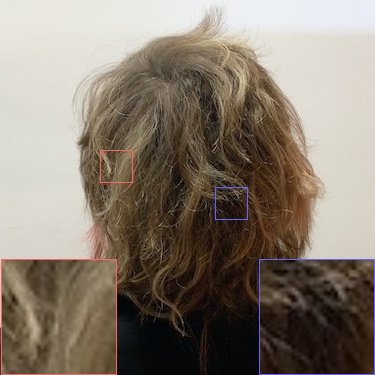} & \hspace{-0.31cm}
        \includegraphics[width=0.185\textwidth]{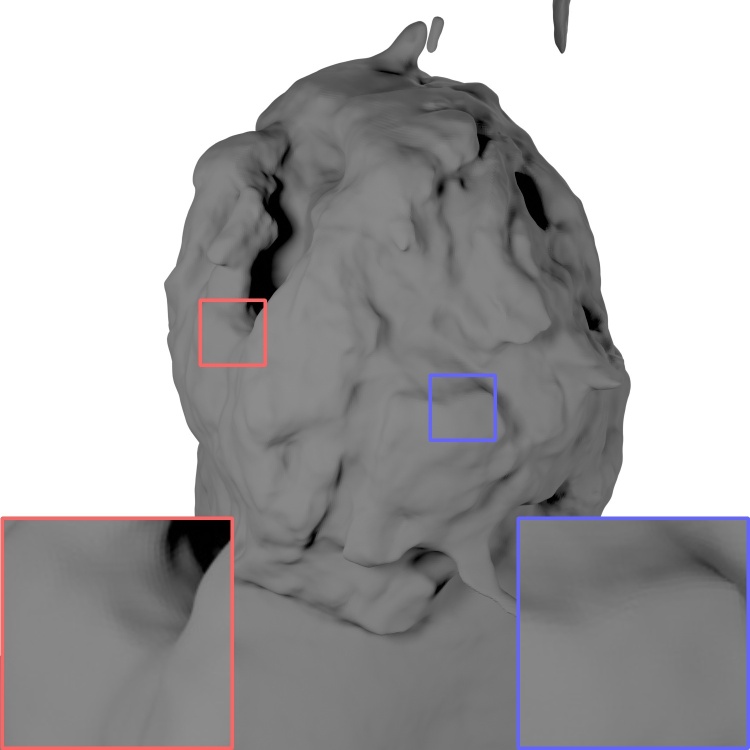} & \hspace{-0.31cm} 
        \includegraphics[width=0.185\textwidth]{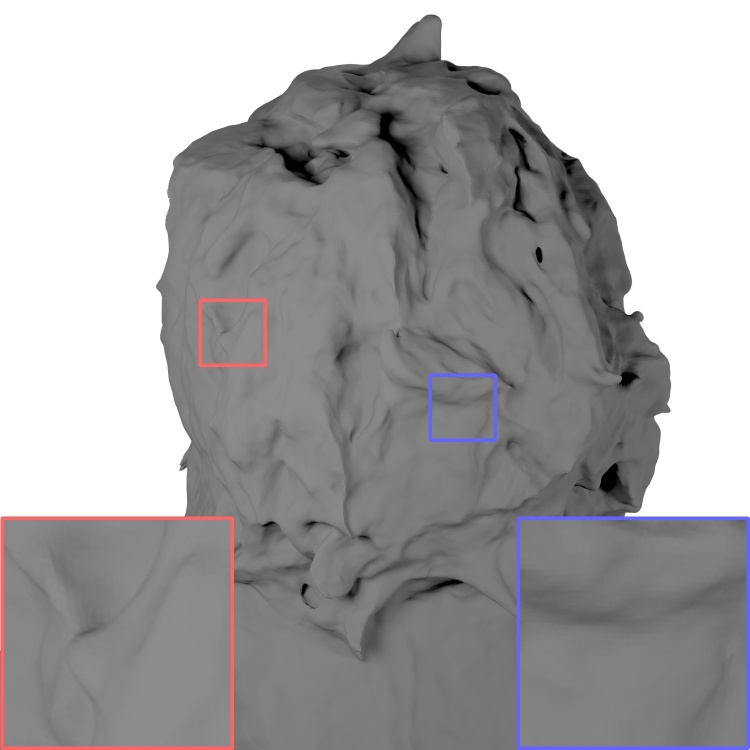} & \hspace{-0.31cm} 
        \includegraphics[width=0.185\textwidth]{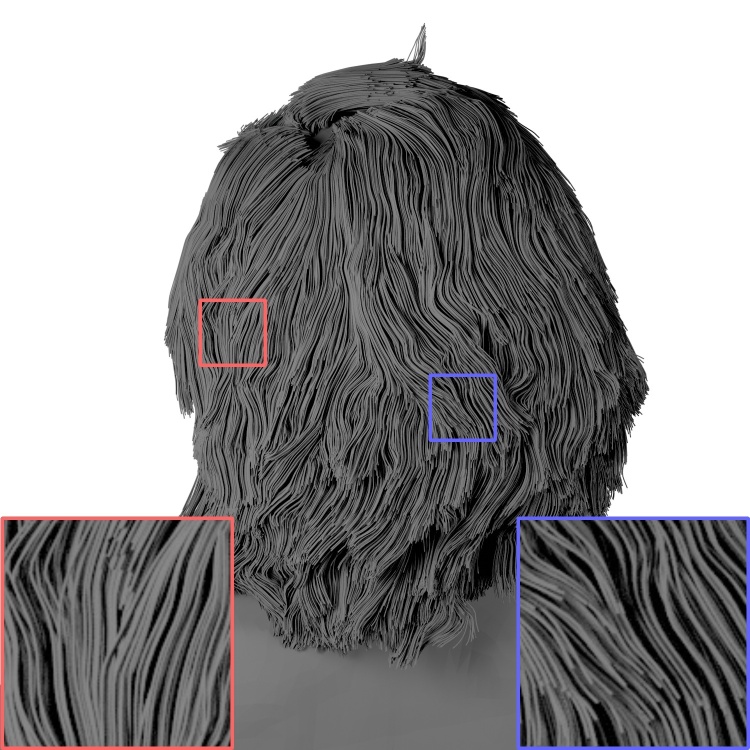} & \hspace{-0.31cm}        
        \includegraphics[width=0.185\textwidth]{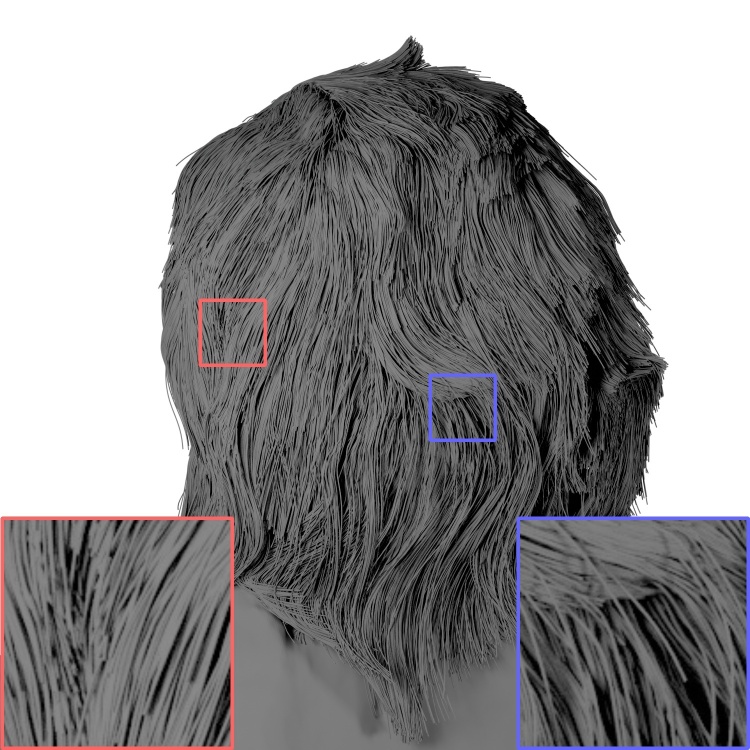} \\ %
         \includegraphics[width=0.185\textwidth]{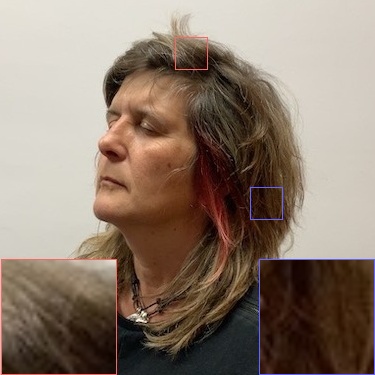} & \hspace{-0.31cm}
        \includegraphics[width=0.185\textwidth]{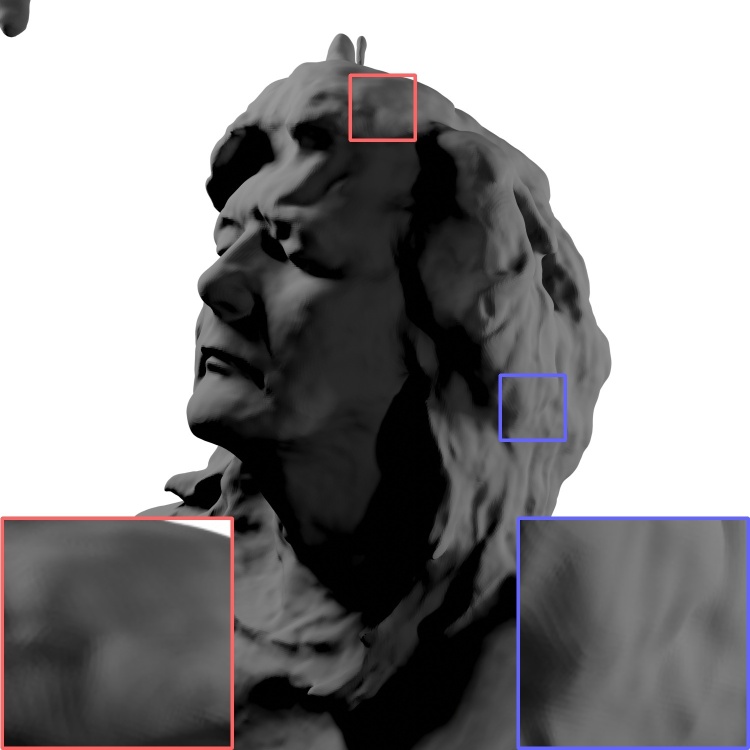} & \hspace{-0.31cm} 
        \includegraphics[width=0.185\textwidth]{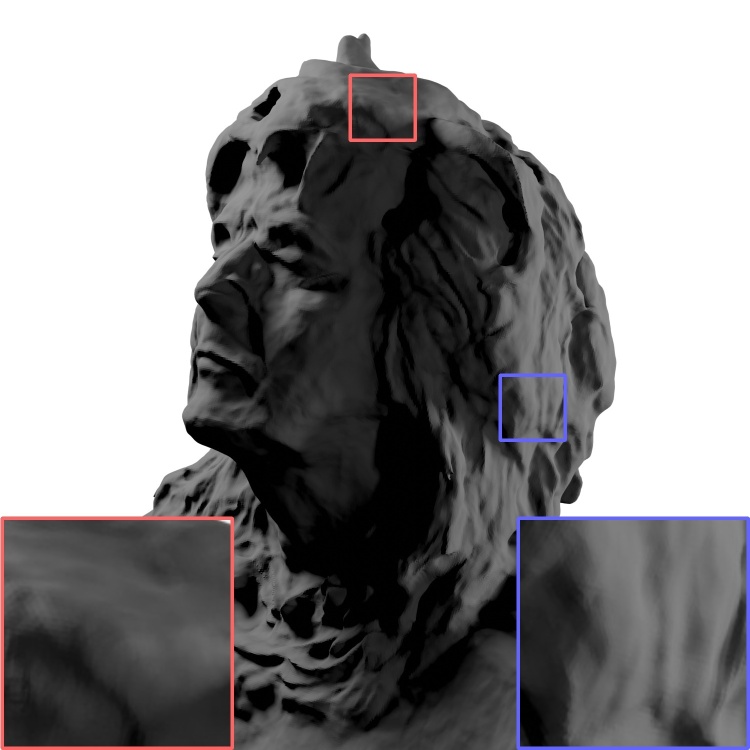} & \hspace{-0.31cm} 
        \includegraphics[width=0.185\textwidth]{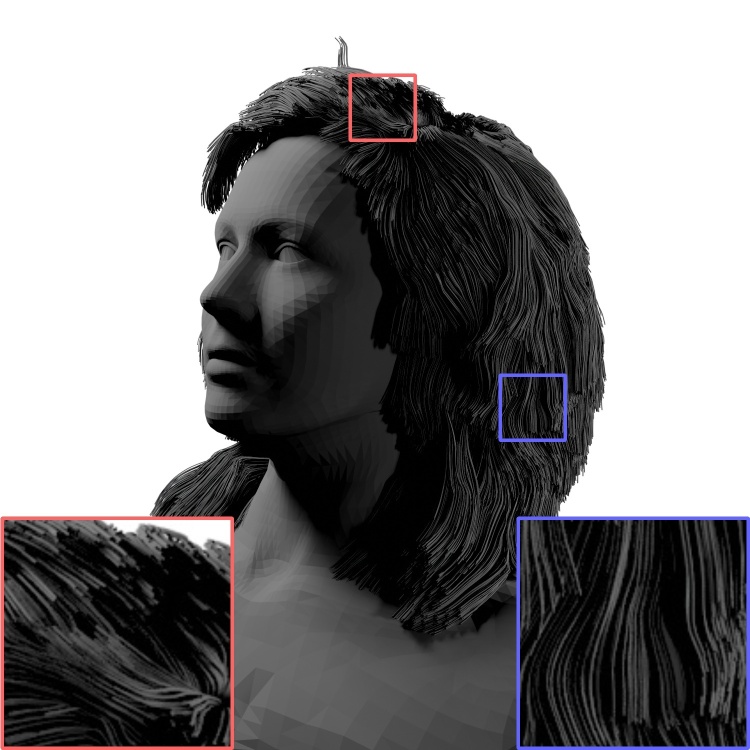} & \hspace{-0.31cm}        
        \includegraphics[width=0.185\textwidth]{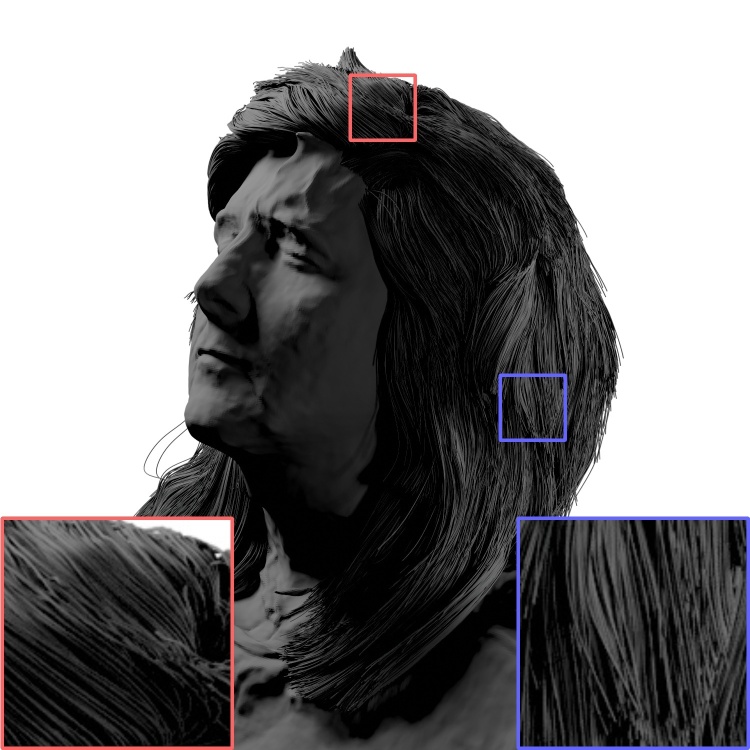} \\ %
        \includegraphics[width=0.185\textwidth]{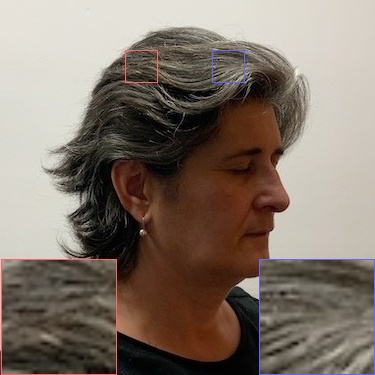} & \hspace{-0.31cm}
       \includegraphics[width=0.185\textwidth]{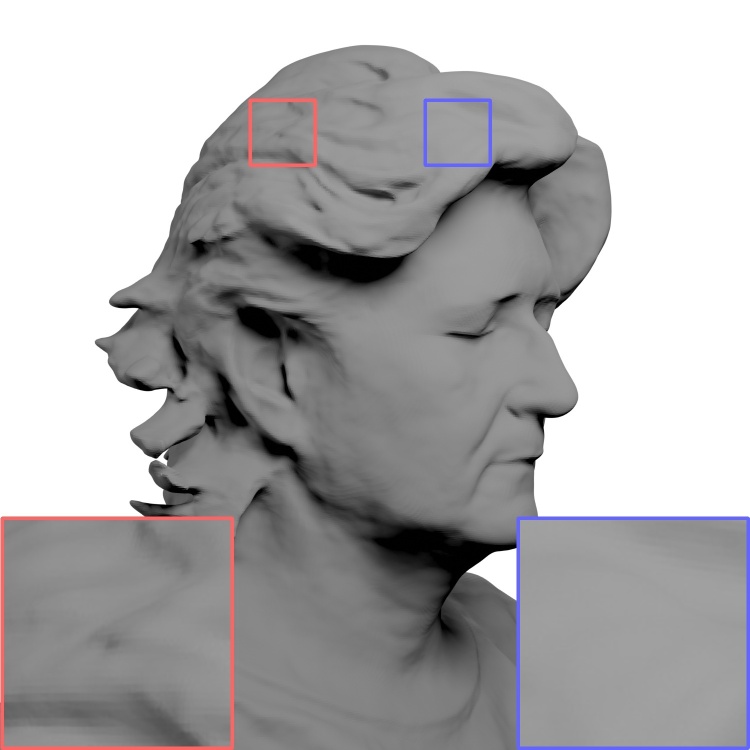} & \hspace{-0.31cm} 
        \includegraphics[width=0.185\textwidth]{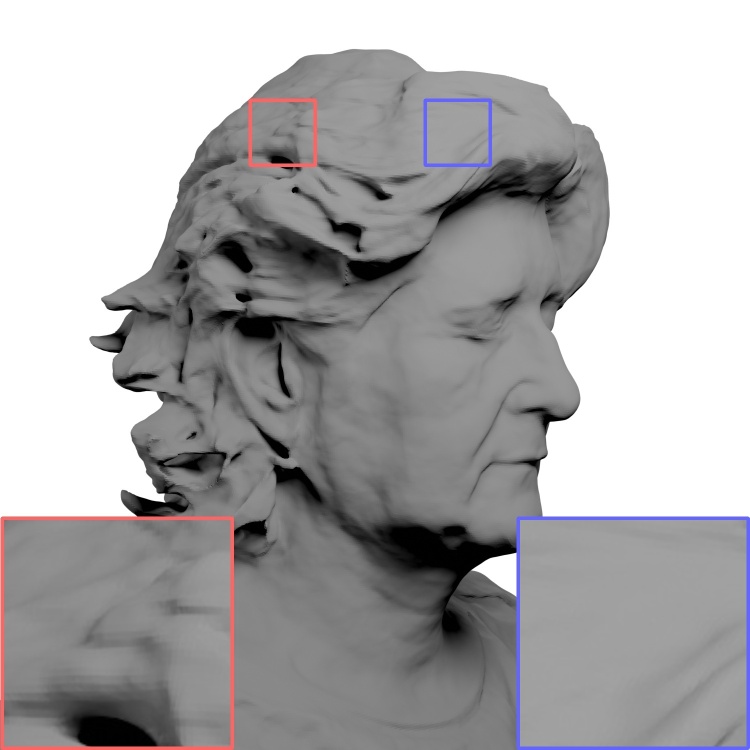} & \hspace{-0.31cm} 
        \includegraphics[width=0.185\textwidth]{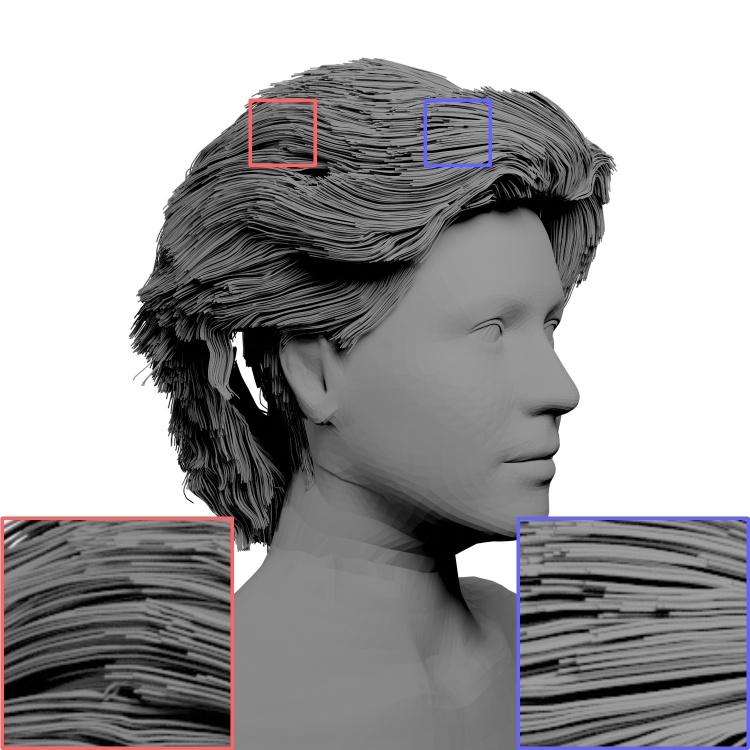} & \hspace{-0.31cm} 
        \includegraphics[width=0.185\textwidth]{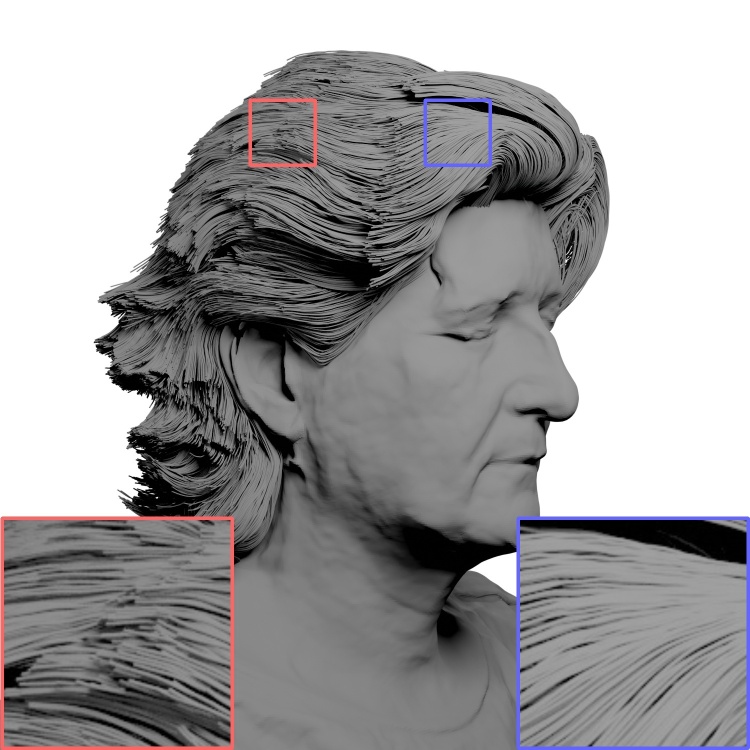} \\ %
        \includegraphics[width=0.185\textwidth]{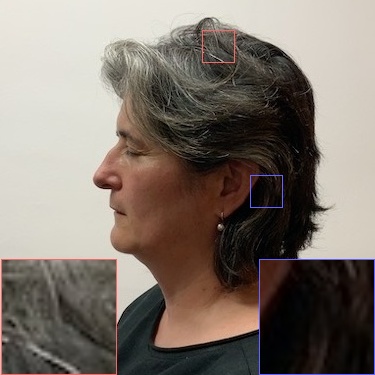} & \hspace{-0.31cm}
       \includegraphics[width=0.185\textwidth]{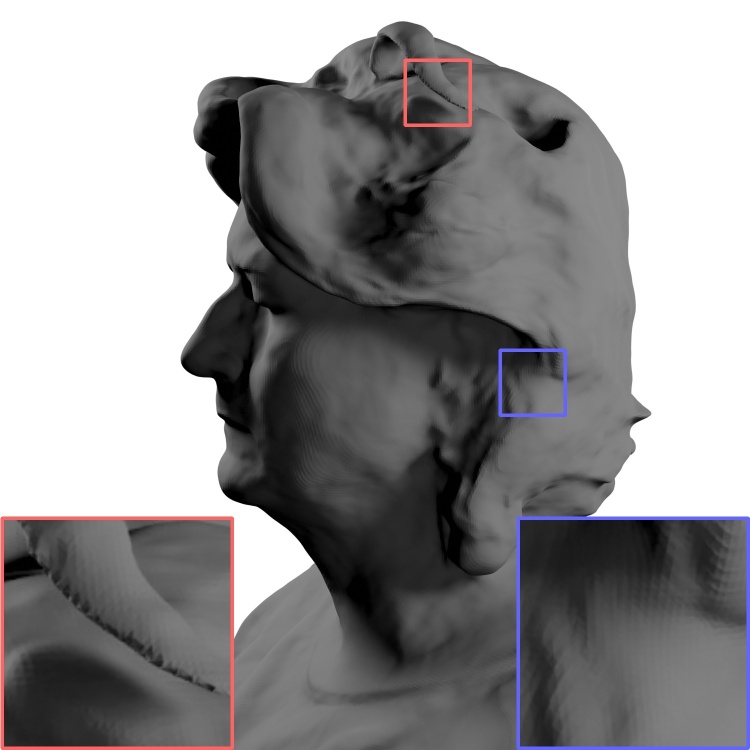} & \hspace{-0.31cm} 
        \includegraphics[width=0.185\textwidth]{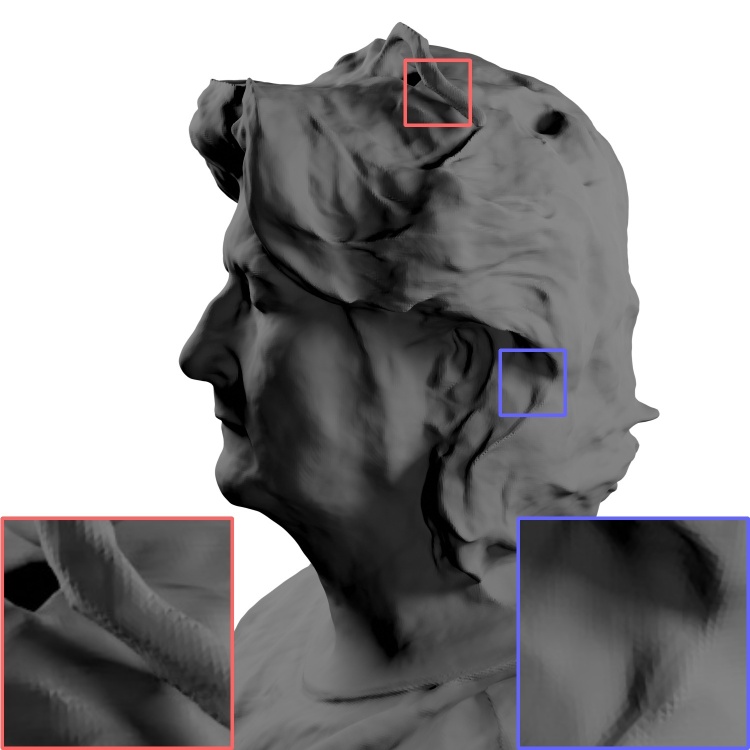} & \hspace{-0.31cm} 
        \includegraphics[width=0.185\textwidth]{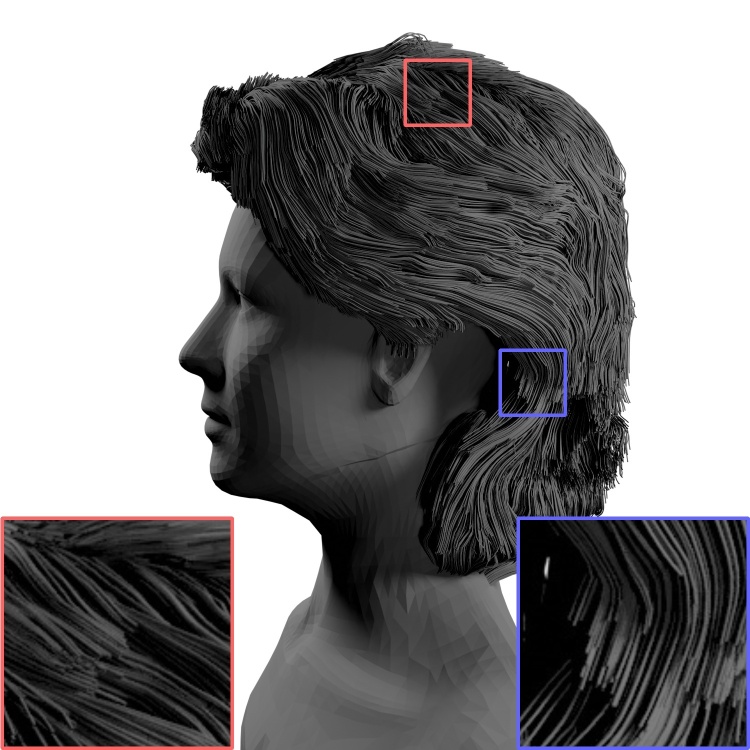} & \hspace{-0.31cm} 
        \includegraphics[width=0.185\textwidth]{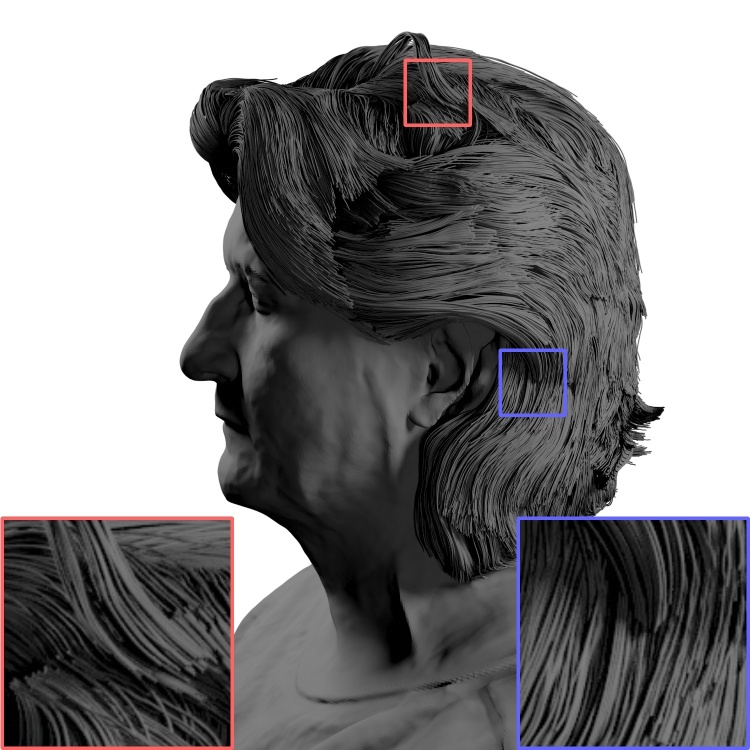} \\ %
        \textbf{Image}  & \hspace{-0.31cm} \textbf{UNISURF}& \hspace{-0.31cm} \textbf{NeuS}  & \hspace{-0.31cm} \textbf{DeepMVSHair} & \hspace{-0.31cm} \textbf{Ours}
    \end{tabular}
    \vspace{0.1cm}
    \caption{Extended qualitative comparison using real-world multi-view scenes~\cite{Ramon2021H3DNetFH}. Digital zoom-in is recommended.}
    \label{fig:geom_compare_suppmat}
\end{figure*}
\clearpage

%% file: figures_suppmat/qualitative/additional_qualitative.tex
\begin{figure*}
    \begin{tabular}{cccc}

         \includegraphics[width=0.24\textwidth]{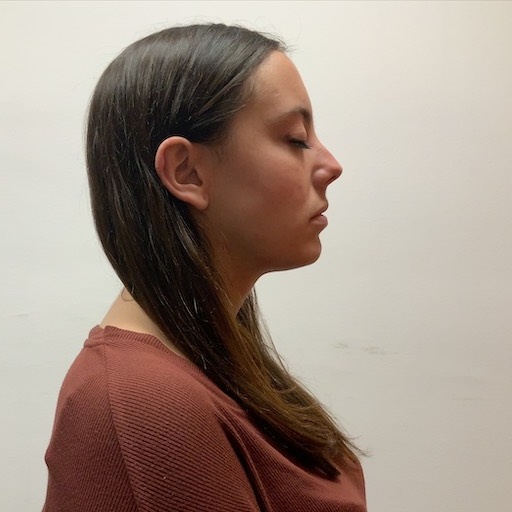} & \hspace{-0.31cm}
       \includegraphics[width=0.24\textwidth]{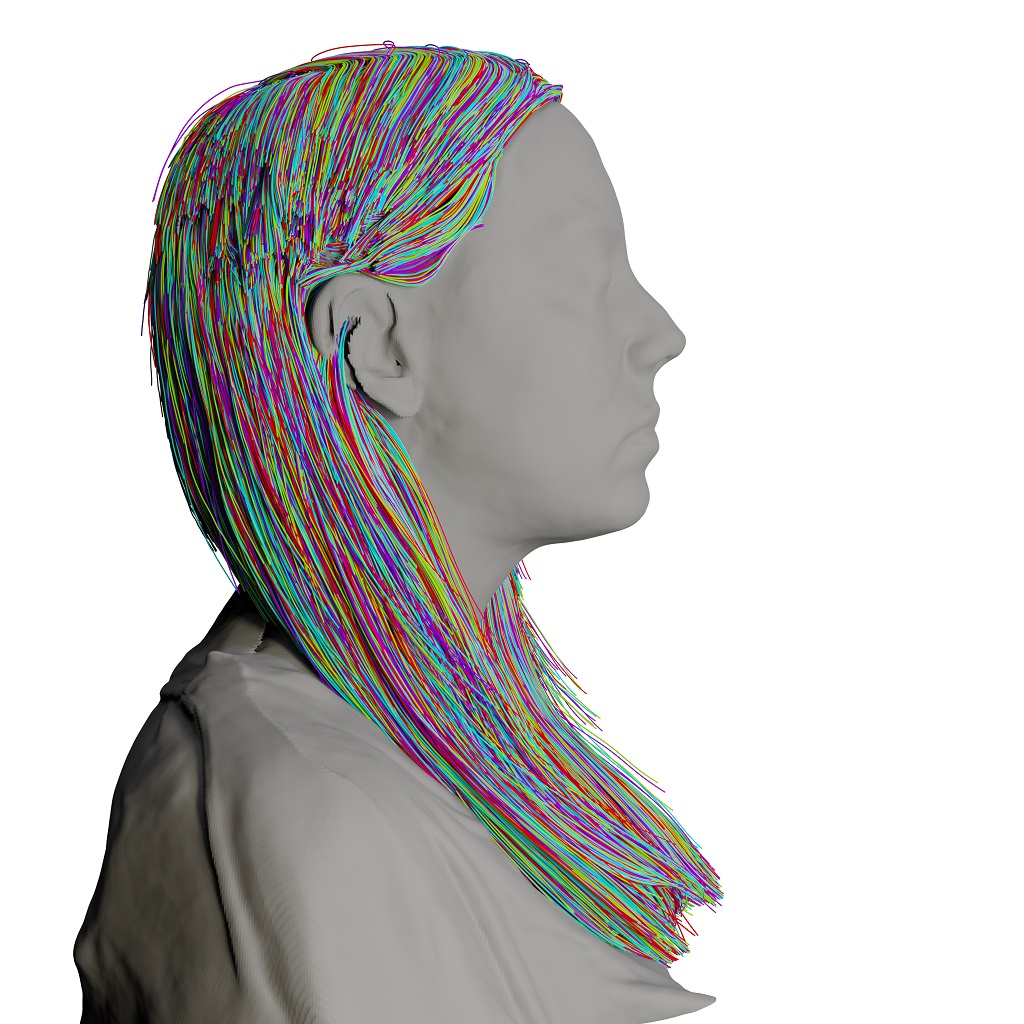} &
       \hspace{-0.31cm} 
        \includegraphics[width=0.24\textwidth]{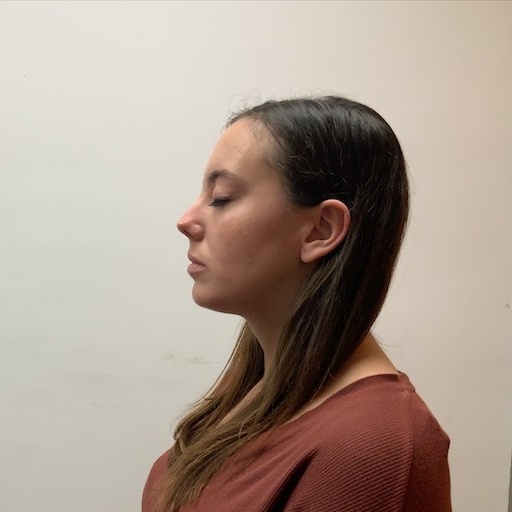} & \hspace{-0.31cm} 
        \includegraphics[width=0.24\textwidth]{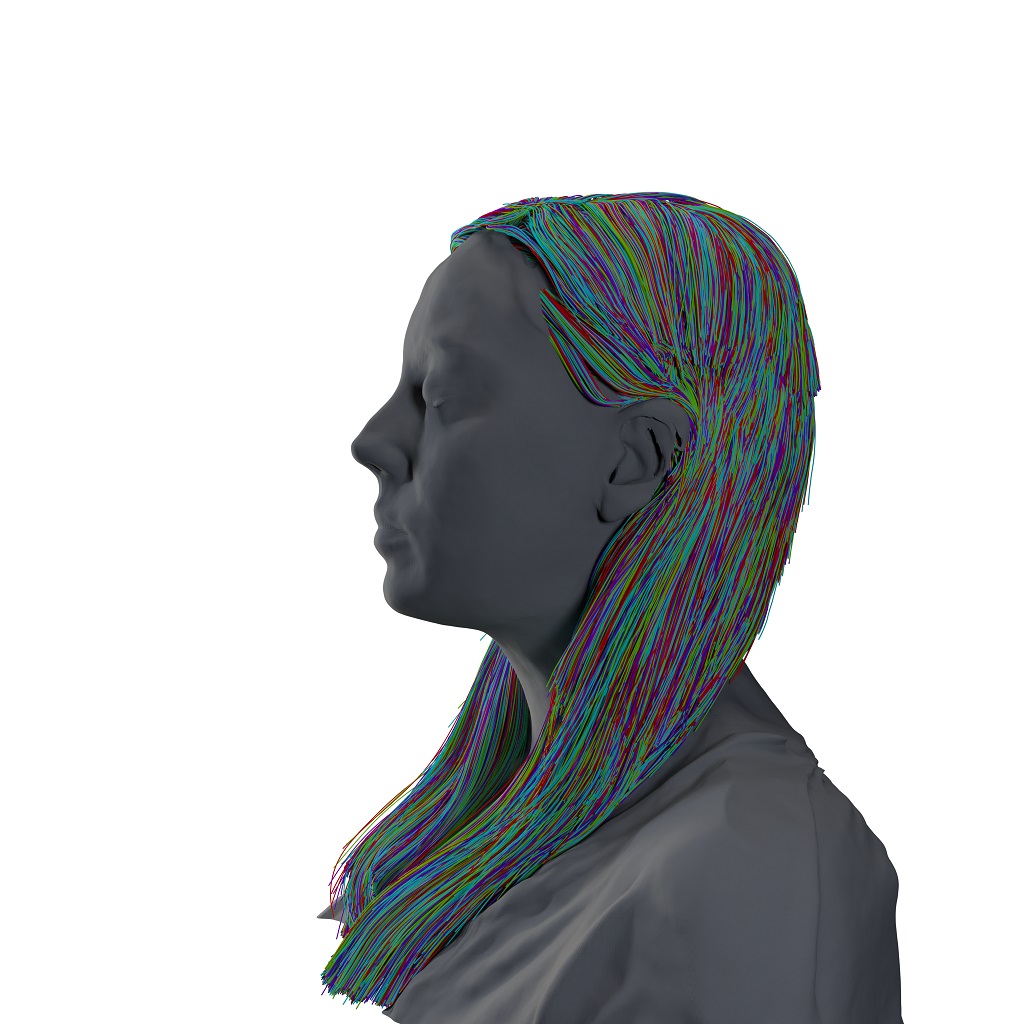} \\ 
        \hspace{-0.31cm}
        \includegraphics[width=0.24\textwidth]{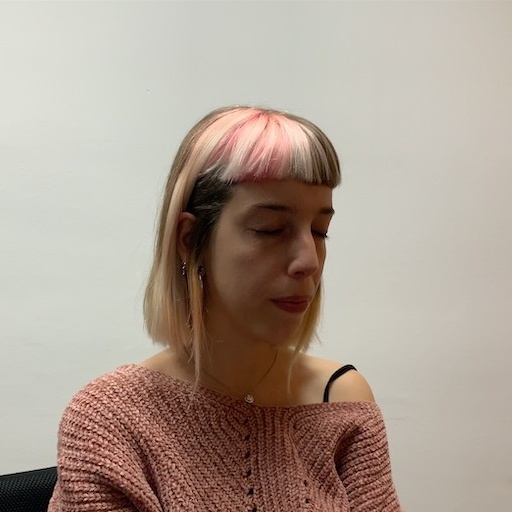} & \hspace{-0.31cm}
        \includegraphics[width=0.24\textwidth]{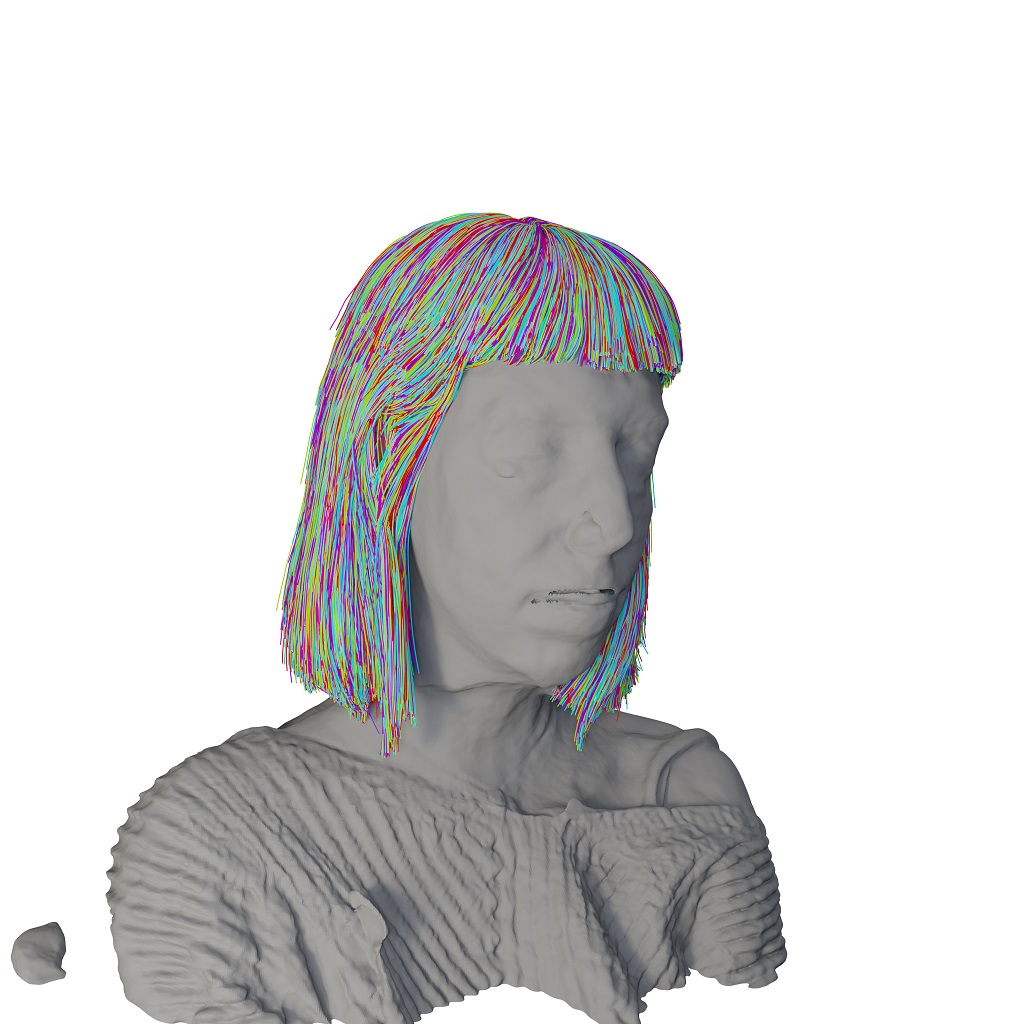} & 
        \hspace{-0.31cm} 
        \includegraphics[width=0.24\textwidth]{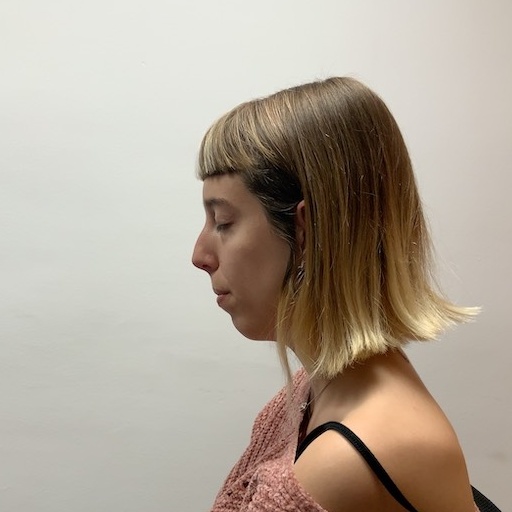} & \hspace{-0.31cm} 
        \includegraphics[width=0.24\textwidth]{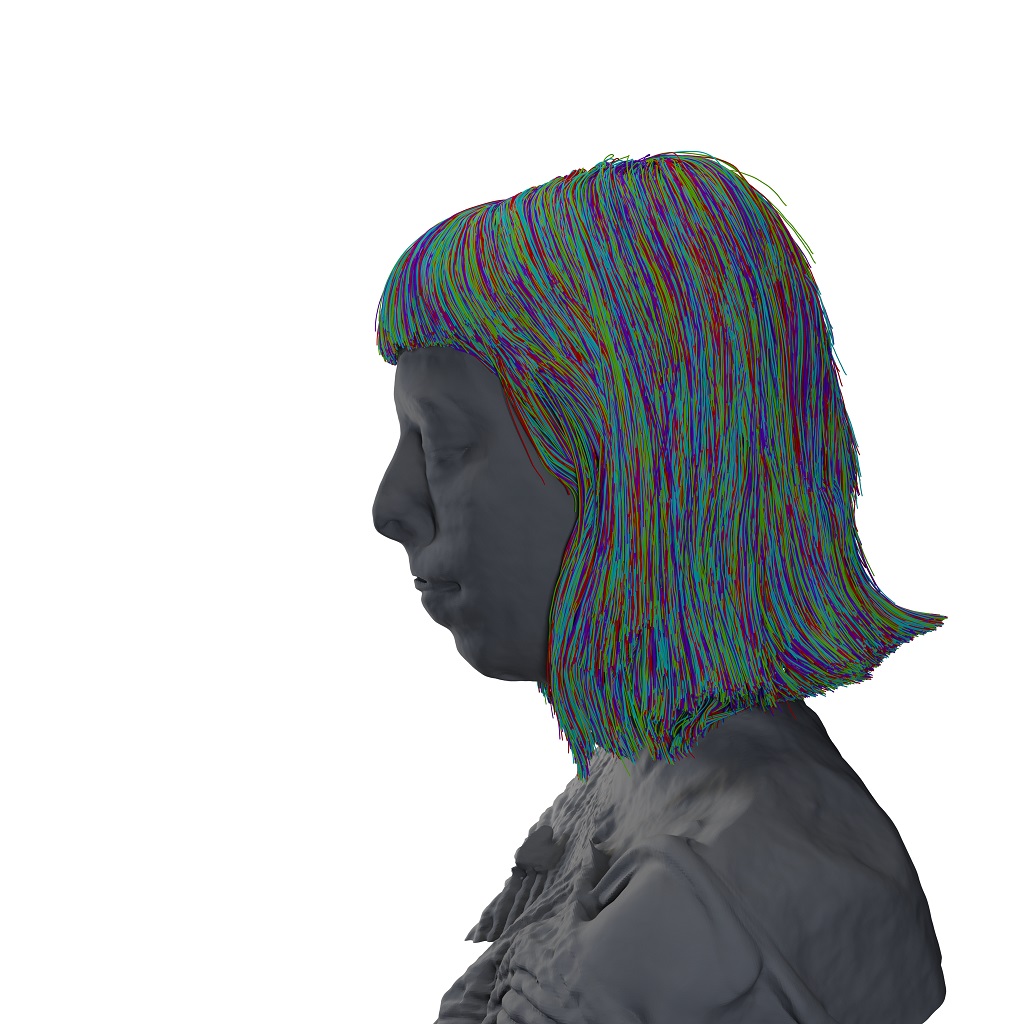}   \\ %
        \includegraphics[width=0.24\textwidth]{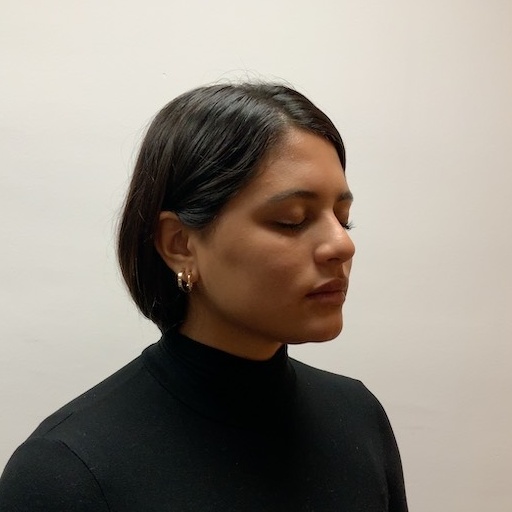} & \hspace{-0.31cm}
       \includegraphics[width=0.24\textwidth]{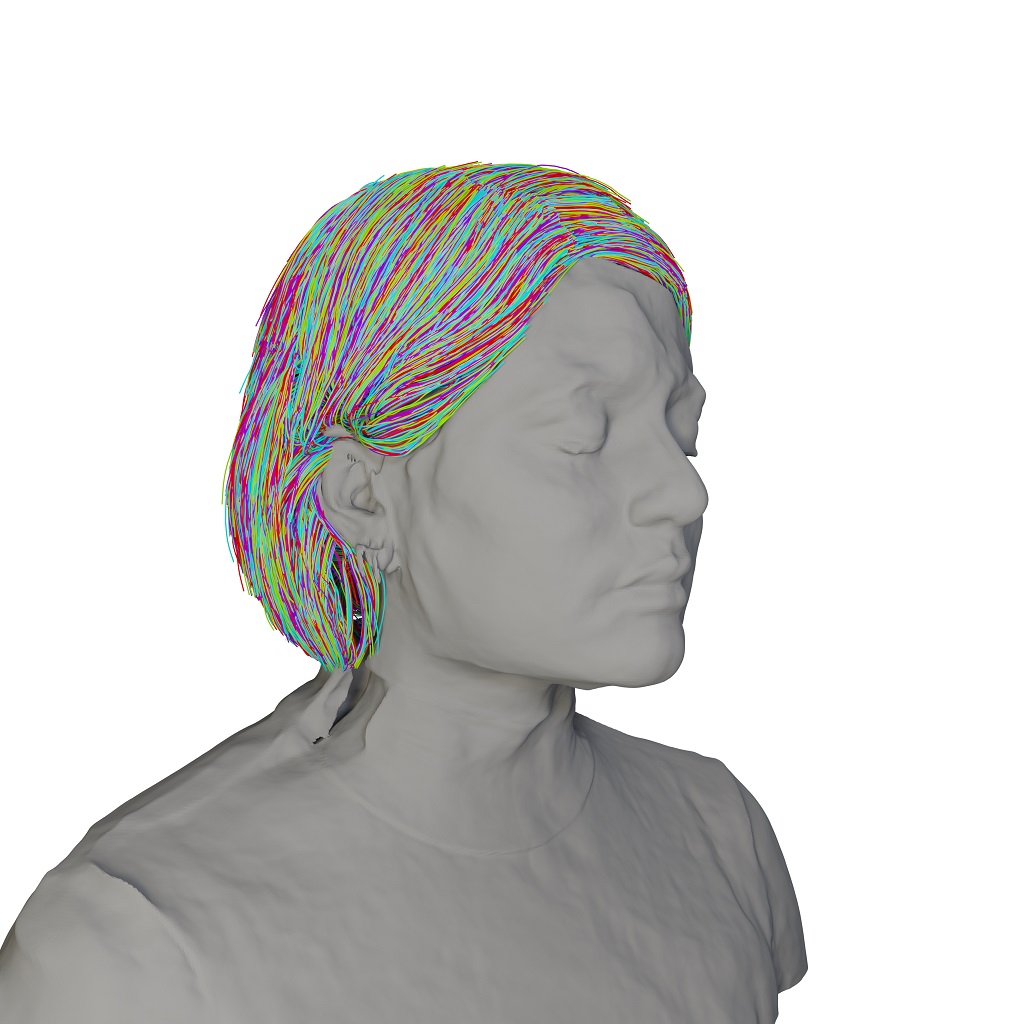} &
       \hspace{-0.31cm} 
        \includegraphics[width=0.24\textwidth]{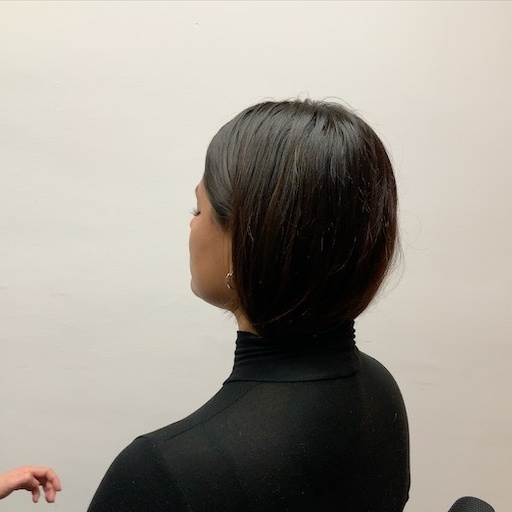} & \hspace{-0.31cm} 
        \includegraphics[width=0.24\textwidth]{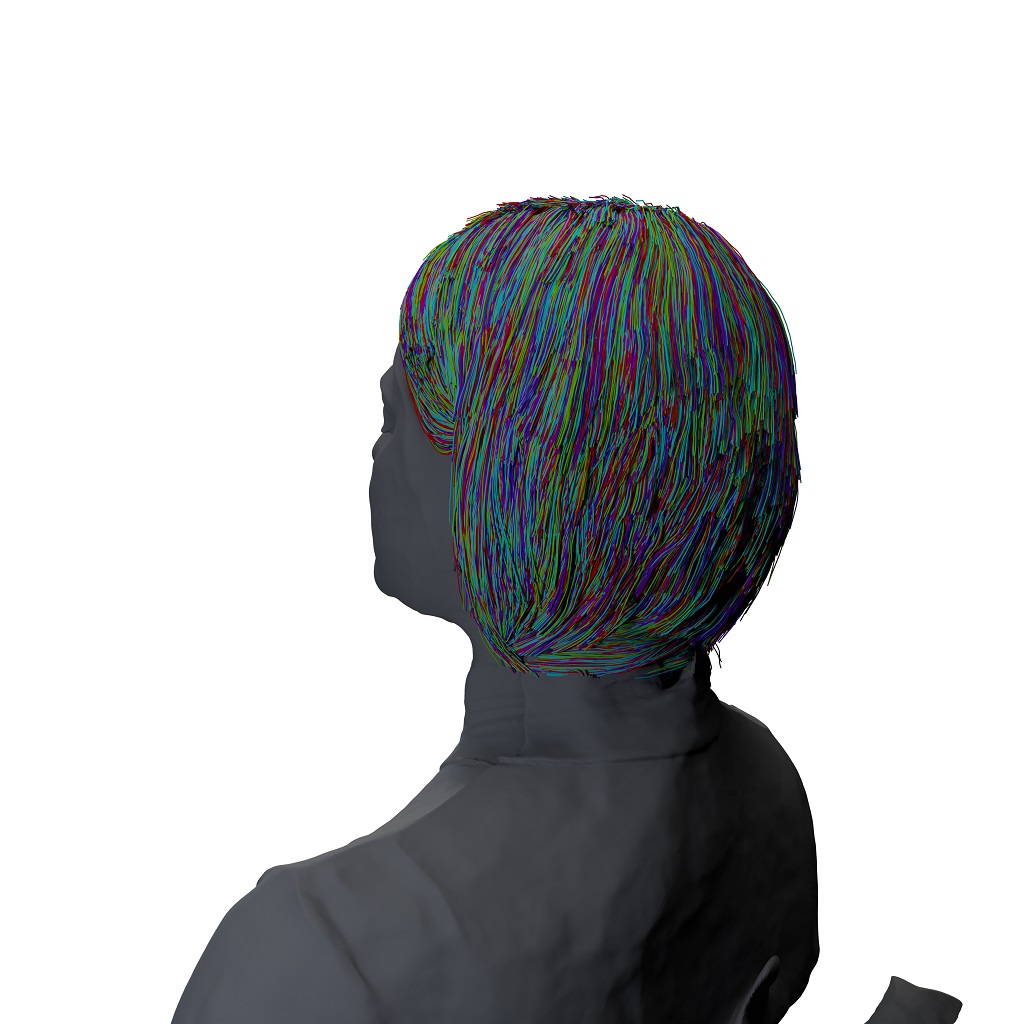} \\
        \textbf{Image}  & \hspace{-0.31cm} \textbf{Ours}& \hspace{-0.31cm} \textbf{Image}  & \hspace{-0.31cm} \textbf{Ours} 
    \end{tabular}
    \caption{Additional results for our method on a multi-view dataset~\cite{Ramon2021H3DNetFH}. Our method is capable of reconstructing various length hairstyles, starting from long (top row) to short (bottom row).  Digital zoom-in is recommended.}
    \label{fig:geom_compare_suppmat_additional}
\end{figure*}
\clearpage

%% file: figures_suppmat/colmap_jenya_last/colmap_jenya.tex
\begin{figure*}
    \begin{tabular}{ccccc}
        \includegraphics[width=0.185\textwidth]{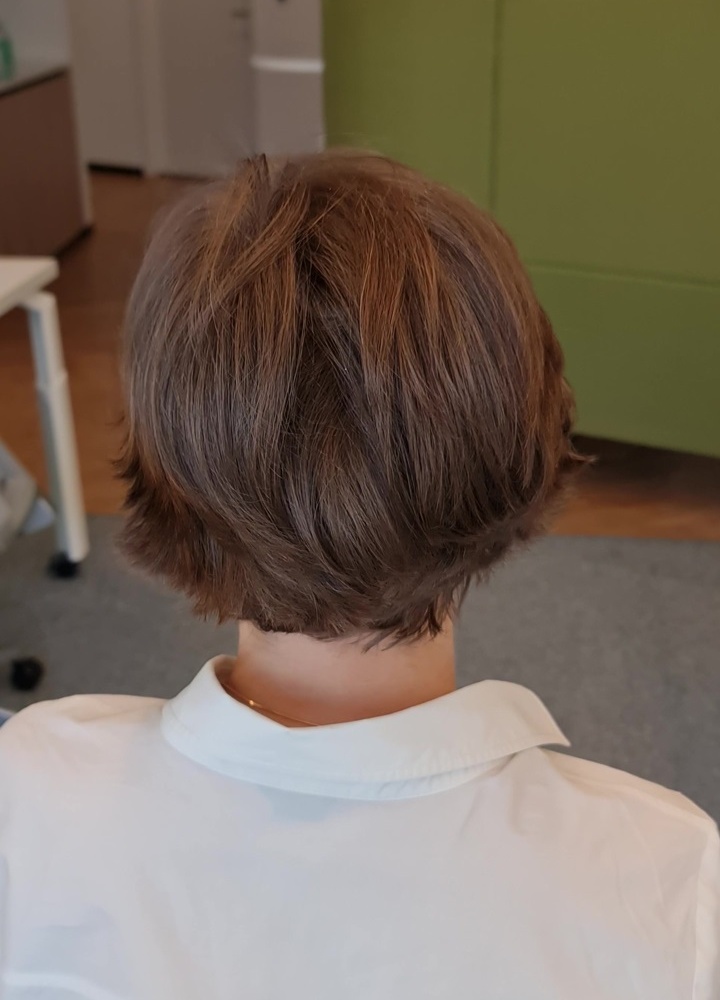} & \hspace{-0.31cm}
        \includegraphics[width=0.185\textwidth]{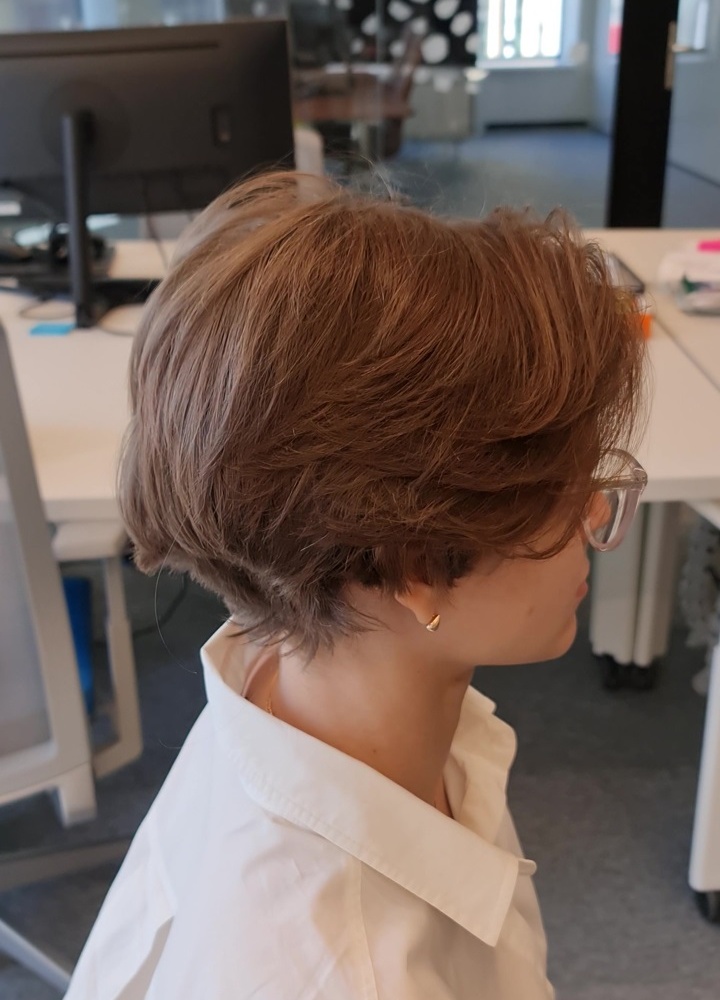} & \hspace{-0.31cm} 
        \includegraphics[width=0.185\textwidth]{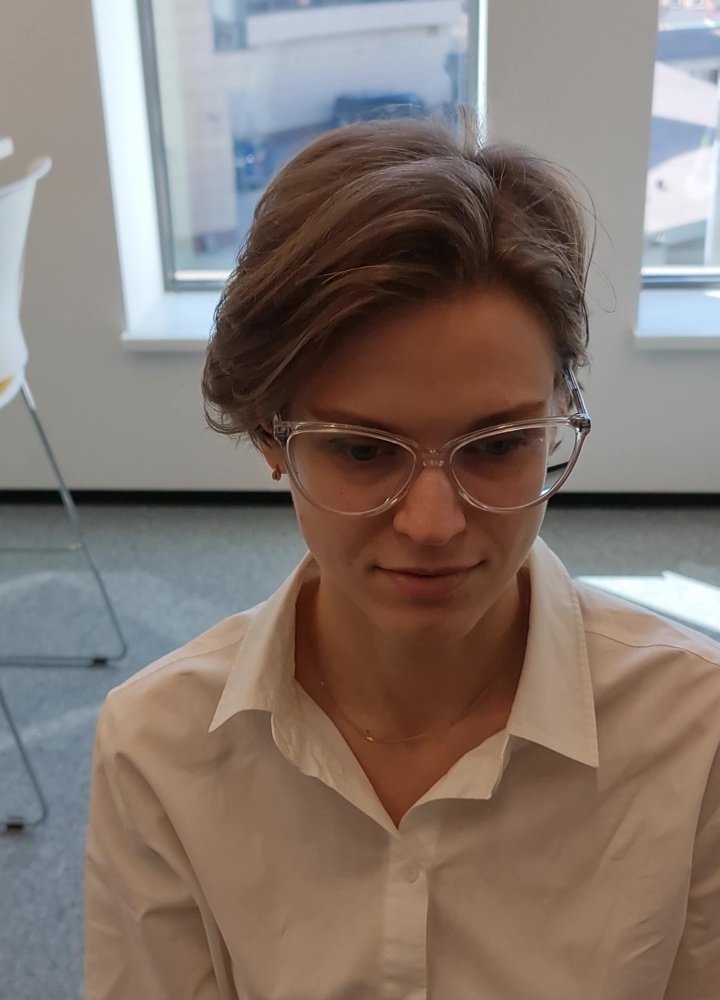} & \hspace{-0.31cm} 
        \includegraphics[width=0.185\textwidth]{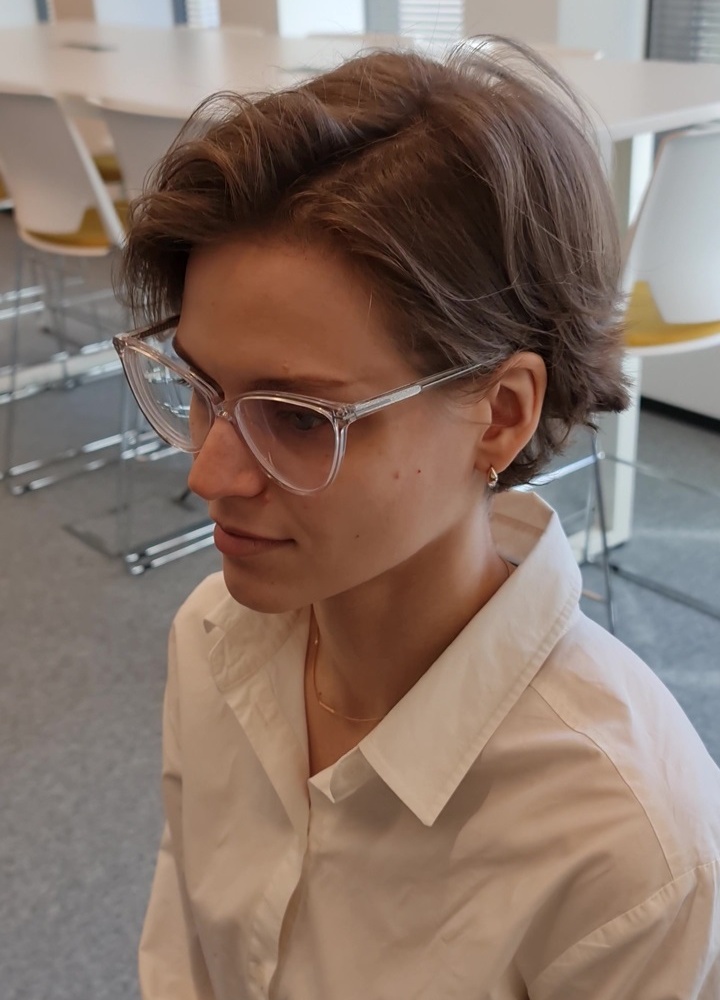} & \hspace{-0.31cm} 
        \includegraphics[width=0.185\textwidth]{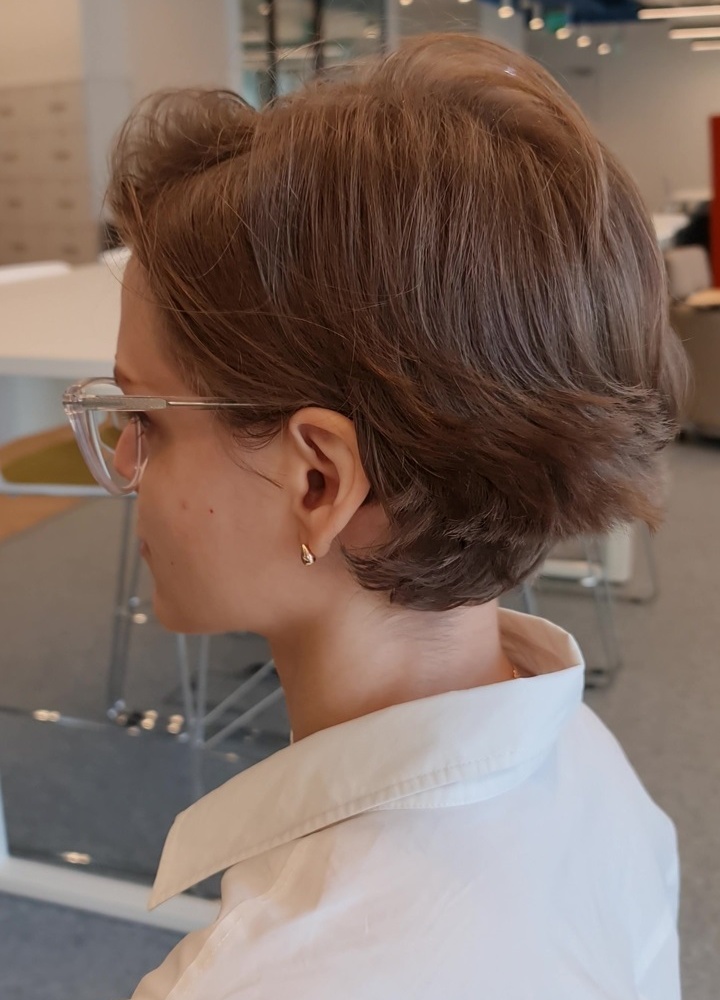} \\ %
        \includegraphics[width=0.185\textwidth]{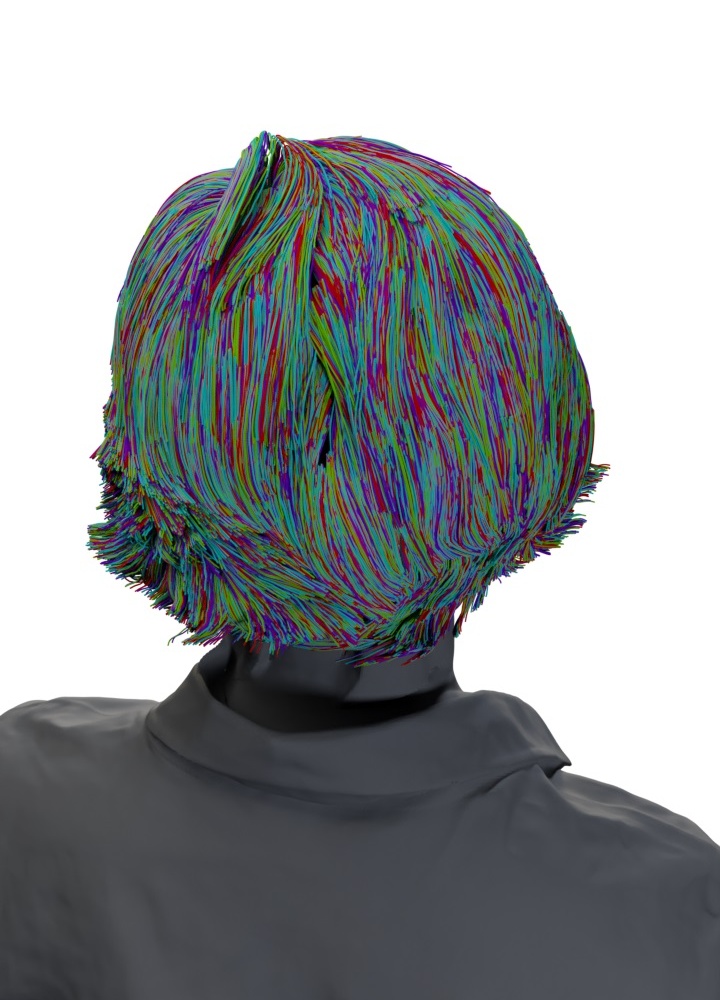} & \hspace{-0.31cm}
       \includegraphics[width=0.185\textwidth]{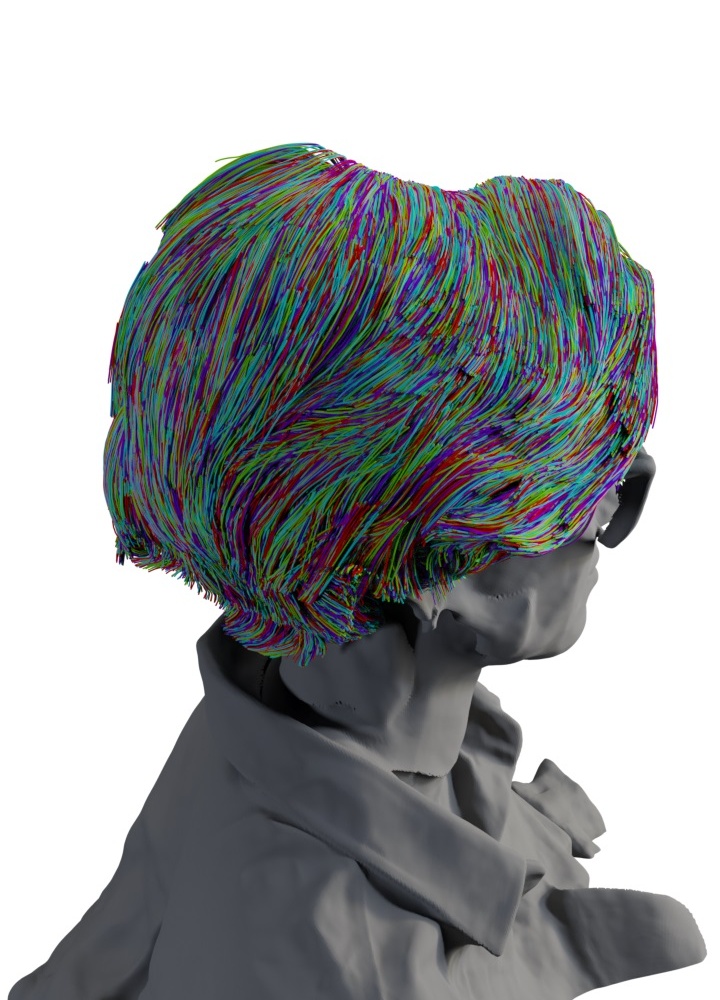} & \hspace{-0.31cm} 
        \includegraphics[width=0.185\textwidth]{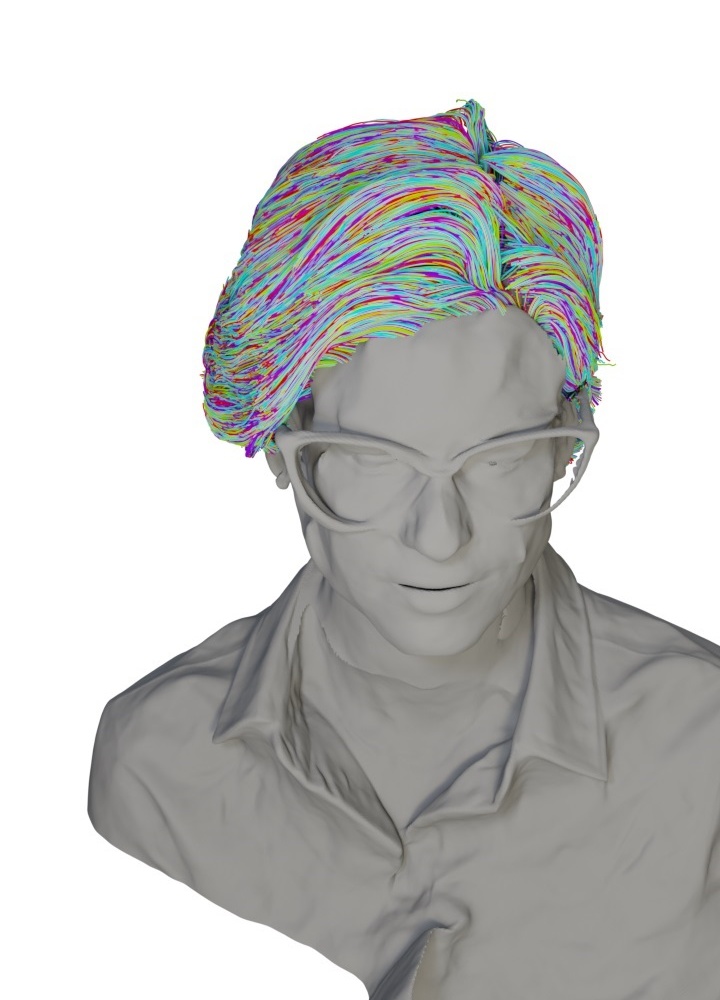} & \hspace{-0.31cm} 
        \includegraphics[width=0.185\textwidth]{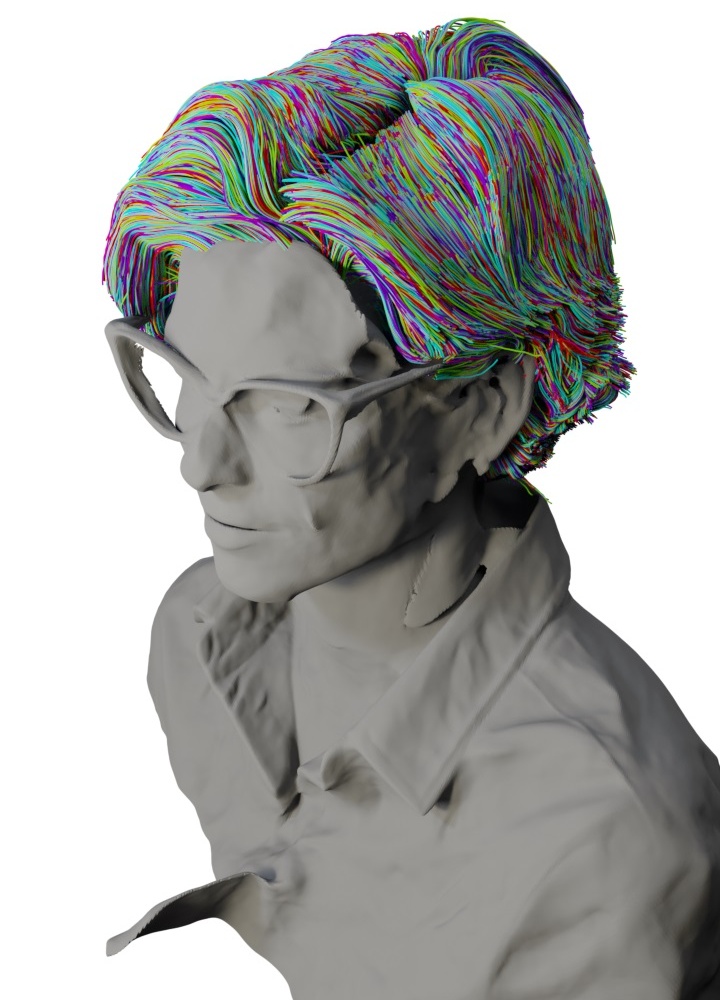} & \hspace{-0.31cm} 
        \includegraphics[width=0.185\textwidth]{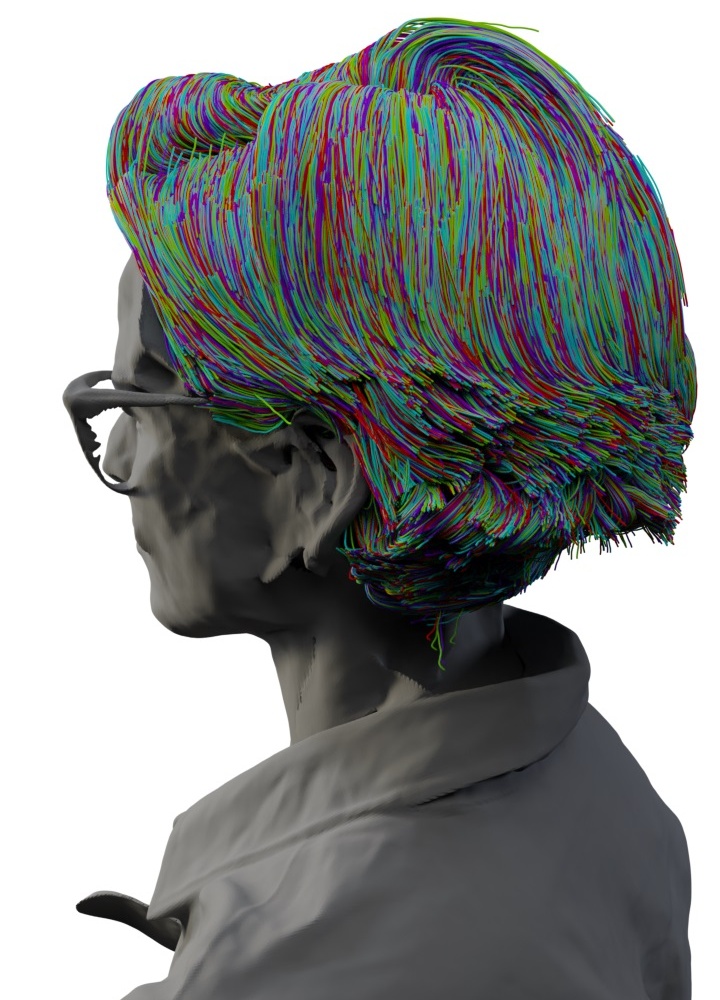}

        \\
               \includegraphics[width=0.185\textwidth]{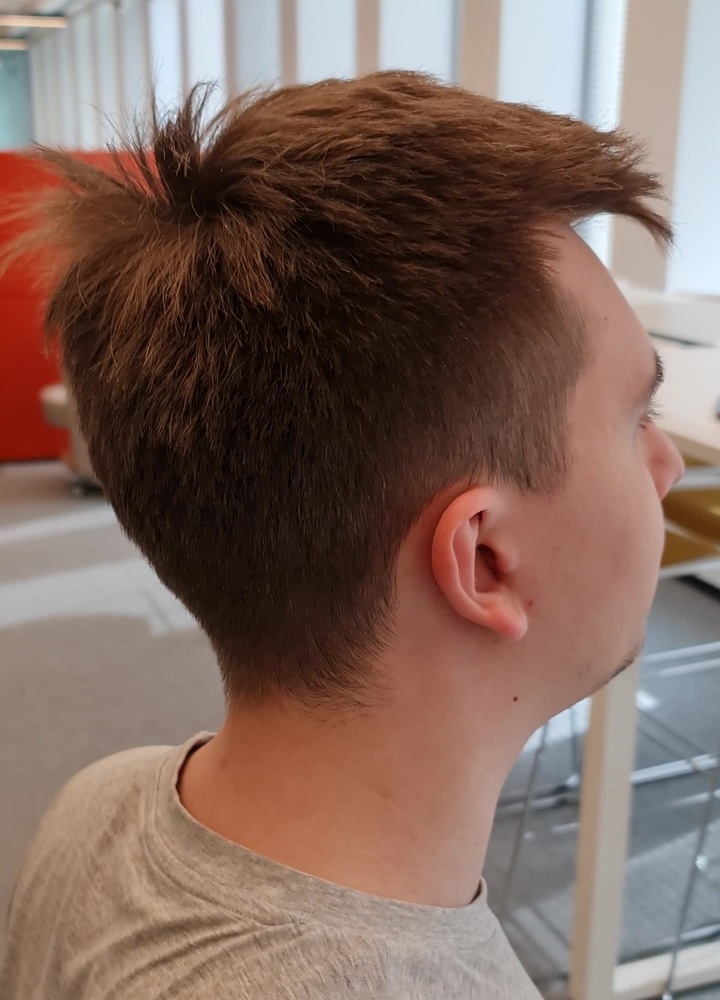} & \hspace{-0.31cm}
        \includegraphics[width=0.185\textwidth]{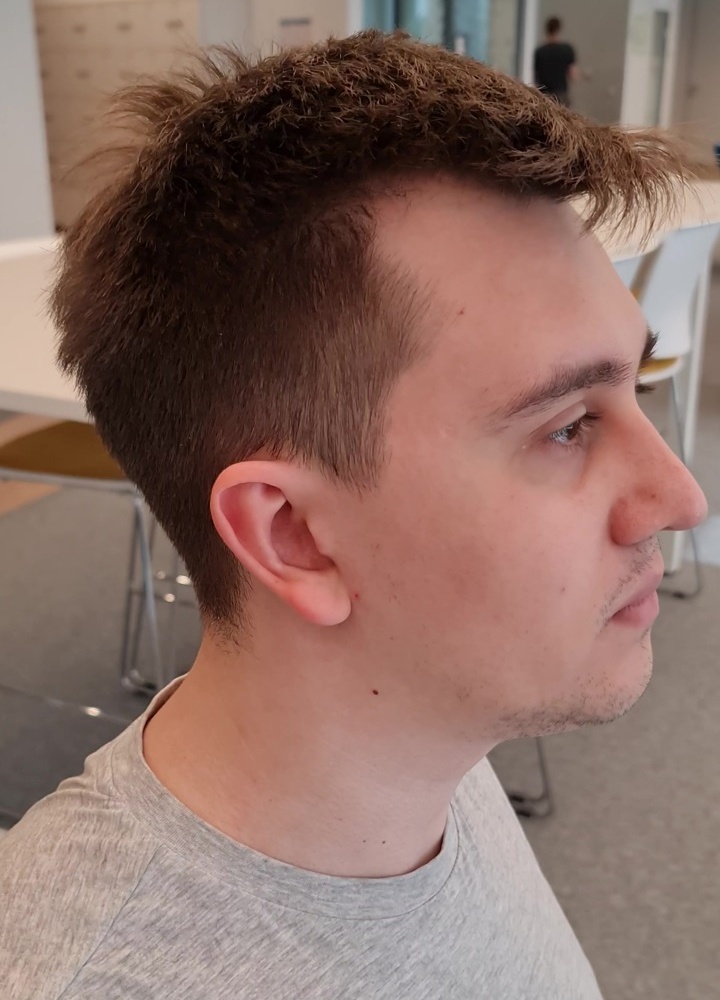} & \hspace{-0.31cm} 
        \includegraphics[width=0.185\textwidth]{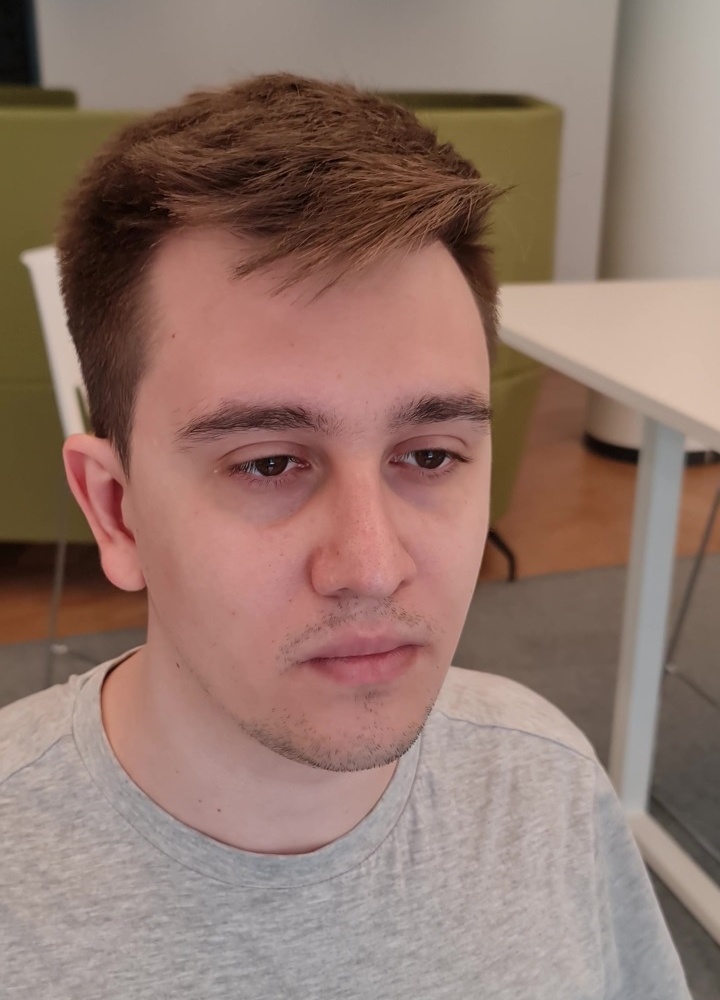} & \hspace{-0.31cm} 
        \includegraphics[width=0.185\textwidth]{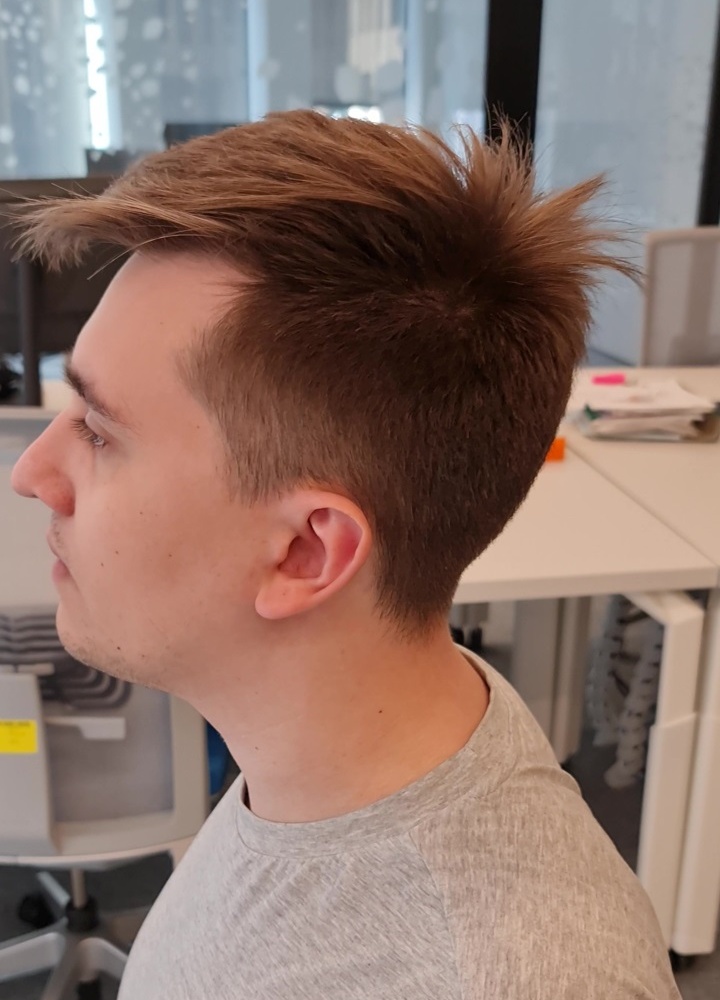} & \hspace{-0.31cm} 
        \includegraphics[width=0.185\textwidth]{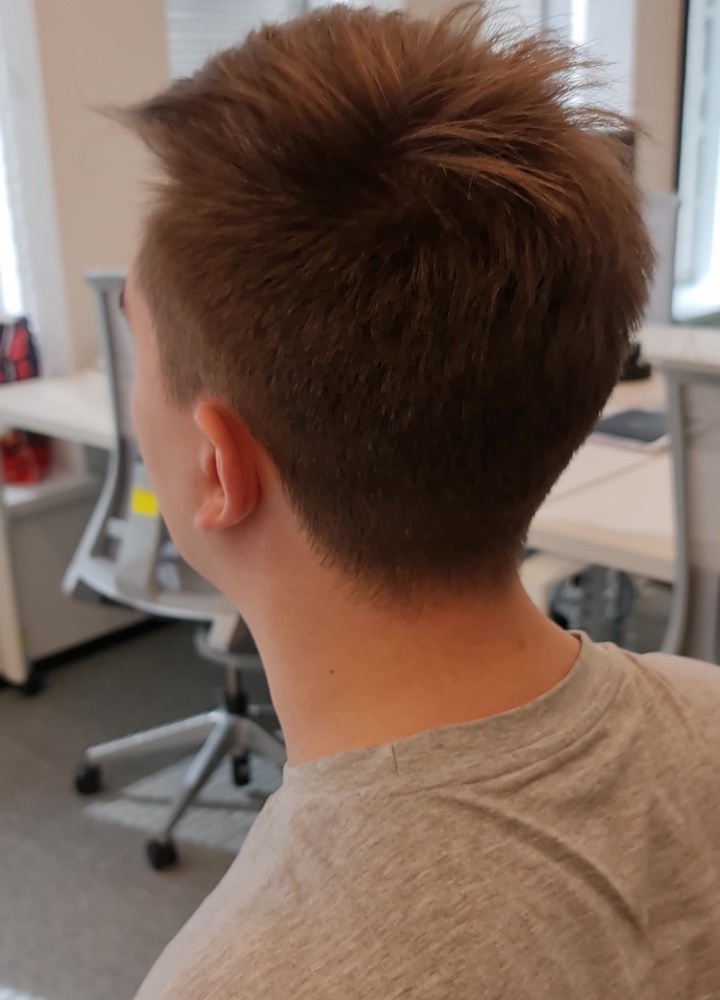} \\ %
        \includegraphics[width=0.185\textwidth]{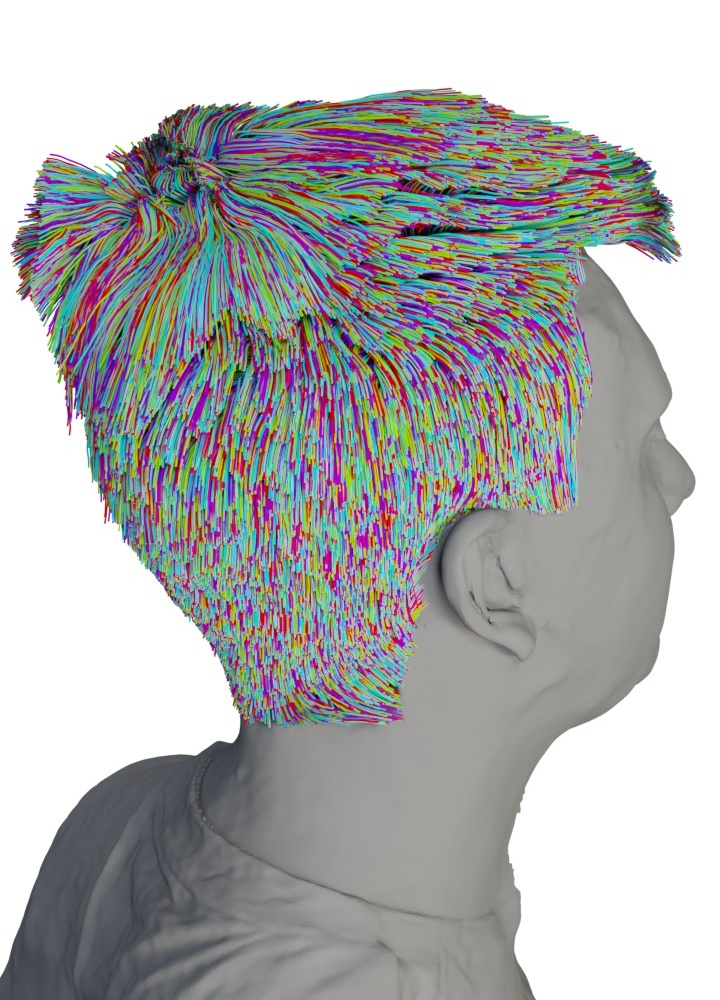} & \hspace{-0.31cm}
       \includegraphics[width=0.185\textwidth]{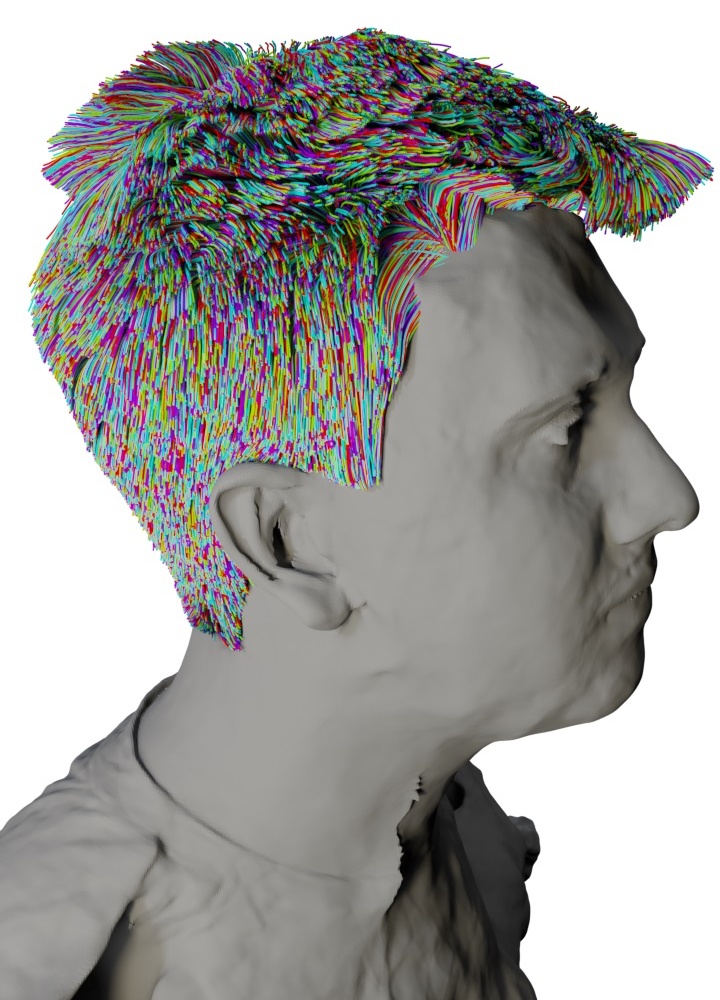} & \hspace{-0.31cm} 
        \includegraphics[width=0.185\textwidth]{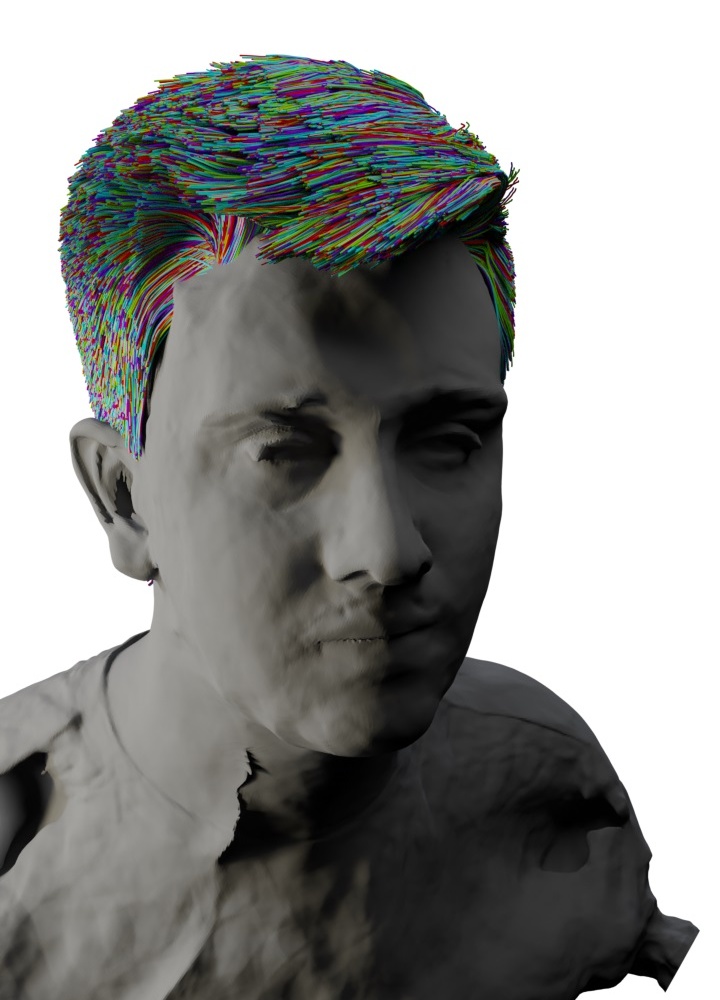} & \hspace{-0.31cm} 
        \includegraphics[width=0.185\textwidth]{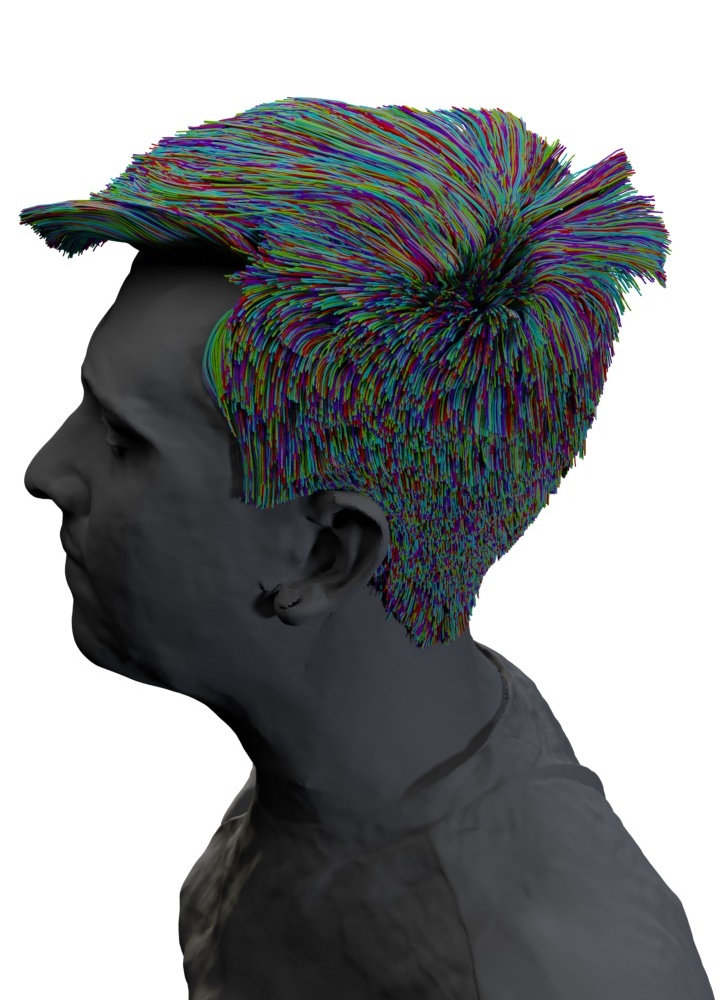} & \hspace{-0.31cm} 
        \includegraphics[width=0.185\textwidth]{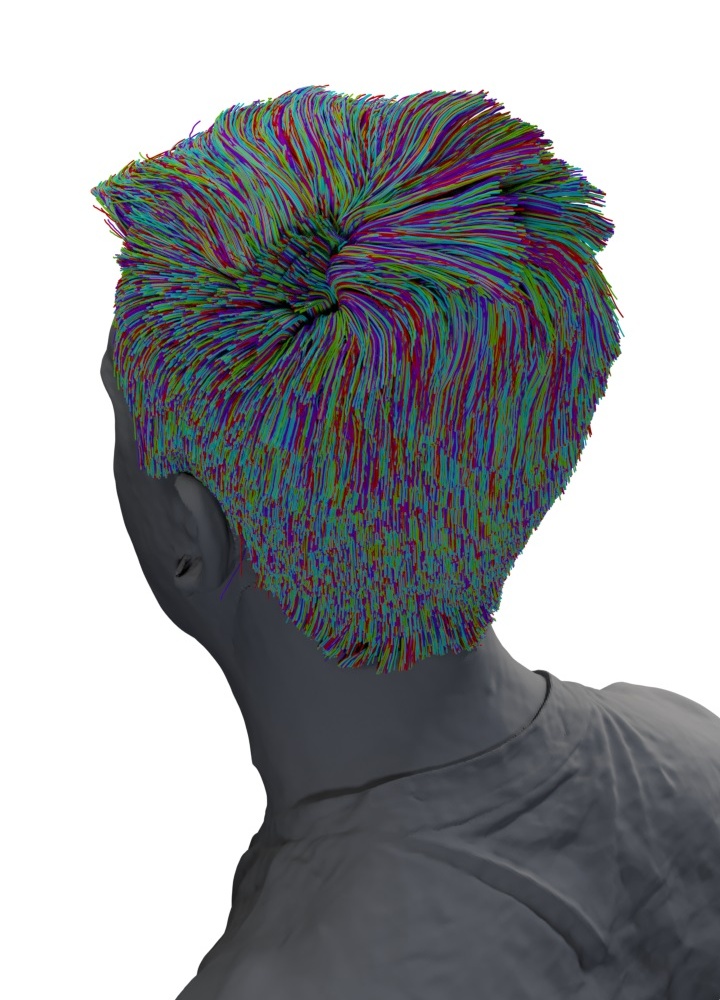}
        
    \end{tabular}
    \caption{Additional reconstruction results of our method on monocular videos in arbitrary lighting conditions. Our method is capable of obtaining personalized reconstructions for various hairstyles. } 
    \label{fig:colmap_scene_jenya_suppmat}
\end{figure*}
\clearpage

%% file: figures_suppmat/colmap_ksyusha_last/colmap_ksyusha_additional.tex
\begin{figure*}
    \begin{tabular}{ccccc}
        \includegraphics[width=0.185\textwidth]{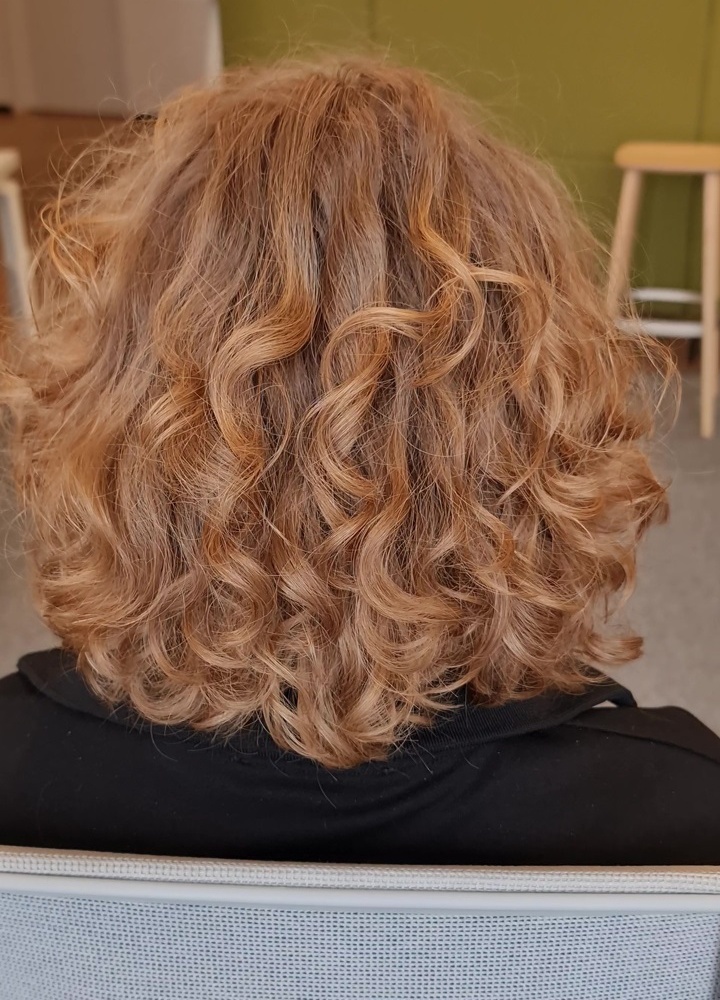} & \hspace{-0.31cm}
        \includegraphics[width=0.185\textwidth]{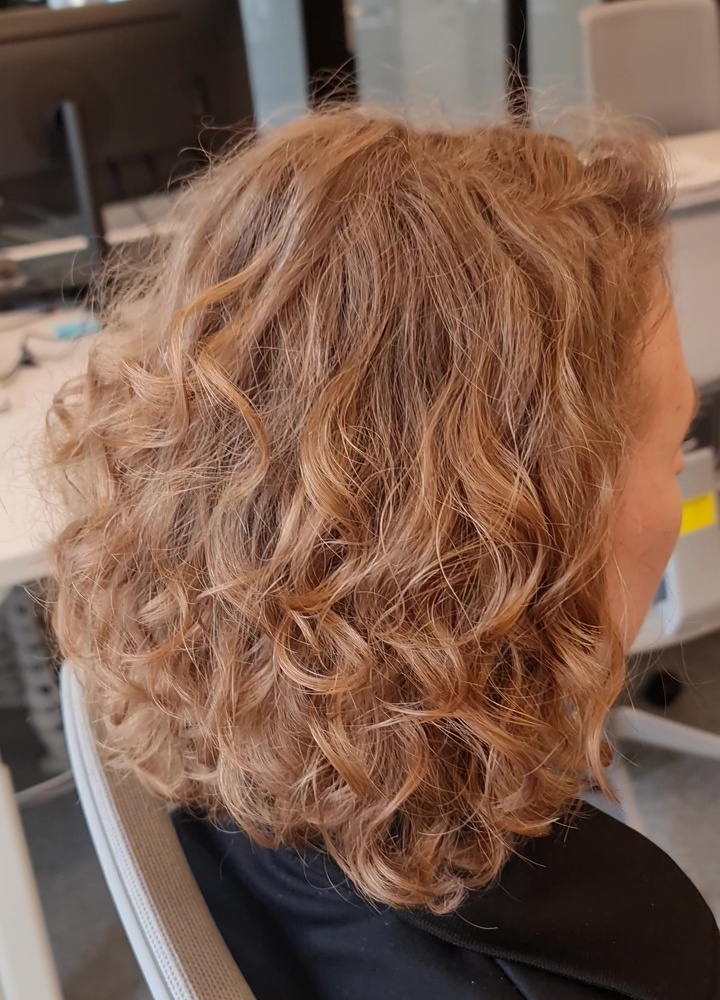} & \hspace{-0.31cm} 
        \includegraphics[width=0.185\textwidth]{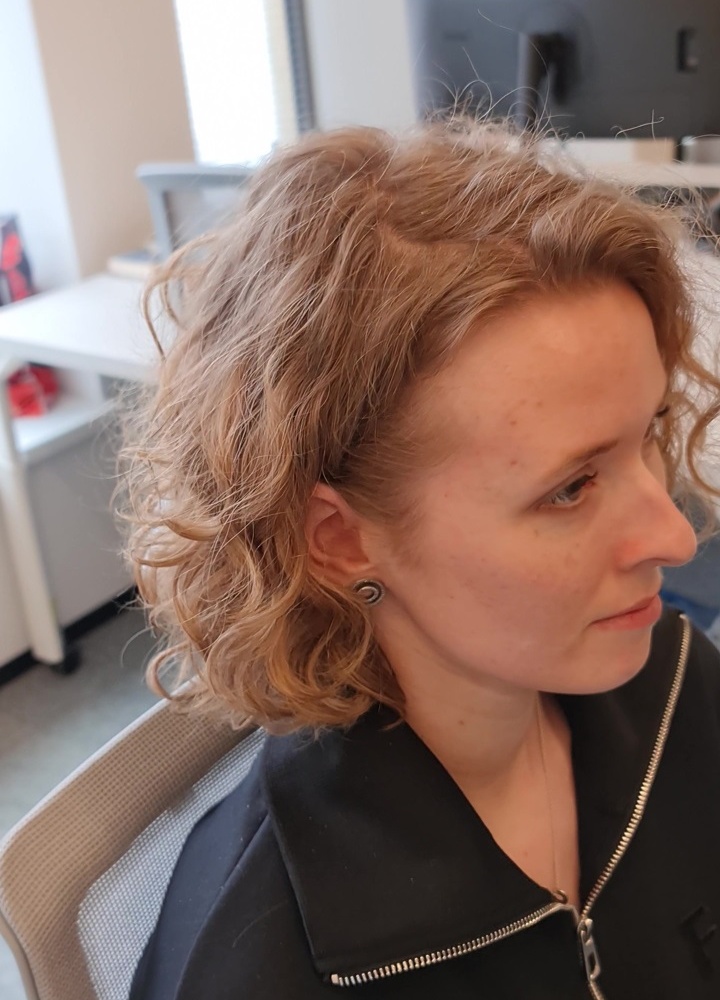} & \hspace{-0.31cm} 
        \includegraphics[width=0.185\textwidth]{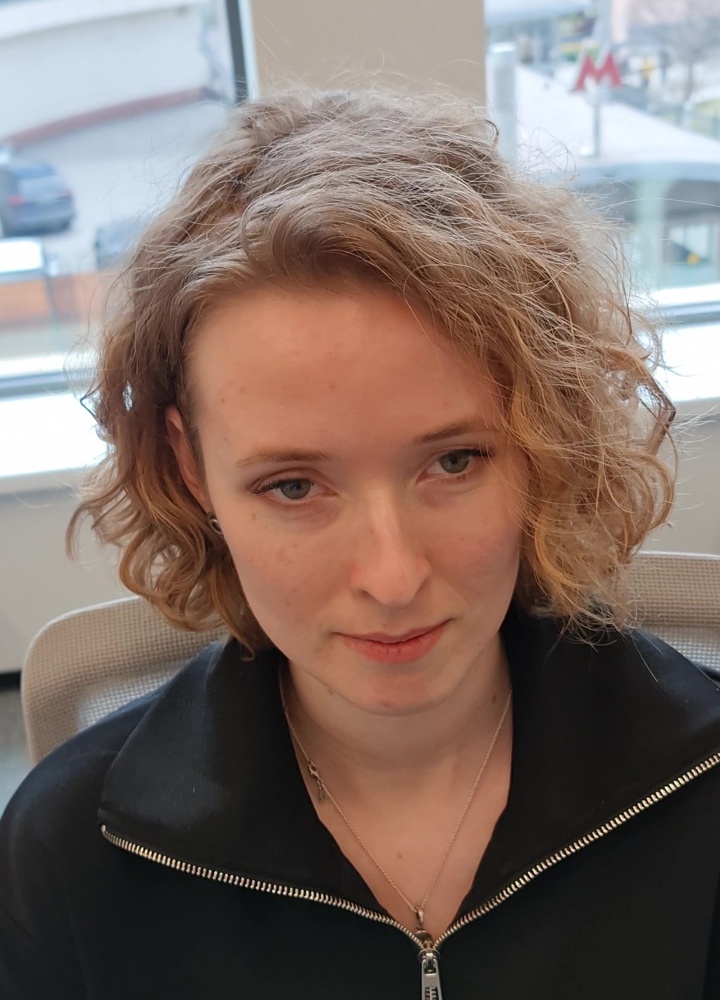} & \hspace{-0.31cm} 
        \includegraphics[width=0.185\textwidth]{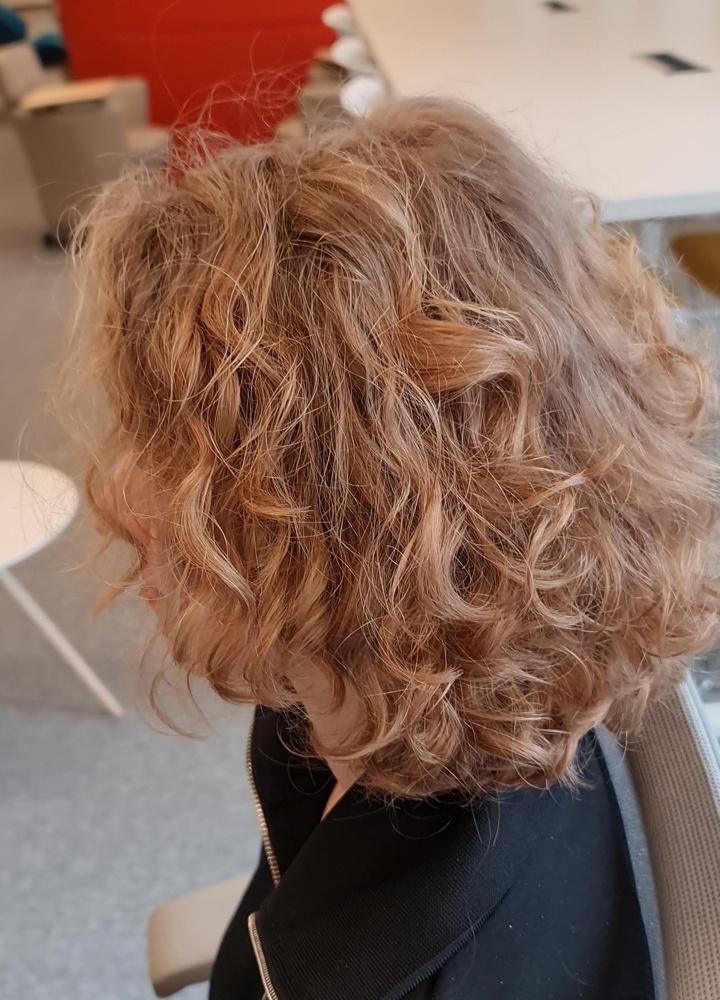} \\ %
        \includegraphics[width=0.185\textwidth]{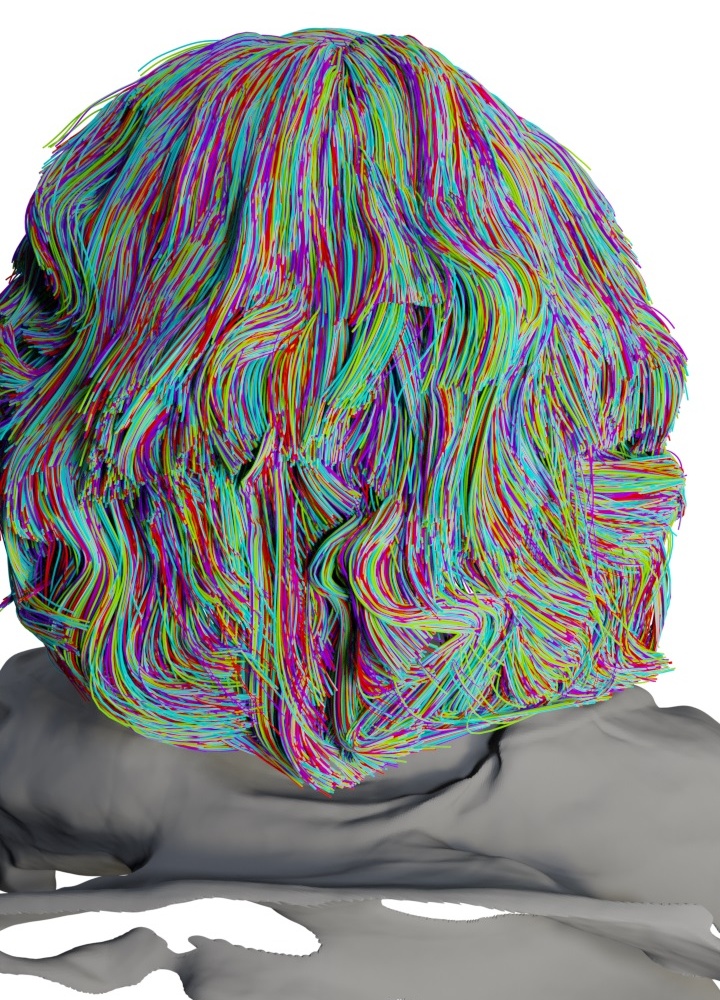} & \hspace{-0.31cm}
       \includegraphics[width=0.185\textwidth]{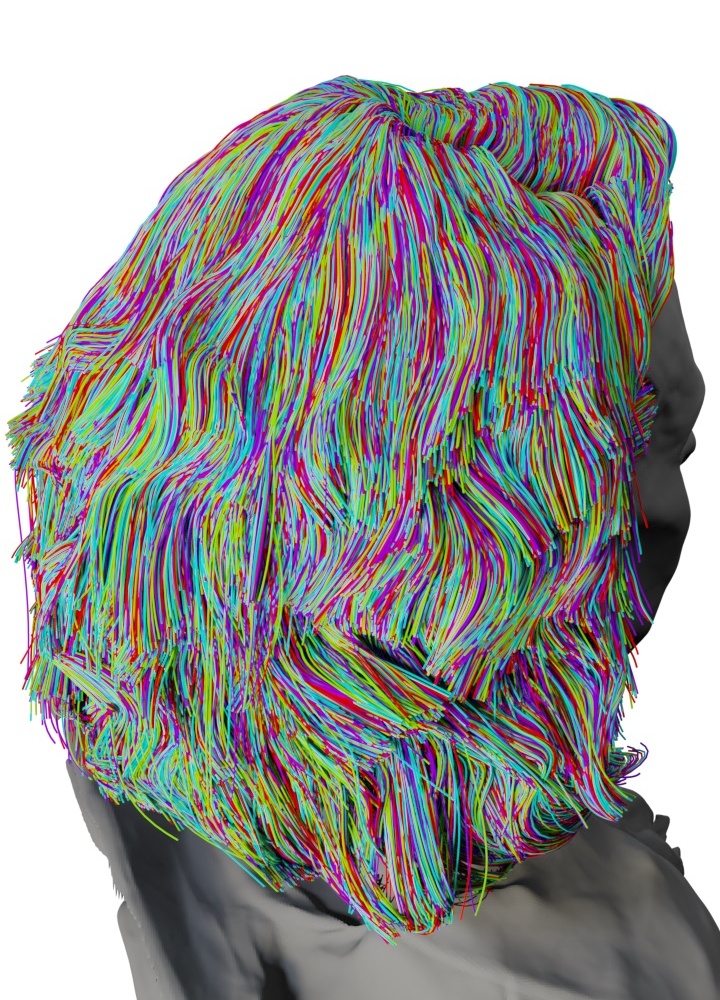} & \hspace{-0.31cm} 
        \includegraphics[width=0.185\textwidth]{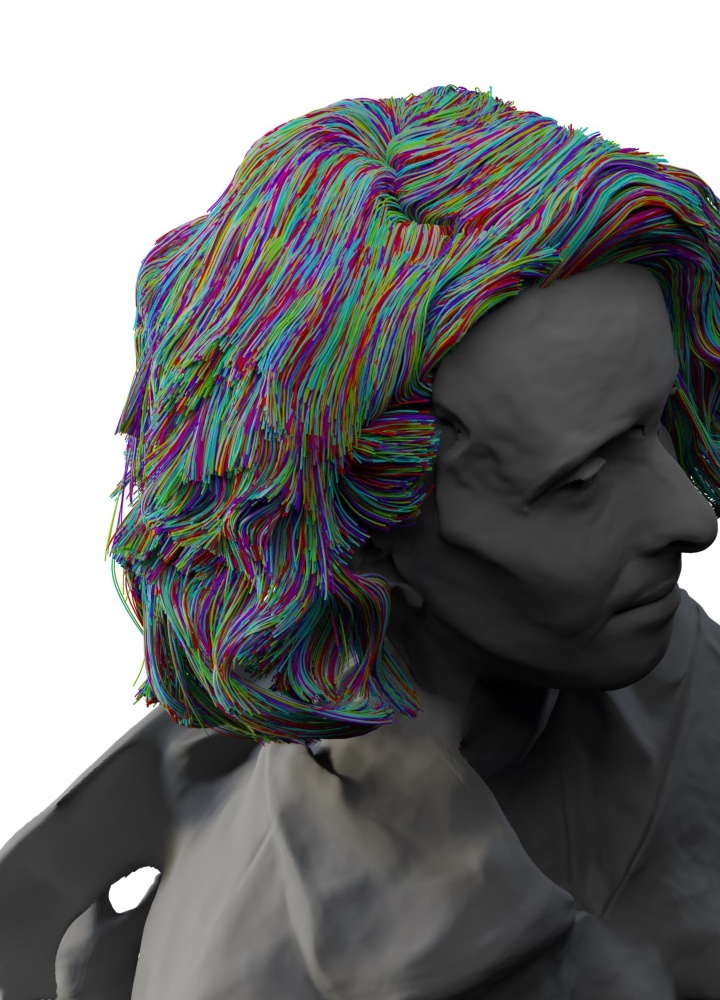} & \hspace{-0.31cm} 
        \includegraphics[width=0.185\textwidth]{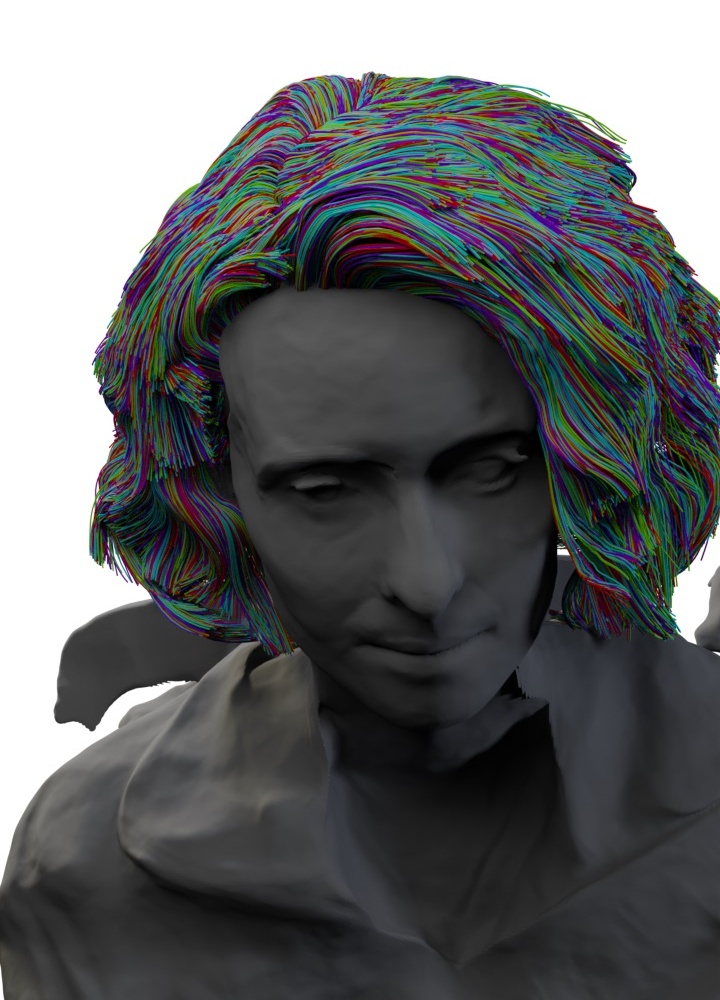} & \hspace{-0.31cm} 
        \includegraphics[width=0.185\textwidth]{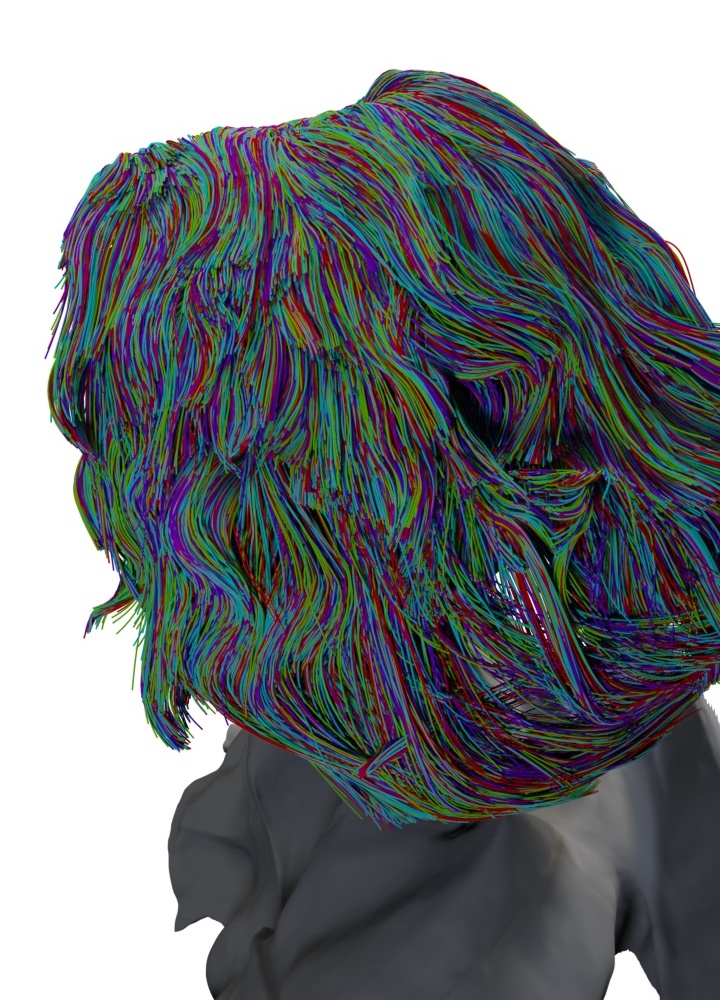}

        \\
               \includegraphics[width=0.185\textwidth]{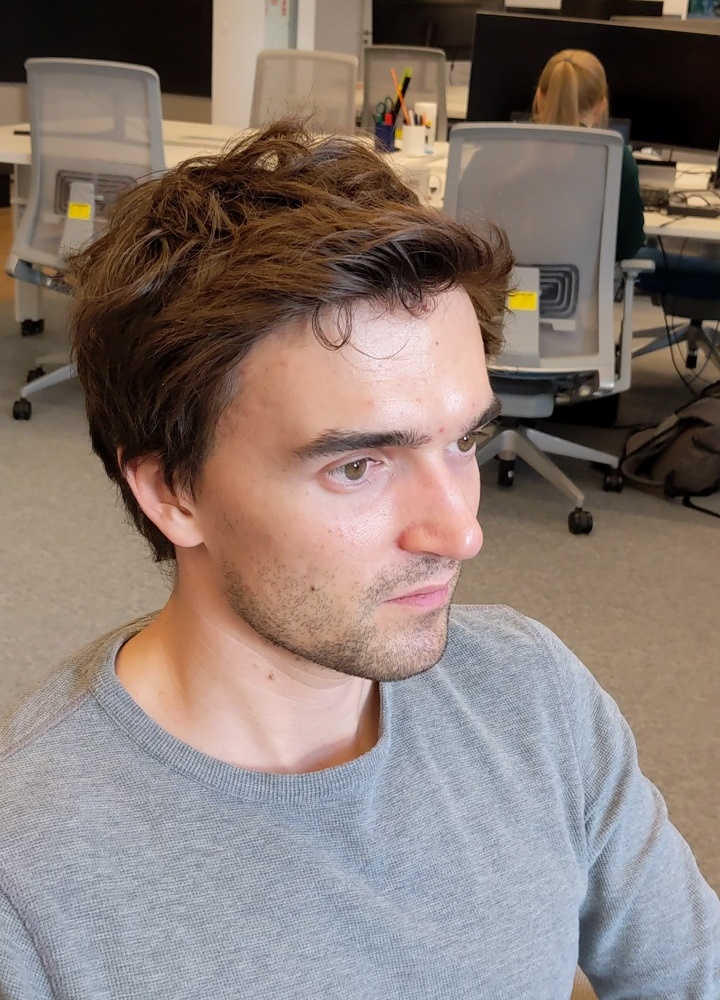} & \hspace{-0.31cm}
        \includegraphics[width=0.185\textwidth]{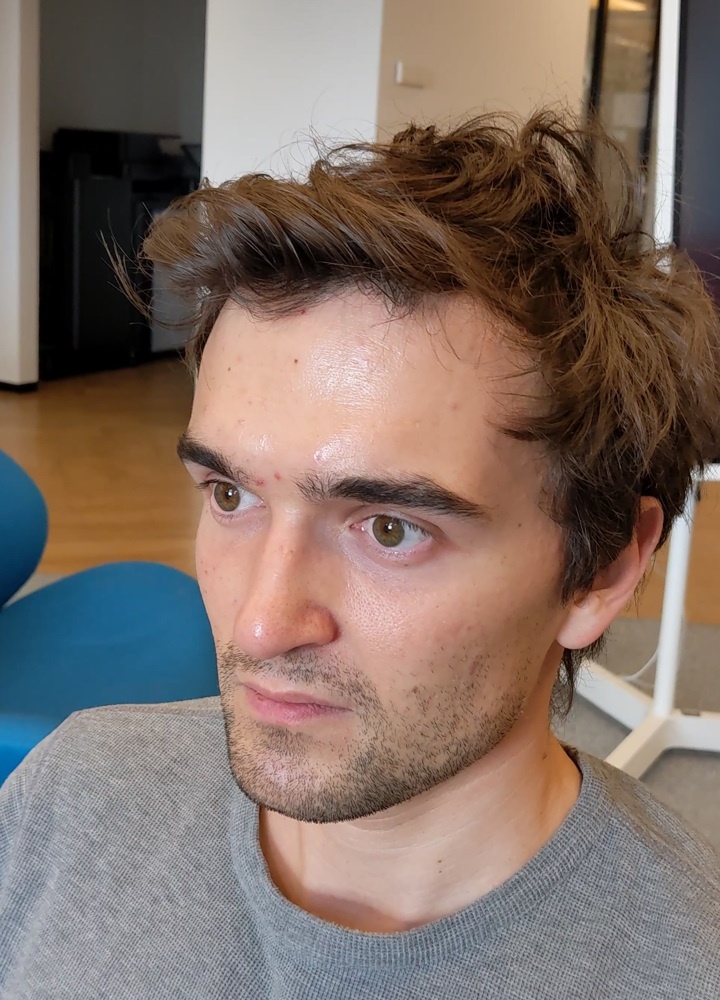} & \hspace{-0.31cm} 
        \includegraphics[width=0.185\textwidth]{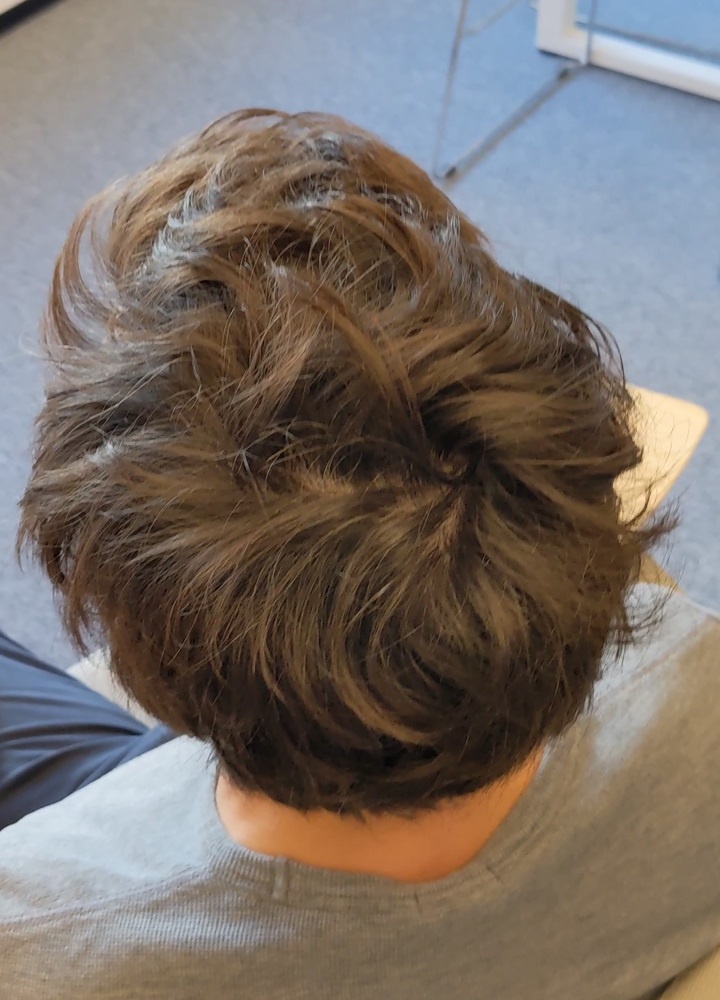} & \hspace{-0.31cm} 
        \includegraphics[width=0.185\textwidth]{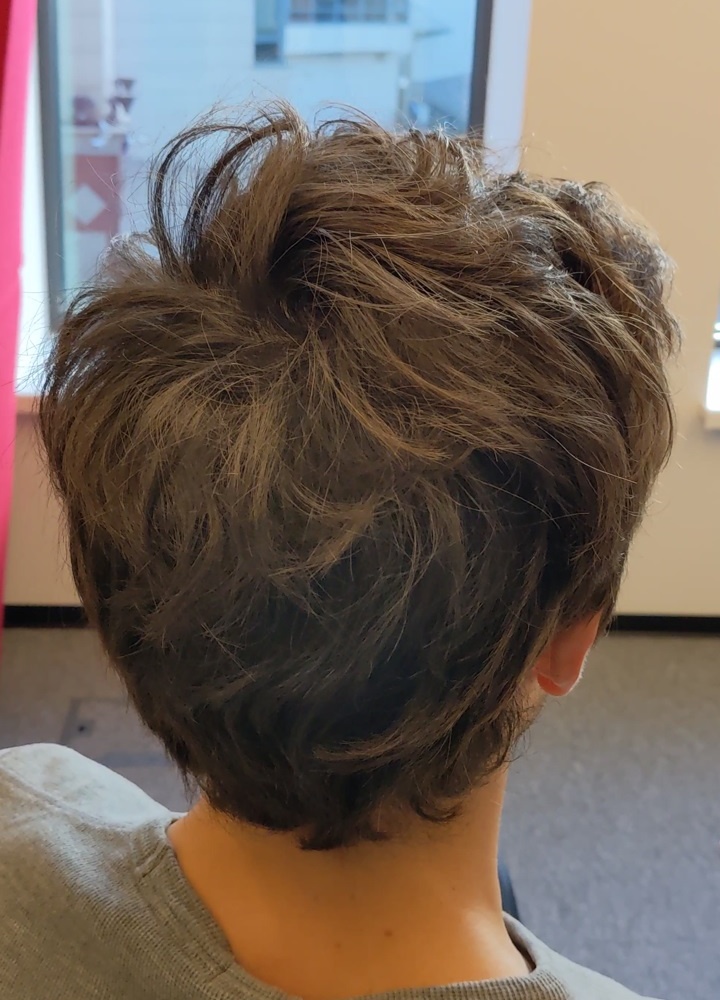} & \hspace{-0.31cm} 
        \includegraphics[width=0.185\textwidth]{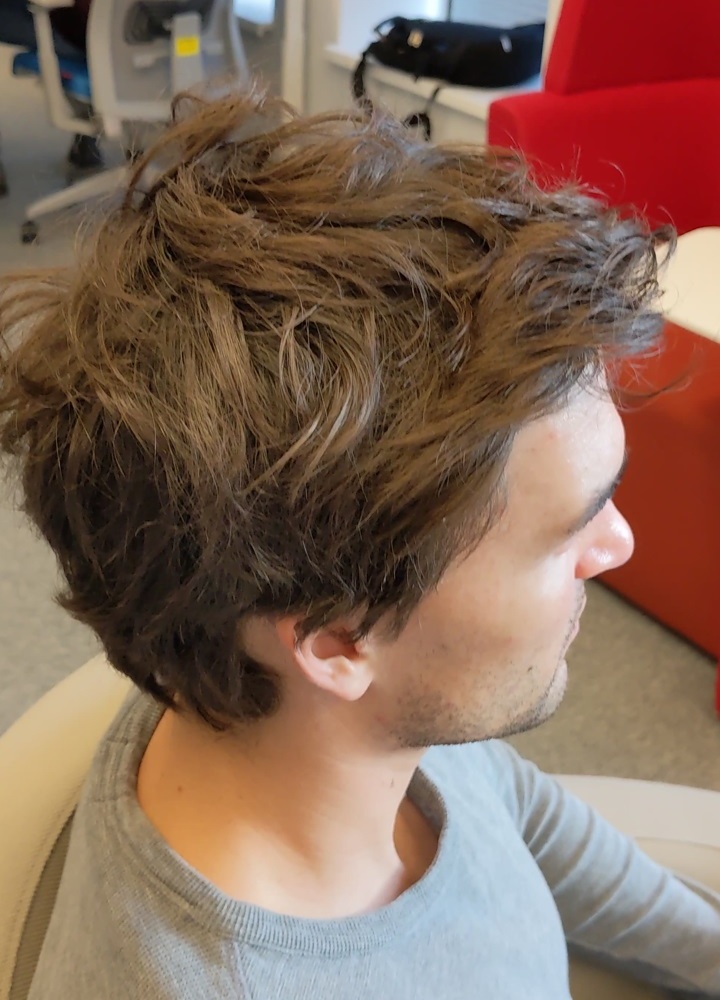} \\ %
        \includegraphics[width=0.185\textwidth]{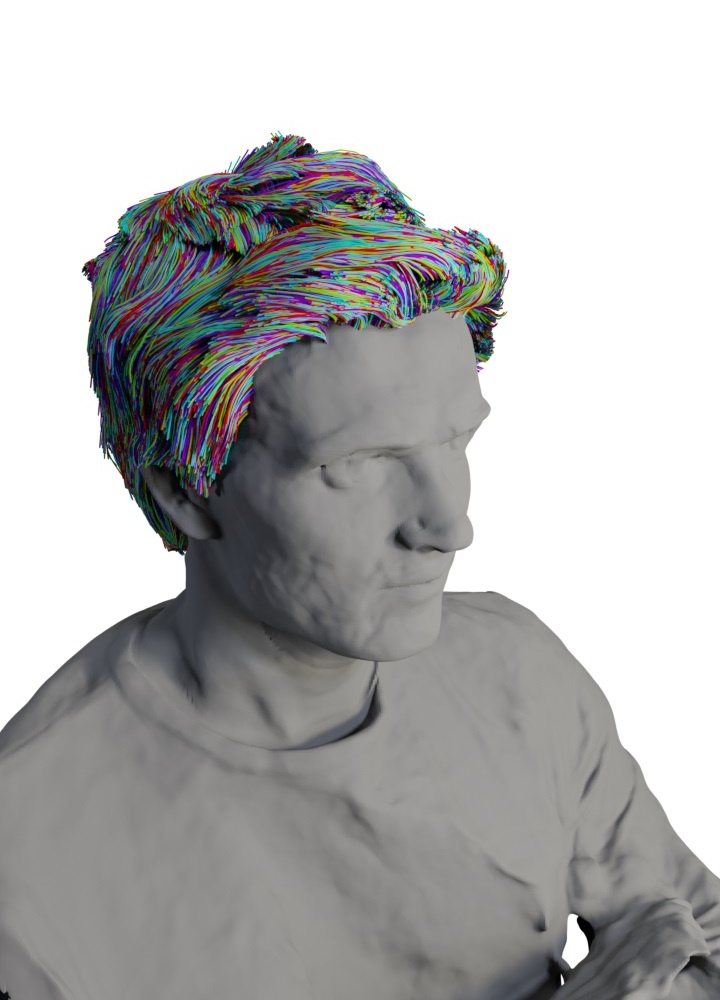} & \hspace{-0.31cm}
       \includegraphics[width=0.185\textwidth]{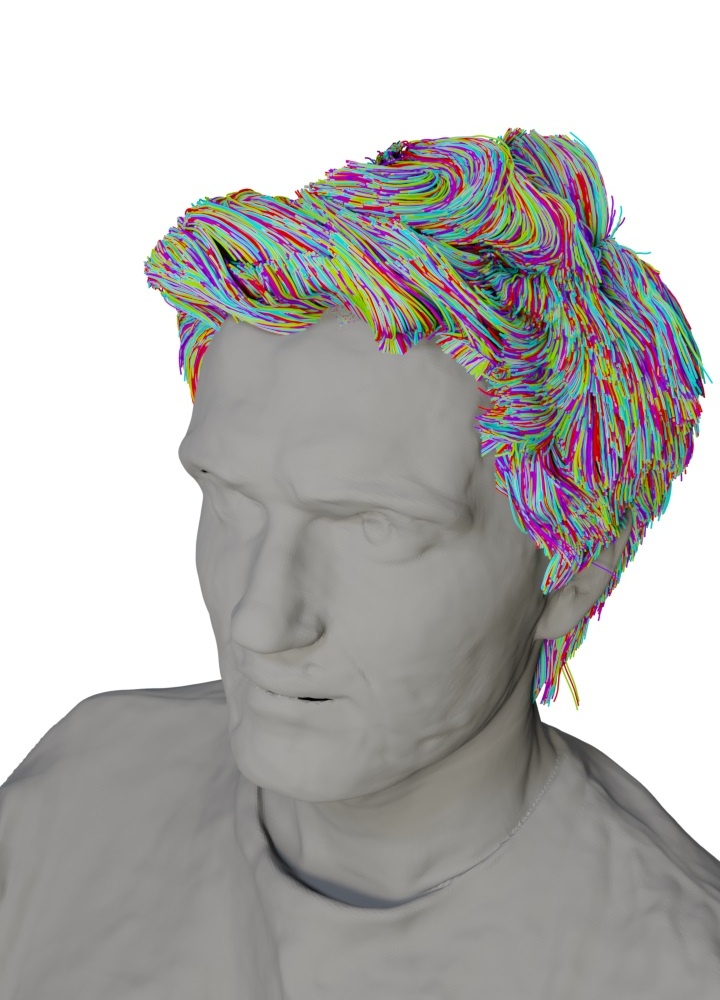} & \hspace{-0.31cm} 
        \includegraphics[width=0.185\textwidth]{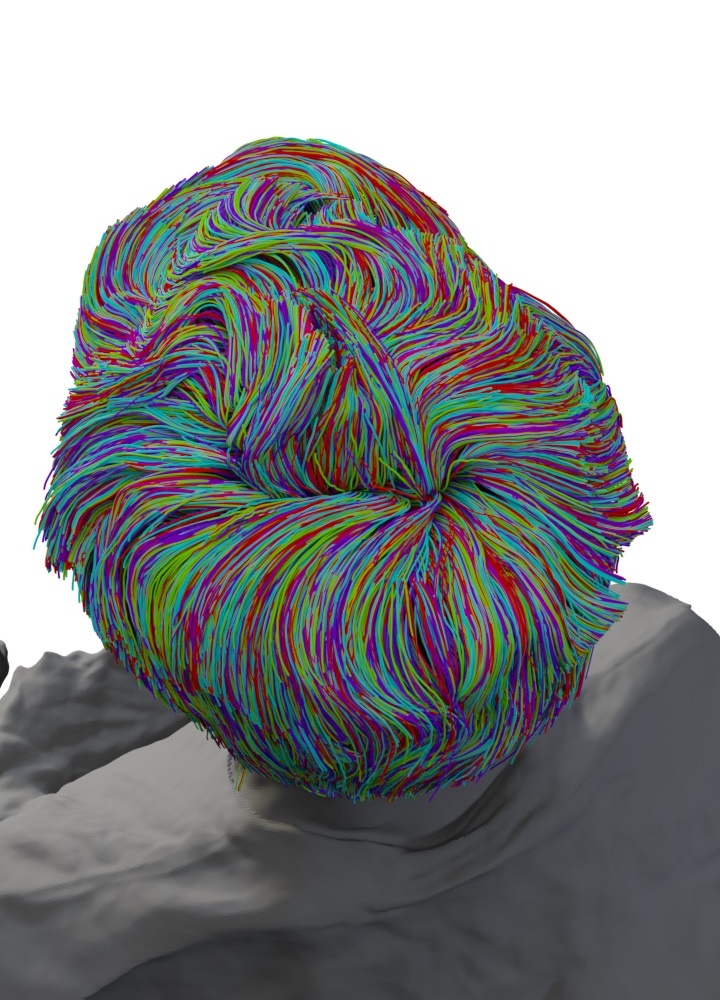} & \hspace{-0.31cm} 
        \includegraphics[width=0.185\textwidth]{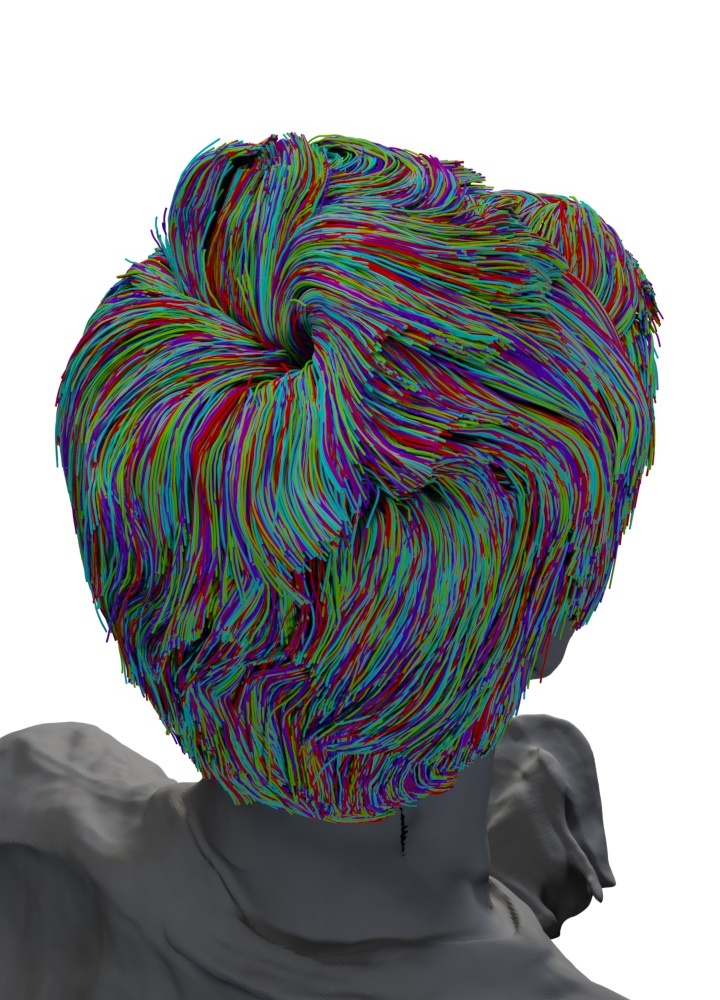} & \hspace{-0.31cm} 
        \includegraphics[width=0.185\textwidth]{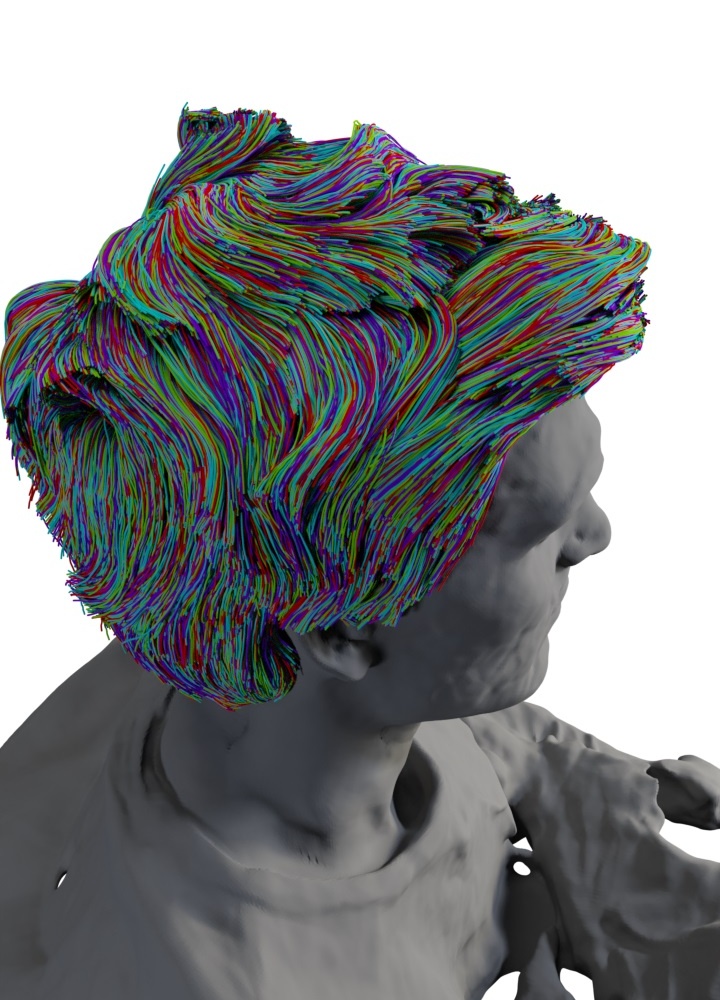}
        
    \end{tabular}
    \caption{Additional reconstruction results of our method on monocular videos in arbitrary lighting conditions. Our method could produce realistic hair geometry for long and short, straight and curly hairstyles.} 
    \label{fig:colmap_scene_ksyusha_suppmat}
\end{figure*}
\clearpage

%% file: figures_suppmat/colmap_nastya/colmap_nastya_additional.tex
\begin{figure*}
    \begin{tabular}{ccccc}

        \includegraphics[width=0.185\textwidth]{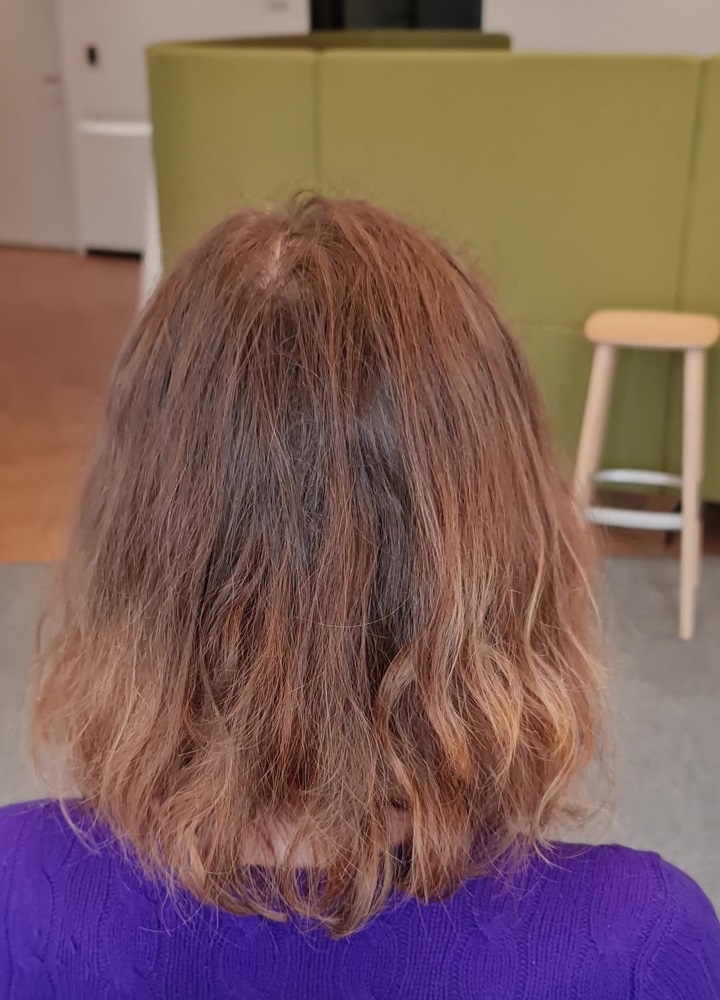} & \hspace{-0.31cm}
        \includegraphics[width=0.185\textwidth]{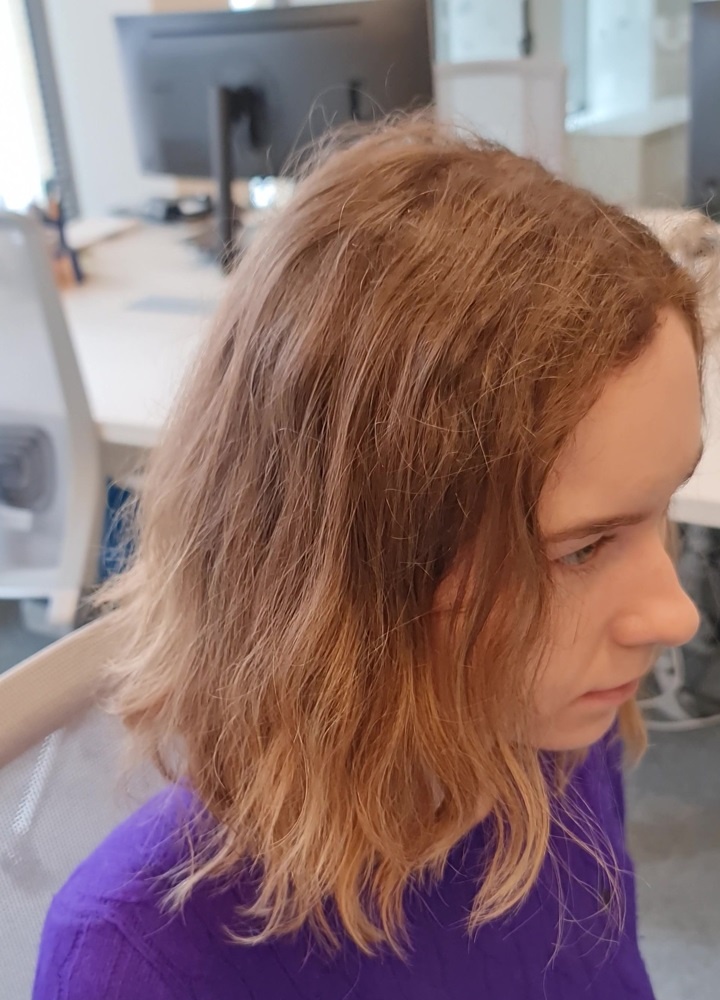} & \hspace{-0.31cm} 
        \includegraphics[width=0.185\textwidth]{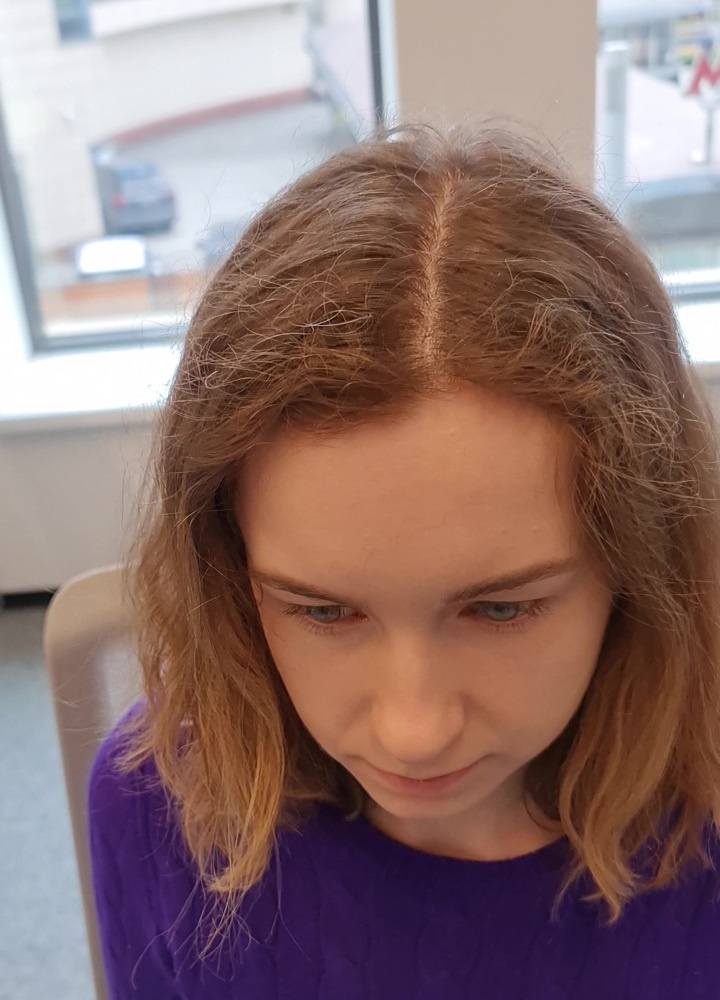} & \hspace{-0.31cm} 
        \includegraphics[width=0.185\textwidth]{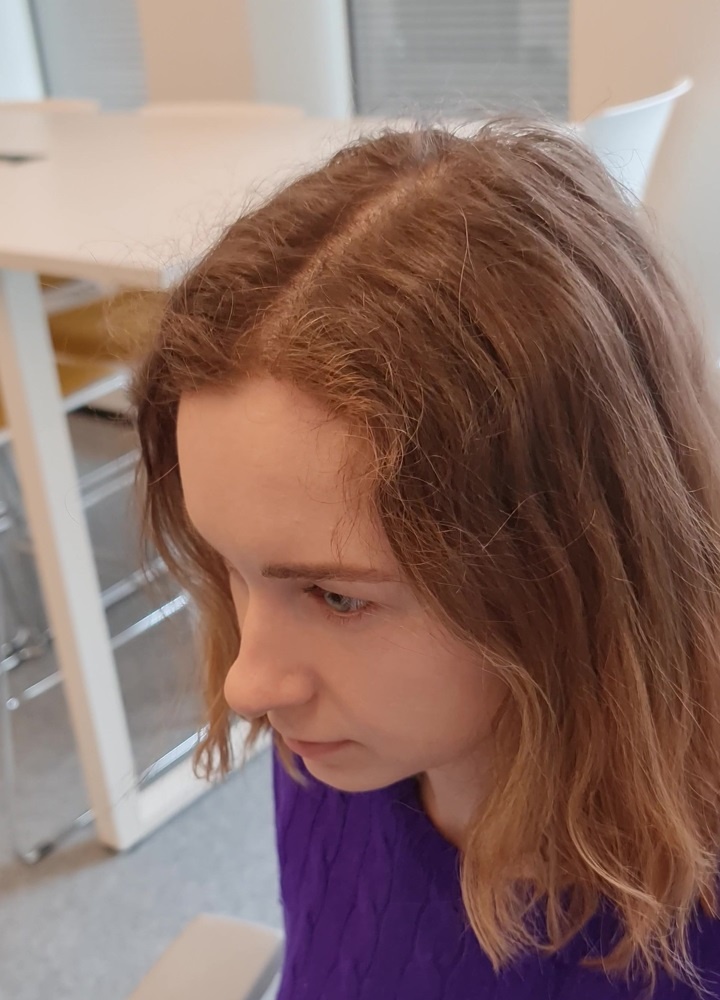} & \hspace{-0.31cm} 
        \includegraphics[width=0.185\textwidth]{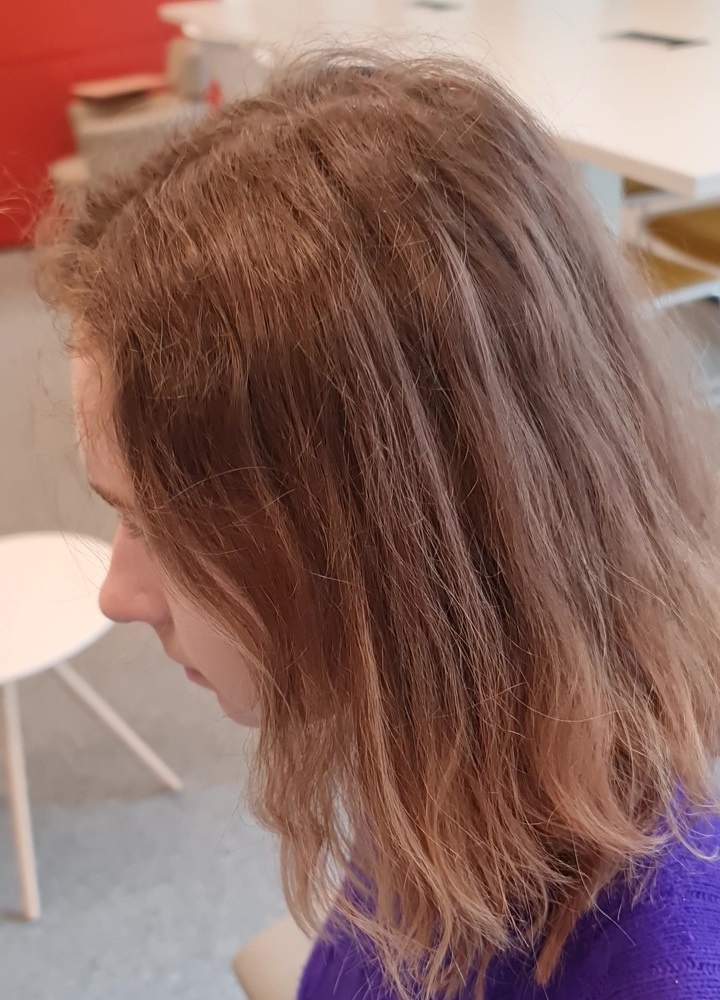} \\ %
        \includegraphics[width=0.185\textwidth]{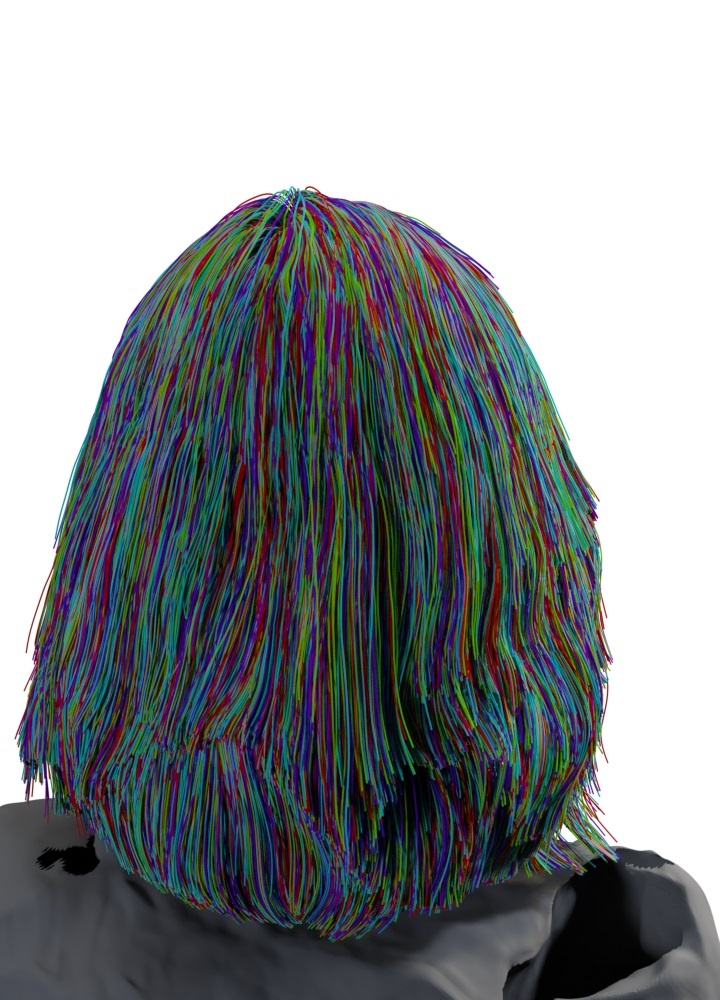} & \hspace{-0.31cm}
       \includegraphics[width=0.185\textwidth]{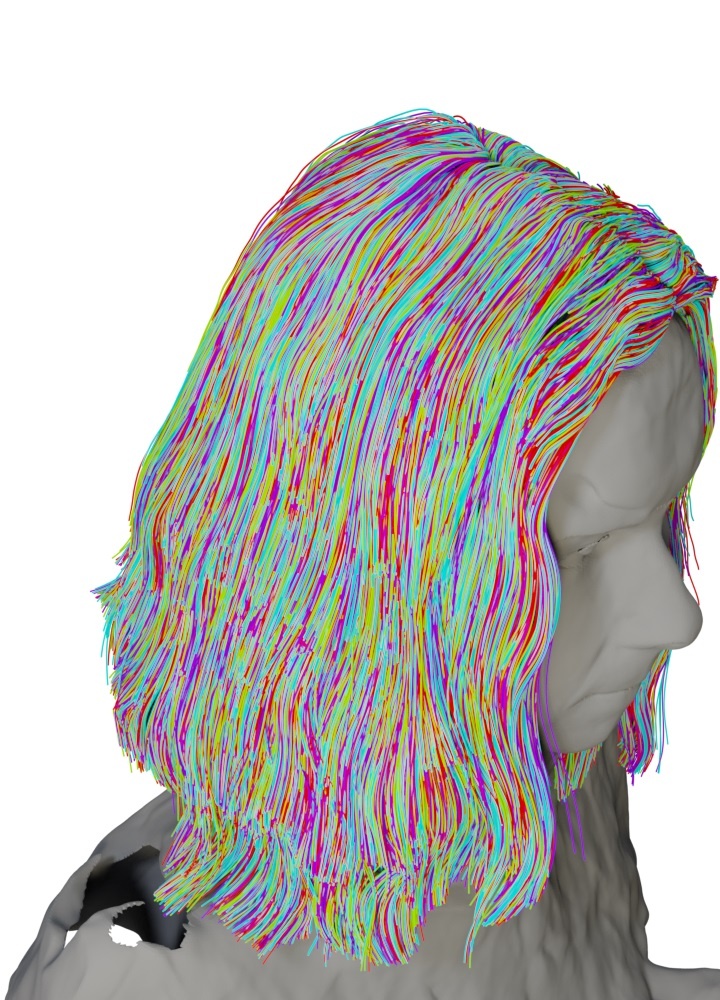} & \hspace{-0.31cm} 
        \includegraphics[width=0.185\textwidth]{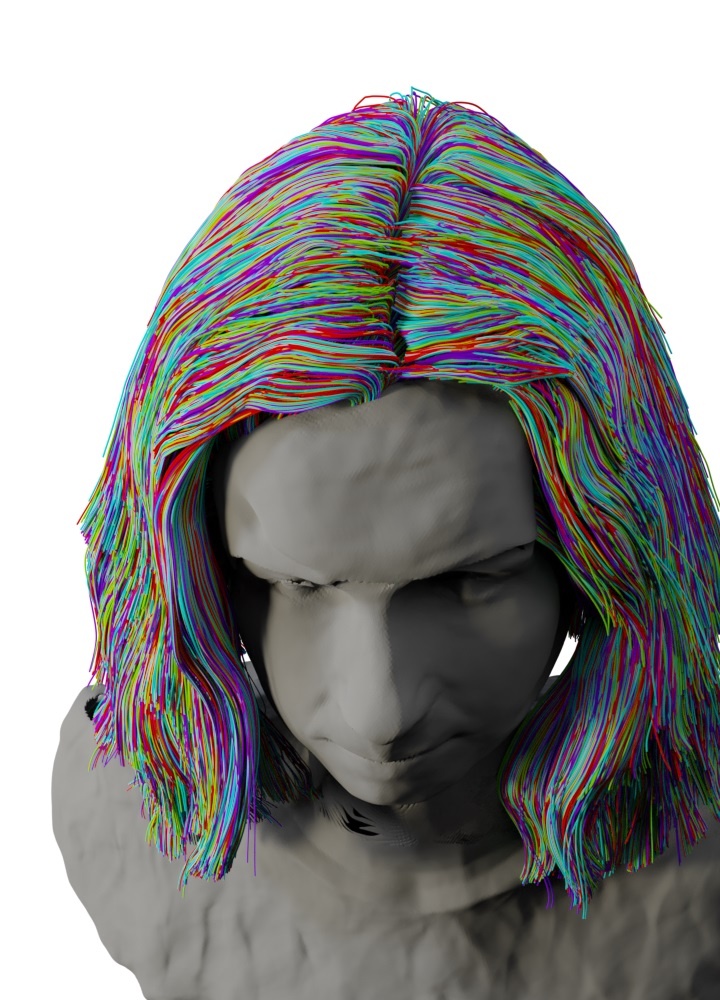} & \hspace{-0.31cm} 
        \includegraphics[width=0.185\textwidth]{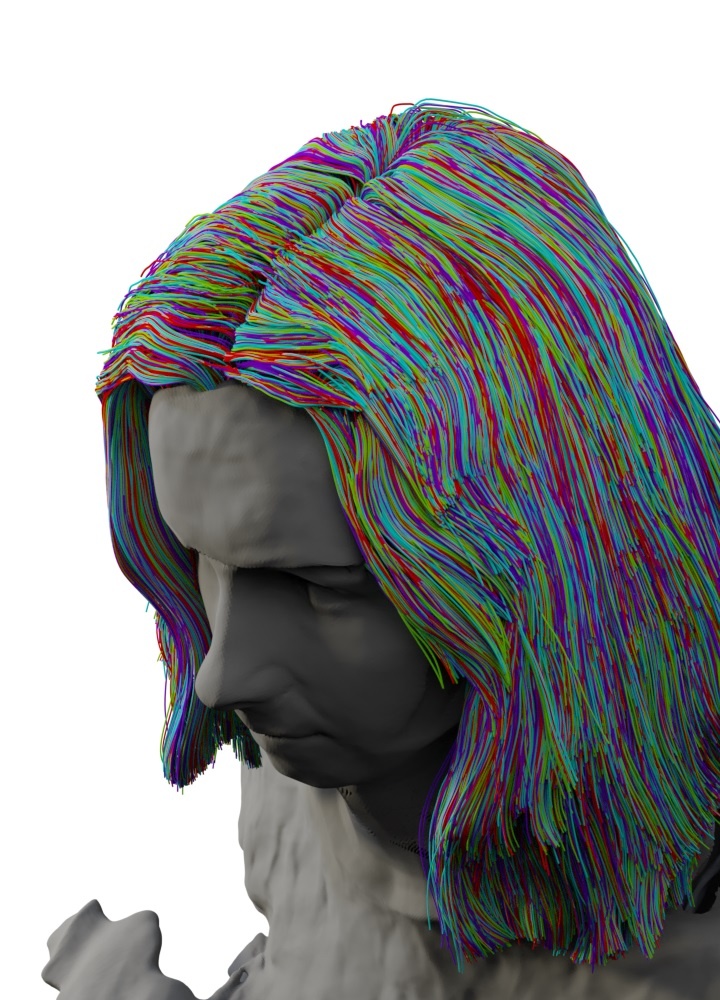} & \hspace{-0.31cm} 
        \includegraphics[width=0.185\textwidth]{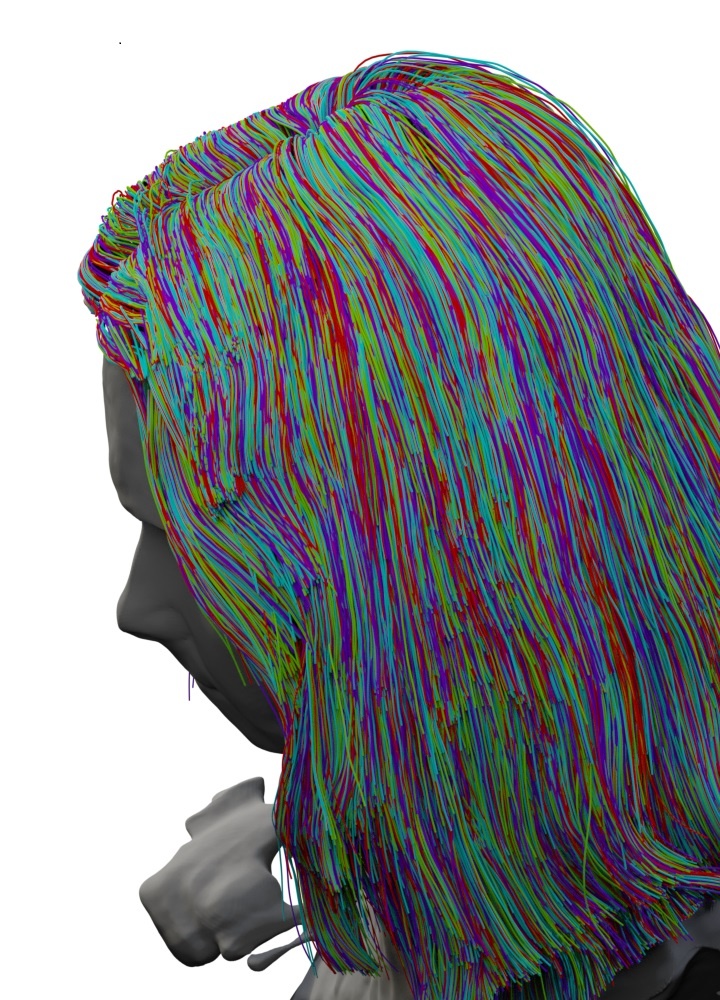}

        \\

    \end{tabular}
    \caption{Additional reconstruction results of our method on monocular videos in arbitrary lighting conditions.} 
    \label{fig:colmap_scene_nastya_suppmat}
\end{figure*}

%% file: figures_suppmat/colmap_person/colmap_person.tex
\begin{figure*}
    \begin{tabular}{ccccc}

        \includegraphics[width=0.185\textwidth]{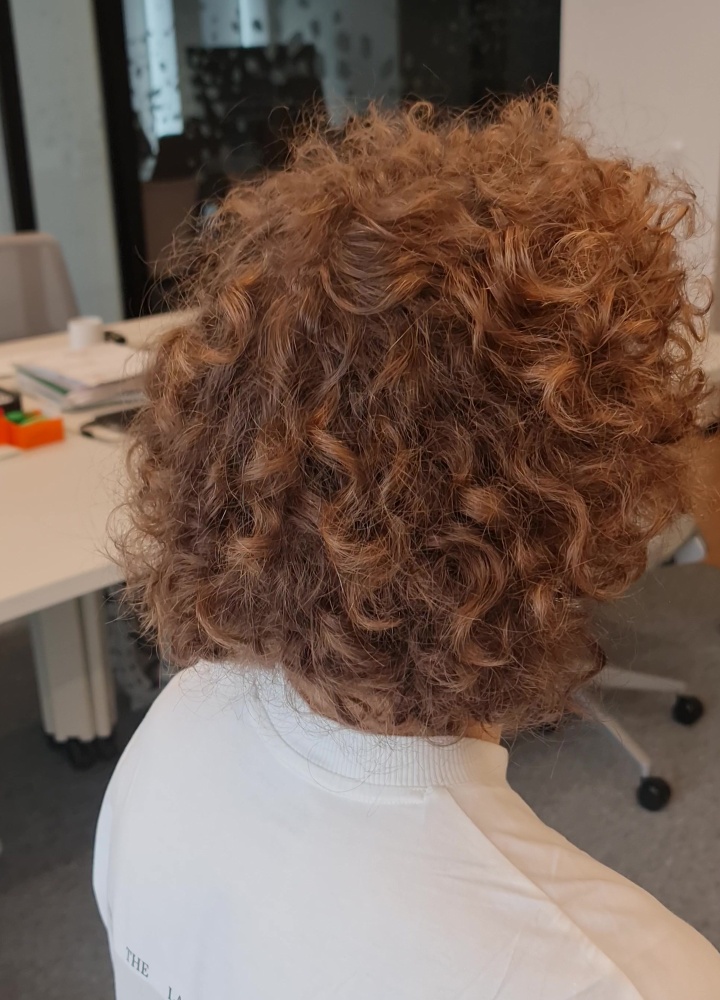} & \hspace{-0.31cm}
        \includegraphics[width=0.185\textwidth]{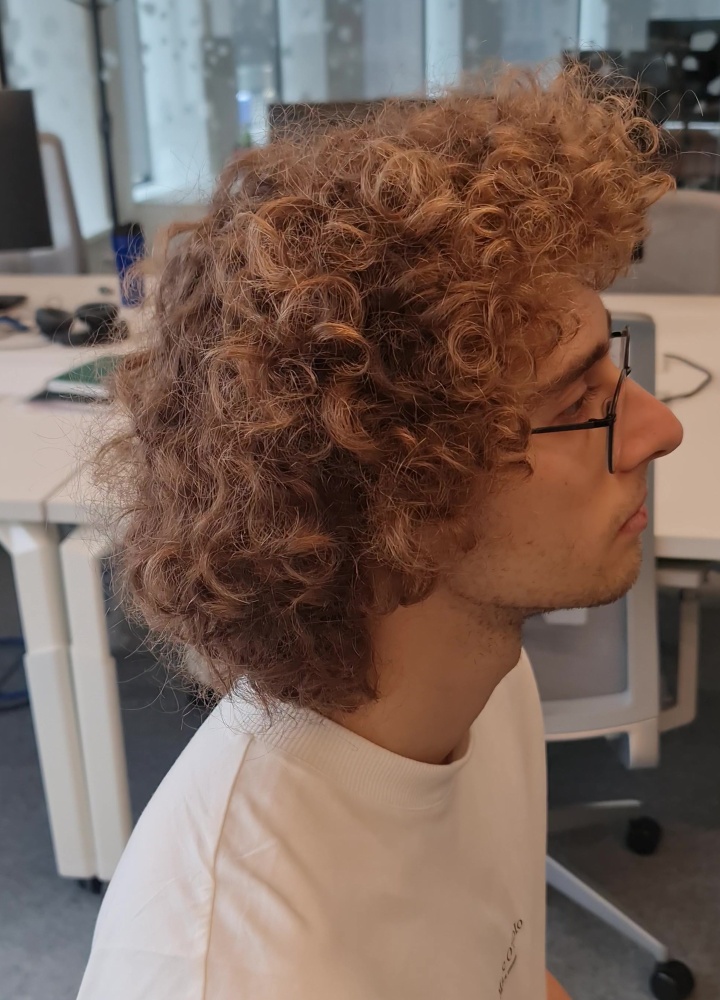} & \hspace{-0.31cm} 
        \includegraphics[width=0.185\textwidth]{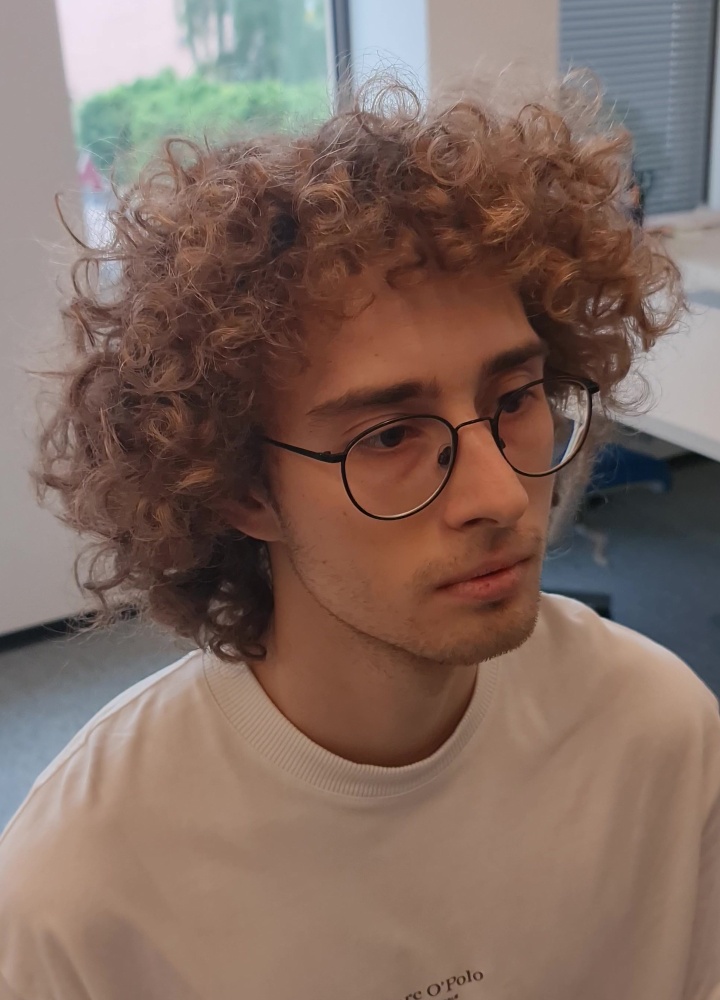} & \hspace{-0.31cm} 
        \includegraphics[width=0.185\textwidth]{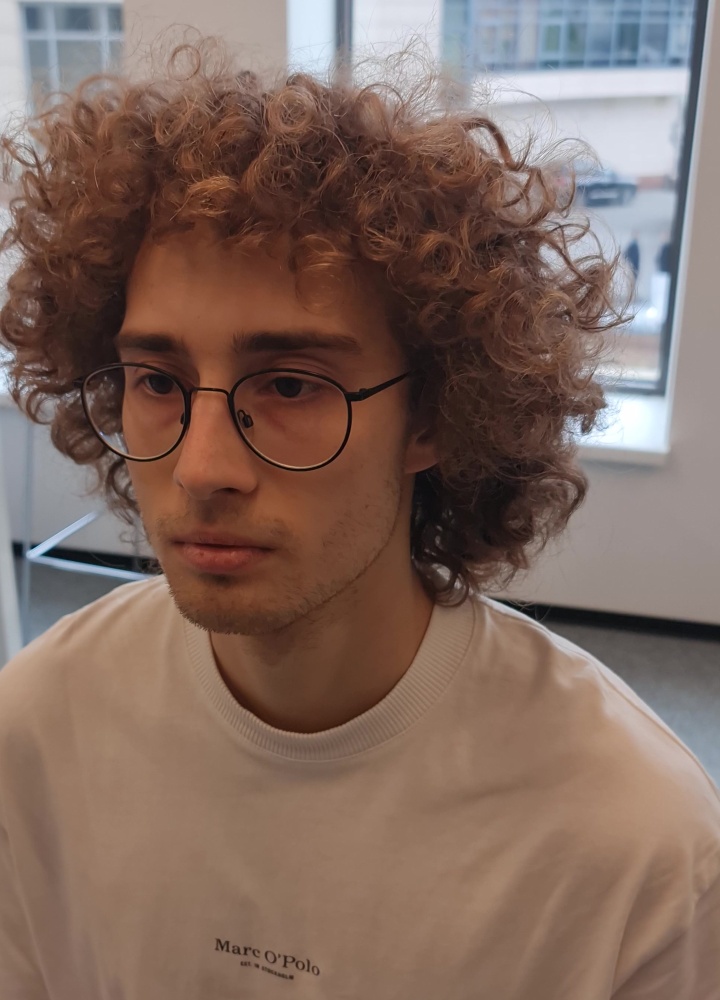} & \hspace{-0.31cm} 
        \includegraphics[width=0.185\textwidth]{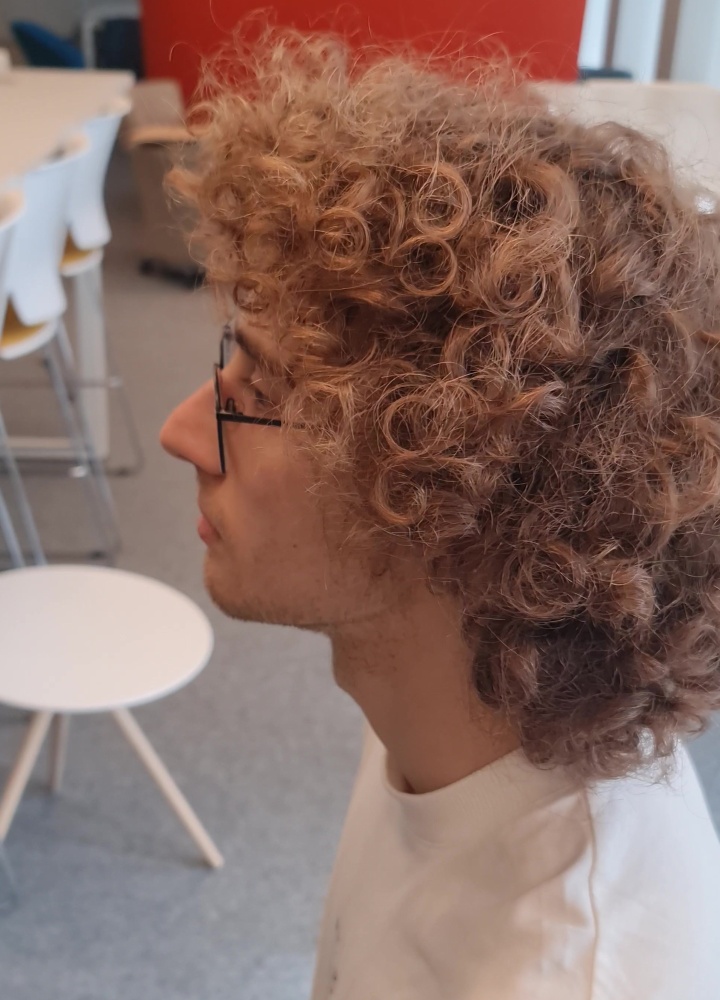} \\ %
        \includegraphics[width=0.185\textwidth]{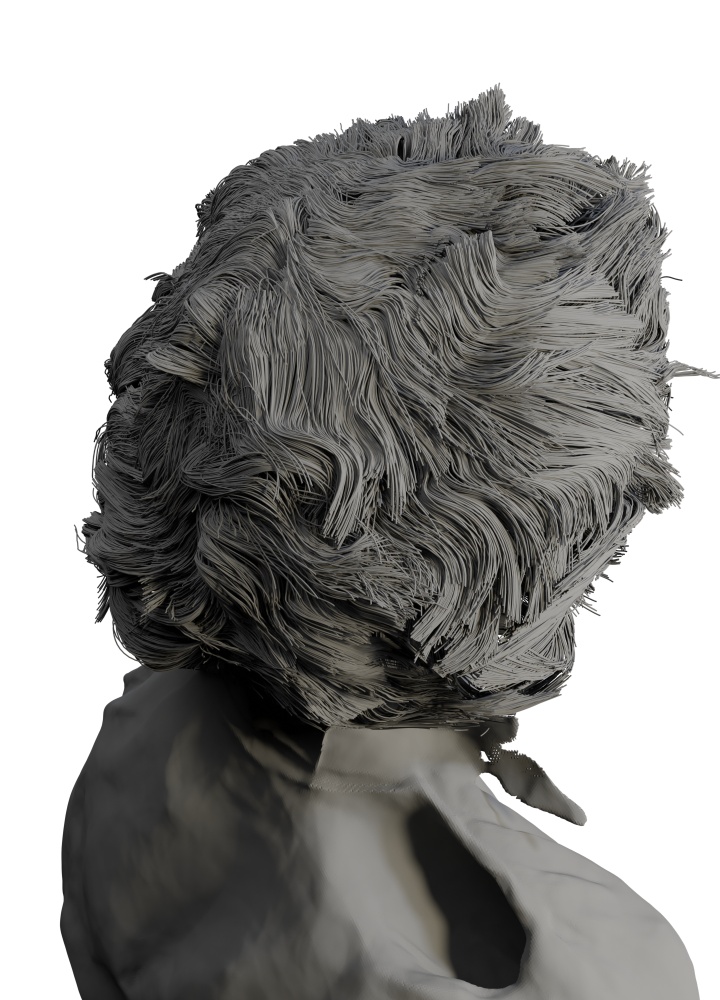} & \hspace{-0.31cm}
       \includegraphics[width=0.185\textwidth]{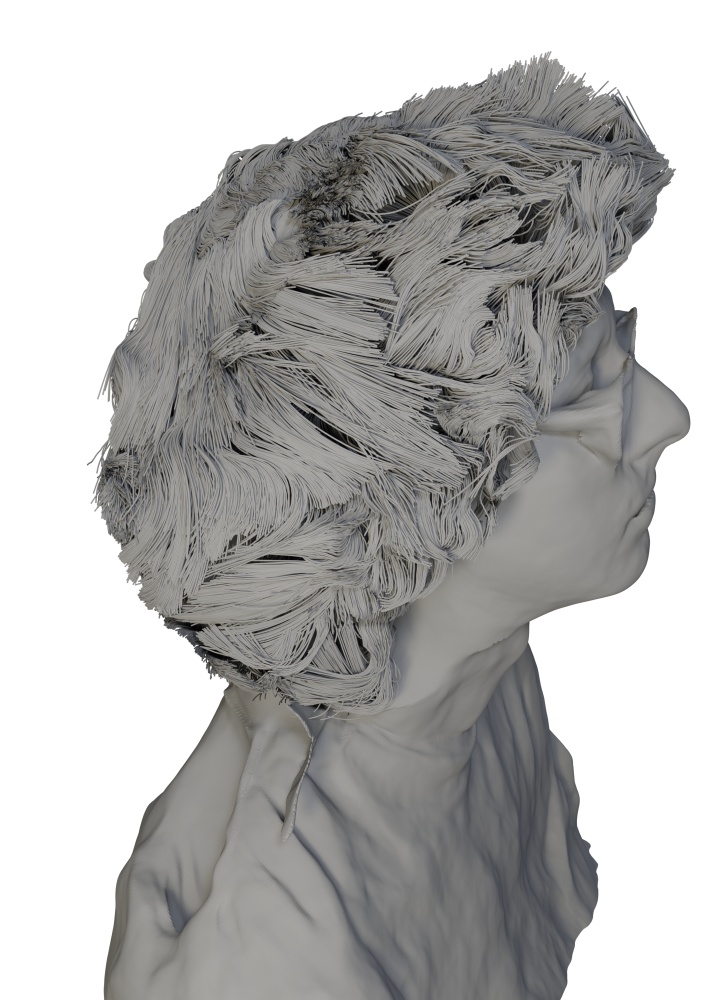} & \hspace{-0.31cm} 
        \includegraphics[width=0.185\textwidth]{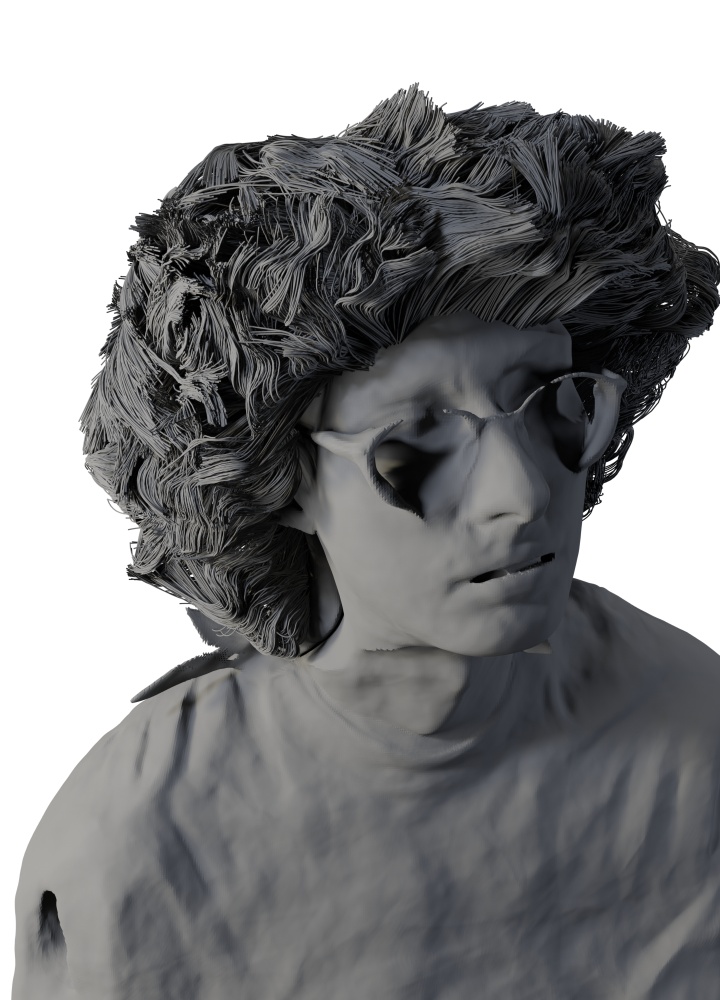} & \hspace{-0.31cm} 
        \includegraphics[width=0.185\textwidth]{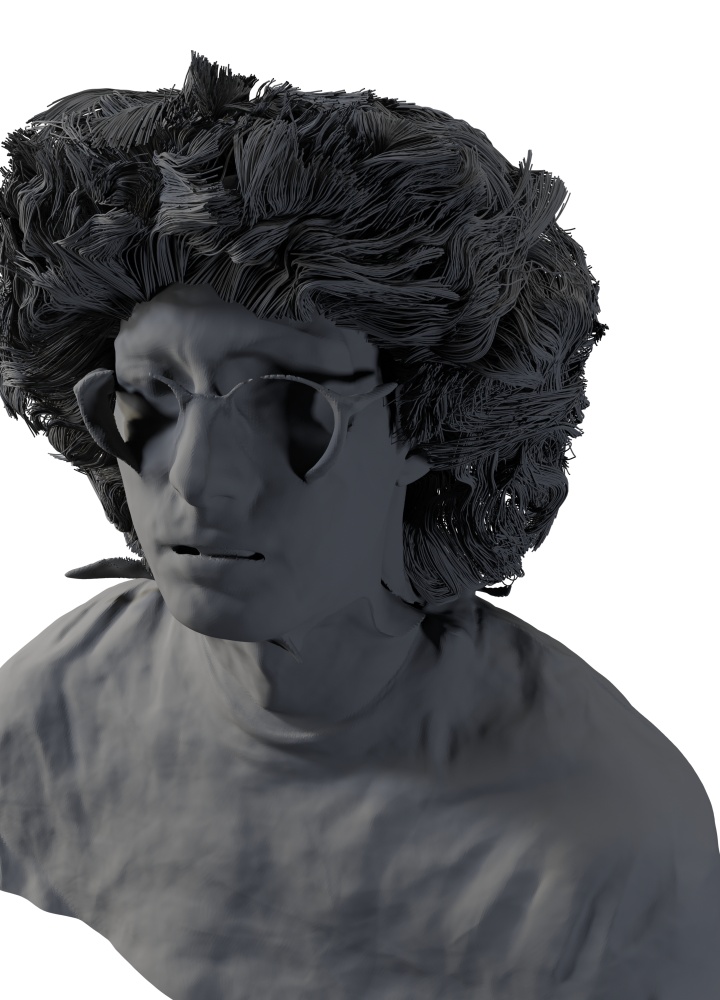} & \hspace{-0.31cm} 
        \includegraphics[width=0.185\textwidth]{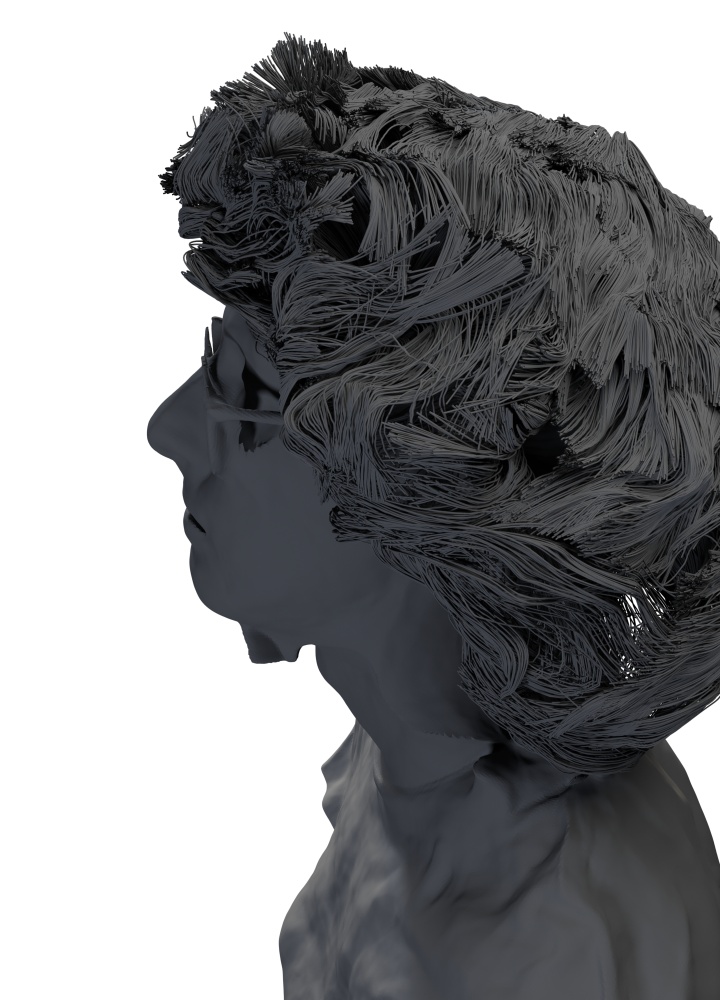}

        \\

    \end{tabular}
    \caption{The main limitation of our method is related to curly hair reconstruction, which will be addressed in future work. } 
    \label{fig:colmap_scene_person_suppmat}
\end{figure*}
\clearpage

%% file: figures_suppmat/qualitative/comparison_12_views.tex
\begin{figure*}
    \begin{tabular}{ccc}
        \includegraphics[width=0.2\textwidth]{figures/final_h3ds_comparison/5d_img_0020.jpg} & \hspace{-0.31cm}
        \includegraphics[width=0.2\textwidth]{figures/final_h3ds_comparison/5d_20_deepmvs.jpg} & \hspace{-0.31cm} 
        \includegraphics[width=0.2\textwidth]{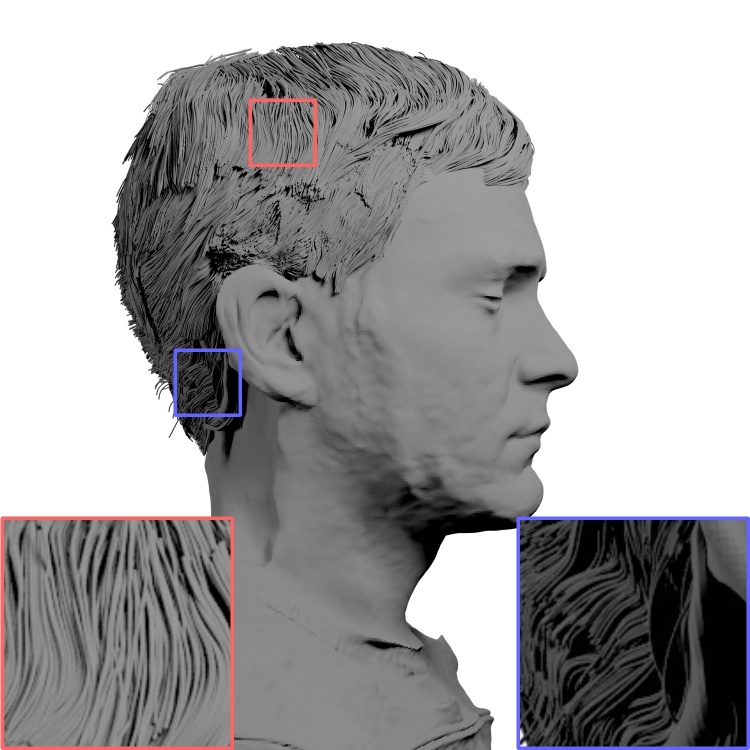} \\ %
        \includegraphics[width=0.2\textwidth]{figures/final_h3ds_comparison/5d_img_0039.jpg} & \hspace{-0.31cm}
        \includegraphics[width=0.2\textwidth]{figures/final_h3ds_comparison/5d_39_deepmvs.jpg} & \hspace{-0.31cm} 
        \includegraphics[width=0.2\textwidth]{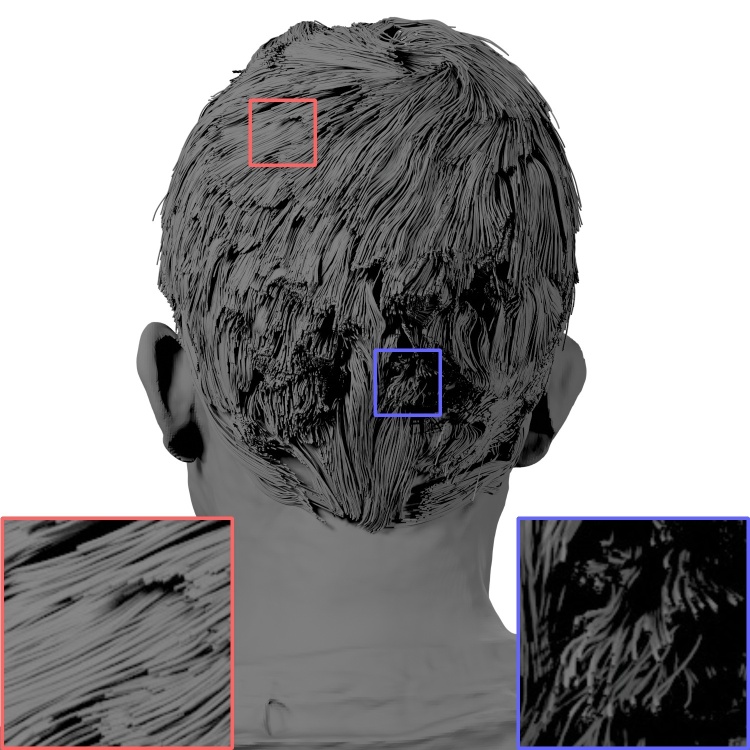} \\ %
        \includegraphics[width=0.2\textwidth]{figures/final_h3ds_comparison/66_img_0031.jpg} & \hspace{-0.31cm}
        \includegraphics[width=0.2\textwidth]{figures/final_h3ds_comparison/66_31_deepmvs.jpg} & \hspace{-0.31cm}        
        \includegraphics[width=0.2\textwidth]{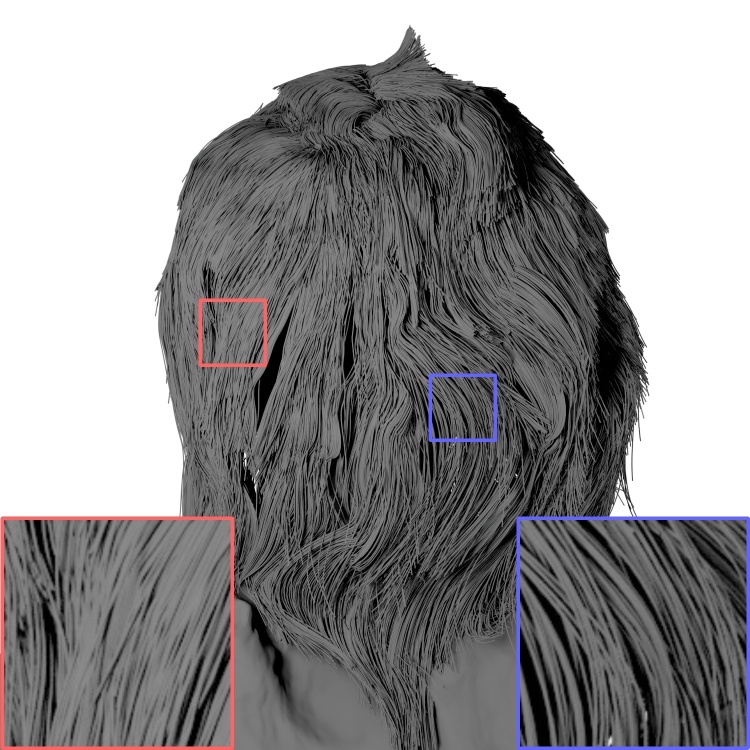} \\ %
         \includegraphics[width=0.2\textwidth]{figures/final_h3ds_comparison/66_img_0013.jpg} & \hspace{-0.31cm}
        \includegraphics[width=0.2\textwidth]{figures/final_h3ds_comparison/66_13_deepmvs.jpg} & \hspace{-0.31cm}        
        \includegraphics[width=0.2\textwidth]{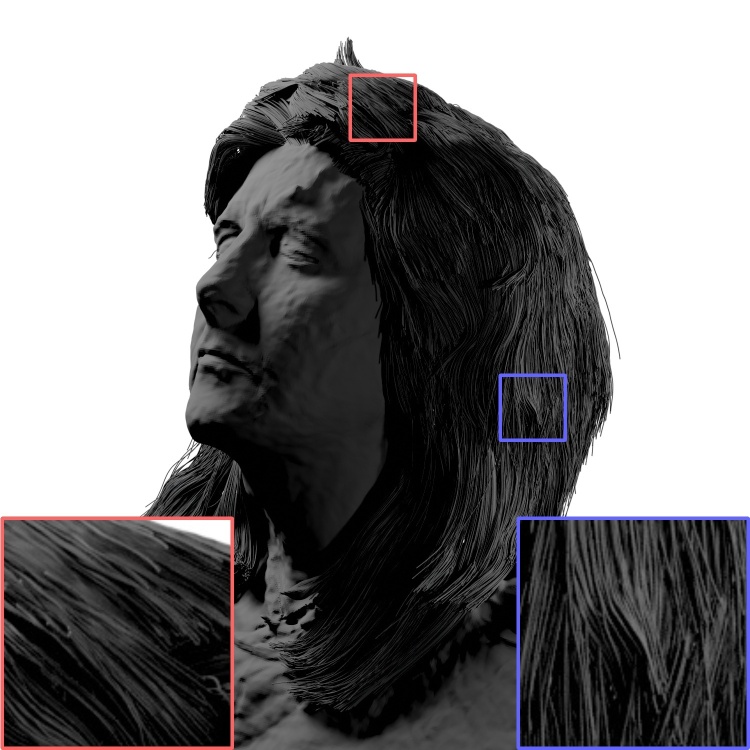} \\ %
        \includegraphics[width=0.2\textwidth]{figures/final_h3ds_comparison/1b_img_0020.jpg} & \hspace{-0.31cm}
        \includegraphics[width=0.2\textwidth]{figures/final_h3ds_comparison/1b_20_deepmvs.jpg} & \hspace{-0.31cm}
        \includegraphics[width=0.2\textwidth]{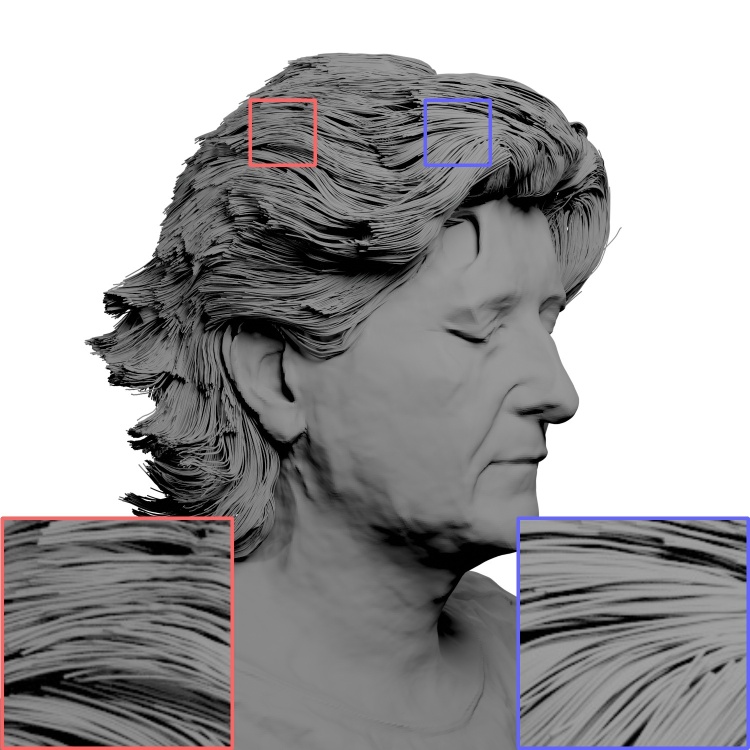} \\ %
        \includegraphics[width=0.2\textwidth]{figures/final_h3ds_comparison/1b_img_0059.jpg} & \hspace{-0.31cm}
        \includegraphics[width=0.2\textwidth]{figures/final_h3ds_comparison/1b_59_deepmvs.jpg} & \hspace{-0.31cm} 
        \includegraphics[width=0.2\textwidth]{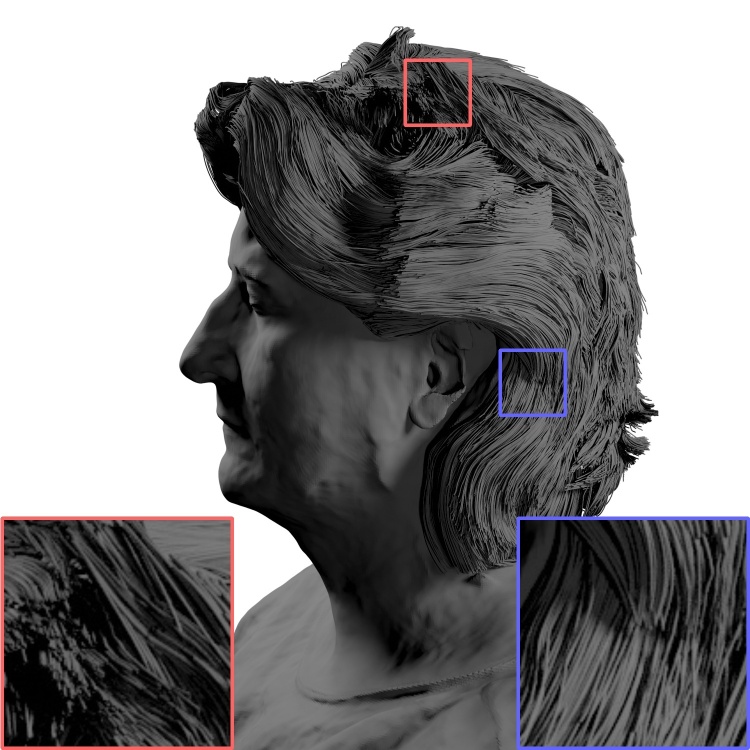} \\ %
        \textbf{Image}  & \hspace{-0.31cm}  \textbf{DeepMVSHair} & \hspace{-0.31cm} \textbf{Ours (12 views)}
    \end{tabular}
    \vspace{0.1cm}
    \caption{Qualitative comparison using 12 views from real-world multi-view scenes~\cite{Ramon2021H3DNetFH}. Digital zoom-in is recommended.}
    \label{fig:geom_compare_suppmat_12_views}
\end{figure*}
\clearpage